\documentclass{article}
\usepackage{codefuse_tech_report}
\usepackage[colorlinks = true,
            linkcolor = blue,
            urlcolor  = blue,
            citecolor = blue,
            anchorcolor = blue]{hyperref}   
\usepackage{microtype}
\usepackage{xurl}
\usepackage{booktabs}
\usepackage{enumitem}
\usepackage{multicol}
\usepackage{CJKutf8}
\usepackage{siunitx}
\usepackage{floatflt}
\usepackage{graphicx}
\usepackage{tikz}
\usetikzlibrary{positioning, arrows.meta, shapes.geometric, calc}
\usepackage{wrapfig}
\usepackage{authblk}
\usepackage{lipsum}
\usepackage{algorithm}
\usepackage{algorithmicx}
\usepackage{algpseudocode}
\usepackage{subfigure}
\usepackage{multirow}
\usepackage{pifont}  

\usepackage{setspace} 
\usepackage{pdflscape} 

\algnewcommand{\LeftComment}[1]{\Statex \(\triangleright\) #1}

\usepackage{array}
\usepackage{amsmath}
\usepackage{amssymb}
\usepackage{mathtools}
\usepackage{amsthm}

\usepackage[capitalize,noabbrev]{cleveref}
\usepackage{adjustbox} 

\theoremstyle{plain}

\theoremstyle{definition}

\theoremstyle{remark}

\newcommand{\ie}{\textit{i.e.},~}
\newcommand{\eg}{\textit{e.g.},~}

\colmfinalcopy

\usepackage[utf8]{inputenc}
\usepackage[T1]{fontenc}

\usepackage{amsmath,amsfonts,bm}

\def\eqref#1{equation~\ref{#1}}

\def\1{\bm{1}}

\def\vc{{\bm{c}}}

\def\ve{{\bm{e}}}

\def\vh{{\bm{h}}}

\def\vu{{\bm{u}}}
\def\vv{{\bm{v}}}
\def\vw{{\bm{w}}}

\def\vz{{\bm{z}}}

\def\mE{{\bm{E}}}

\def\mH{{\bm{H}}}

\DeclareMathAlphabet{\mathsfit}{\encodingdefault}{\sfdefault}{m}{sl}
\SetMathAlphabet{\mathsfit}{bold}{\encodingdefault}{\sfdefault}{bx}{n}

\def\sM{{\mathbb{M}}}

\def\sS{{\mathbb{S}}}

\def\sX{{\mathbb{X}}}

\newcommand{\E}{\mathbb{E}}

\newcommand{\R}{\mathbb{R}}

\usepackage{caption} 
\usepackage{adjustbox} 
\usepackage{fontawesome5}

\usepackage{tcolorbox}
\tcbuselibrary{skins,breakable}

\tcbuselibrary{skins}

\newcommand{\highlightbox}[3]{
\begin{tcolorbox}[
    enhanced jigsaw,
    breakable,
    pad at break*=1mm,
    colback=#1!5!white,
    colframe=#1,
    title=#2
]
#3
\end{tcolorbox}
}

\usepackage{color}
\usepackage[table]{xcolor}
\usepackage{soul} 

\definecolor{tred}{RGB}{251, 130, 132}
\definecolor{torange}{RGB}{247, 162, 116}
\definecolor{tyellow}{RGB}{251, 218, 140}
\definecolor{tgreen}{RGB}{127, 204, 181}
\definecolor{tblue}{RGB}{89, 177, 215}
\definecolor{insightblue}{RGB}{162, 210, 255}
\definecolor{questionred}{RGB}{255, 175, 204}

\newcommand{\simfunc}{\operatorname{sim}}

\newcommand{\yes}{$\bullet$}
\newcommand{\avail}{$\circ$}
\newcommand{\na}{\textendash}

\definecolor{paperDark}{HTML}{0F172A}       
\definecolor{paperSubText}{HTML}{334155}    
\definecolor{paperBorder}{HTML}{94A3B8}     
\definecolor{paperArrow}{HTML}{475569}      
\definecolor{bgNeutral}{HTML}{F8FAFC}       

\definecolor{c1Bg}{HTML}{FDF0E7}  \definecolor{c1Border}{HTML}{D35400}  \definecolor{c1Text}{HTML}{A04000} 
\definecolor{c2Bg}{HTML}{EAF3FA}  \definecolor{c2Border}{HTML}{2980B9}  \definecolor{c2Text}{HTML}{21618B} 
\definecolor{c3Bg}{HTML}{E8F6F3}  \definecolor{c3Border}{HTML}{16A085}  \definecolor{c3Text}{HTML}{117A64} 
\definecolor{depBg}{HTML}{F4F6F6} \definecolor{depBorder}{HTML}{7F8C8D} \definecolor{depText}{HTML}{616D6E} 
\definecolor{coreBg}{HTML}{F1F5F9} \definecolor{coreBorder}{HTML}{64748B} \definecolor{coreText}{HTML}{0F172A} 

\definecolor{decorBlueBg}{HTML}{EFF6FF}   \definecolor{decorBlueBorder}{HTML}{60A5FA}
\definecolor{decorOrangeBg}{HTML}{FFF7ED} \definecolor{decorOrangeBorder}{HTML}{FB923C}

\title{From Fragmentation to Fusion: A Survey and Roadmap for Omnimodal Embeddings}

\newcommand{\smallfootnotesize}{\fontsize{7.57pt}{9pt}\selectfont}

\vspace{-7pt}
\author{%
Ziyin Zhang\thanks{\smallfootnotesize Core contributors: Ziyin \textless daenerystargaryen@sjtu.edu.cn\textgreater, Yuyu \textless yuyuguo1994@gmail.com\textgreater, Ying \textless ying.zhang@anu.edu.au\textgreater, Siqiao \textless siqiao\_xue@163.com\textgreater. \faGithub ~\url{https://github.com/codefuse-ai/Awesome-Omnimodal-Embeddings}}$^{\phantom{*},1,2}$
~~Yuyu Guo$^{*,1}$
~~Ying Zhang$^{*,1,3}$
~~Siqiao Xue$^{*,1}$
\\
\vspace{-7pt}
\bf
~~Zihan Liao$^{1}$ 
~~Jin Qin$^{1}$
~~Wenjie Yang$^{1}$
~~Siyuan Yang$^{1}$
~~Shijie Li$^{1,2}$
~~Cheng Chen$^{1}$
\\
\vspace{2.5pt}
\bf
~~Hao Wu$^{1,4}$
~~Zhaoyang Zhu$^{1}$
~~Ruipeng Li$^{1}$
~~Zhaokun Jiang$^{5}$
~~Lei Lei$^{1}$
~~Rui Wang$^{1}$
\\
\vspace{2.5pt}
\bf
~~Rui Wang$^{\dagger,2}$
~~Hang Yu\thanks{\smallfootnotesize Correspondence to: Rui Wang \textless wangrui12@sjtu.edu.cn\textgreater, Hang Yu \textless hyu1@e.ntu.edu.sg\textgreater.}$^{\phantom{\dagger},1}$ 

\vspace{6.5pt}
$^1$Ant Group\ \ \ \ \ $^2$Shanghai Jiao Tong University%
\ \ \ \ \ $^3$Australian National University\\
\vspace{3pt}
\bf
$^4$Tongji University\ \ \ \ \ $^5$National University of Singapore\\
\vspace{5pt}
}

\colmfinalcopy 

\begin{document}

\maketitle

\vspace{-10pt}
\begin{abstract}
Embedding models underpin virtually every modern AI system, from search engines to recommendation systems.
Despite shared mathematical foundations, research on embedding models remains fragmented by modality: the communities develop largely in isolation, seldom exchanging insights.
We argue that this fragmentation obscures a deeper convergence. Driven by contrastive learning, Transformer architectures, and a granularity shift, embedding research across modalities is converging toward a common set of principles, even as modality-specific challenges remain underexplored.
To reason across this convergence without conflating it, we propose a framework of three progressively harder capabilities: \emph{coverage}, \emph{comparability}, and \emph{compositionality}. Three paths toward unification satisfy these in turn: \emph{transformation} broadens coverage, \emph{alignment} enables comparability, and \emph{joint embedding}, the terminus we advocate, aims at compositionality.
This position paper traces the resulting evolution, providing the first unified, cross-cutting treatment of omnimodal embedding research.
\end{abstract}

\begin{figure*}[th]
    \centering
    \vspace{-0.3cm}
    \includegraphics[width=\textwidth]{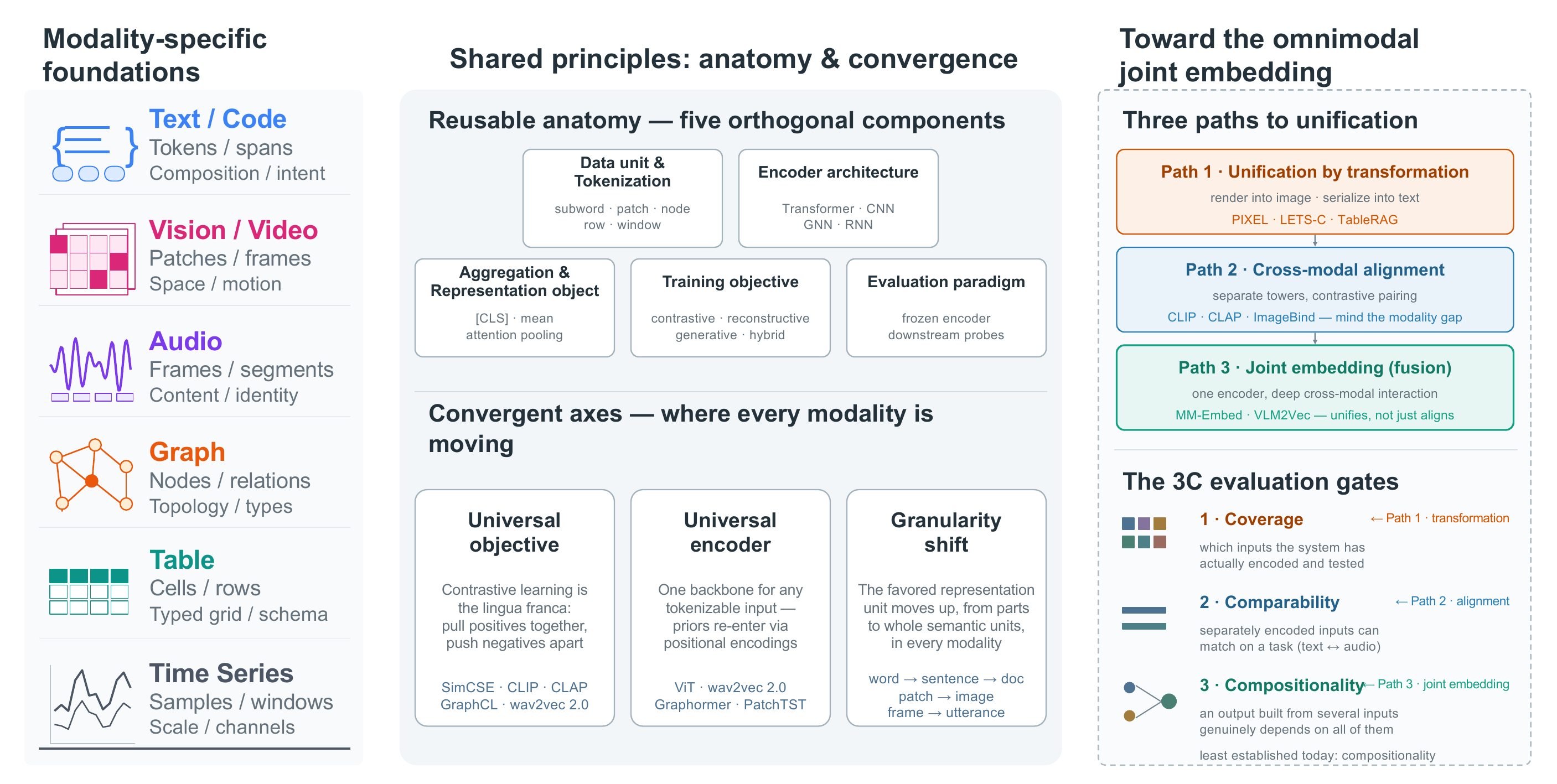}
    \vspace{-0.65cm}
    \caption{A research map of omnimodal embeddings. \textbf{Left}: modality-specific foundations---each with its own data units and challenges. \textbf{Middle}: top, a reusable anatomy that any modality decomposes into; bottom, the three convergent axes. \textbf{Right}: top, the three paths to unification, from coarse to deep; bottom, the 3C lens any omnimodal claim must pass.}
    \label{fig:overall-convergence-map}
\end{figure*}

\newpage
\begin{spacing}{0.9}
\tableofcontents
\end{spacing}
\newpage

\part{The Unified Foundation}
\label{part:foundation}
\section{Introduction}
\label{sec:intro}

At the heart of modern machine learning lies a deceptively simple idea:
map complex, unstructured data into a vector space where geometric
relationships carry semantic meaning.
The ubiquity of such embedding models is striking: in information retrieval
dense embeddings have largely replaced sparse bag-of-words representations like
BM25~\citep{robertson2009bm25} to enable semantic
search~\citep{karpukhin2020dpr, izacard2022contriever}; in recommendation,
learned item embeddings power web-scale
retrieval and ranking; and in cross-modal settings, aligned embedding
spaces---popularized at web scale by CLIP~\citep{radford2021clip} and
CLAP~\citep{clap}---let us
search images with text and describe audio with language.

Despite this central role, the study of embedding models has been
fragmented along modality boundaries.
We argue that this fragmentation is the central obstacle to progress, and
that it runs deeper than is usually acknowledged, manifesting on three
interlocking levels: in research, in evaluation, and in surveys.

\subsection{The Fragmentation Problem: Silos of Research, Evaluation, and Surveys}
\label{sec:intro:silos}

\paragraph{What counts as a modality?}
Our unit of analysis is the coarse modality family: text and code,
image and video, audio, graph, table, and time series. The fragmentation
we diagnose lives at family level:
it is communities, venues, and benchmarks that are siloed, and it is at that
level that convergence must be measured.
In contrast, modality counts in the literature may also include
intra-family derivatives---views and channels of a single underlying object
produced by signal processing or annotation.
As an example, 4M-21~\citep{bachmann20244m21} claims to cover twenty-one modalities,
yet almost all of them (such as depth, segmentation, edges, human pose, and feature
maps) are views of the same visual scenes. In
our taxonomy they are not counted as peer modalities, but derived views within the vision family (refer to \Cref{tab:modality-ontology} in \Cref{sec:frontier:properties} for details).

\paragraph{Research silos.}
Text embedding research~\citep{gao2021simcse, wang2022e5, su2022instructor,
lee2025nvembed} has evolved largely independently of vision embedding
research~\citep{caron2021dino, he2022mae}, audio self-supervised
learning~\citep{baevski2020wav2vec2, hsu2021hubert}, graph representation
learning~\citep{bordes2013transe, grover2016node2vec, khoshraftar2024survey},
time-series embeddings~\citep{ts2vec, nie2023times}, and table
representation learning~\citep{tabert2020, tapas2020, turl2020}.
Each community holds its own workshops, releases its own checkpoints, and
optimizes its own benchmarks; methods that would be ported
straightforwardly within one literature are repeatedly reinvented from
scratch in another, and the connections between them go unnamed.

\paragraph{Evaluation silos.}
This research-level fragmentation is faithfully mirrored in evaluation.
Text, code, audio, and vision have each matured their own frozen-encoder
benchmark---MTEB and its multilingual extension
MMTEB for text~\citep{muennighoff2023mteb,enevoldsen2025mmteb},
CoIR for code~\citep{coir2025}, SUPERB and its successors for
audio~\citep{superb, hear, mseb, maeb}, and retrieval/classification
suites for vision~\citep{russakovsky2015imagenet,radenovic2018revisiting}.
In contrast, tables and time series are still scored
mainly through task heads and prediction archives: table QA/fact-verification
suites~\citep{wtq2015, spider2018, tabfact2020}, with MTEB analogs
only now emerging~\citep{tembed2026,tabembed2026}, and
time series classification archives~\citep{bagnall2018ueaarchive, dau2019ucrarchive}
alongside forecasting archives~\citep{godahewa2021monash}. Yet, there is no single
yardstick that measures how a
methodology (not just a model) generalizes across modalities.
We have islands of evaluation, but no shared sea to compare them on.

\paragraph{Survey silos.}
Finally, the fragmentation reaches surveys.
We have excellent monolithic reviews of single modalities: text embedding
surveys~\citep{surveytextembed2024, surveytextembed2025}, vision
self-supervised reviews~\citep{surveyvisionssl2023}, speech representation
surveys~\citep{surveyspeechrep2023}, graph embedding
reviews~\citep{khoshraftar2024survey}, and tabular representation
reviews~\citep{jiangtabularsurvey2026}.
These surveys are invaluable within their fields, but they are themselves
evidence of the fragmentation: they almost never cross the modality boundary to
systematically compare and connect the core methodologies of sibling communities.
A graph-embedding survey may discuss view augmentation at length without ever
observing that text's SimCSE achieves implicit positive pair
construction through dropout~\citep{gao2021simcse};
a vision survey may extol masked autoencoding
without noting that audio arrived at the same principle a year
apart~\citep{baevski2020wav2vec2, hsu2021hubert, audiomae}.
Each survey charts its own domain; none builds the bridges between them.

This fragmentation across research, evaluation, and surveys is not an
impression but a measurable structure, as we quantify in Appendix~\ref{app:quant}.
A combined analysis of 1{,}352 formally published embedding papers
(2023--2025) and a decade of arXiv preprints (2015--2025) indicates that the overwhelming majority of embedding research is single-modality, each modality publishing in its home community's venues, while works that explicitly cross modality boundaries
account for under 5\% of both samples and emerge only recently.
This is the core obstacle---and the opportunity---this paper sets out to
address.

\subsection{Why Fragmentation Matters}
\label{sec:intro:why}

At first sight, fragmentation is an organizational complaint rather than a
scientific one: a field can be fragmented and healthy, with each silo
producing the best possible encoder for its own modality.
But two independent lines of evidence suggest that embedding research cannot
stop there---one from the science of intelligence, one from the trajectory
of AI itself.

\paragraph{Concepts must be grounded in more than symbols.}
The physical world is inherently multimodal: real-world entities and events
manifest simultaneously across visual, auditory, tactile, and temporal channels,
rather than existing in isolated sensory silos. Mirroring this reality, 
the first line of evidence is that human intelligence does not organize
knowledge by modality.
Meaning cannot be assembled from a closed system of symbol-to-symbol
mappings~\citep{harnad1990symbol}: human concepts are grounded in multimodal
perceptual and motor experience~\citep{barsalou1999perceptual}, and are
organized in the brain in a hub-and-spoke
architecture---a modality-invariant hub carrying conceptual knowledge
for all modalities, flanked by modality-specific
spokes~\citep{patterson2007where, lambonralph2017neural}.
Rather than keeping channels isolated, human cognition fuses multisensory
cues~\citep{ernst2002humans} and bootstraps word meaning from
cross-modal co-occurrence before language
acquisition~\citep{gogate1998intersensory, gogate2001intersensory, smith2008infants}.
While language models recover some perceptual
statistics from corpus co-occurrence~\citep{abdou2021can,
huh2024platonic}, text remains a lossy projection of experience: it cannot
supervise knowledge that is never put into words, nor provide the
cross-modal consistency check that disambiguates and stabilizes concepts.
A representation literature fragmented by modality is, in these terms,
manufacturing spokes while nobody builds the hub.

\paragraph{Representation is falling behind generation.}
The second line of evidence is more recent, and more measurable.
On the generative side of AI, omnimodality is deployed reality: GPT-4o has
jointly consumed text, audio, image, and video within a single model since
May 2024~\citep{openai2024gpt4o}, and any-to-any generative research covers
the same four natural modalities in both directions~\citep{wu2023nextgpt}.
On the embedding side, breadth is achieved without fusion: pairwise-aligned
spaces have spanned many modalities since ImageBind and
LanguageBind~\citep{imagebind, languagebind, chen2026codebind}, but
verified joint fusion (where measurable compositional gains are
confirmed beyond processing inputs together) has so far been
obtained only for text--image pairs~\citep{yang2026solar}
and partially for composed text, vision, and audio
queries~\citep{huynh2026omniret, wang2026omniue}.
Fused omni embedding models and benchmarks covering
text, image, video, and audio appeared only in late
2025~\citep{lcoembedding, omniembednemotron, wave2025, e5omni2026, jinav5omni, huang2026mmebv3}---almost two years after the generative side
crossed the same threshold.
The consequence is an asymmetry inside the very substrate on which
generative AI depends: models that jointly perceive four modalities are
indexed and remembered by retrieval systems that are bimodal,
pairwise-aligned, or text-centric, with the retrieval-augmented generation (RAG)
pipeline being the most common case.
Fragmentation is not slowing a peripheral subfield; it is the
bottleneck of the omni stack itself.

The remedy is neither to abandon the single-modality literatures nor to
merely catalog them.
Our contribution is not to replace the deep, expert, single-modality
surveys but to provide a
meta-level synthesis: to build the conceptual bridges between those
continents and reveal the shared principles beneath them.
The synthesis is worth more than scholarly tidiness, because what a joint
embedding built on those principles would give is already visible in
miniature: multimodal retrieval over a single shared
index~\citep{wei2024uniir}, generation grounded in retrieved images and text
rather than text alone~\citep{murag, yasunaga2023racm3, mragbench}, robot memory stored as text
captions rather than raw experience~\citep{remembr}, and emergent zero-shot
retrieval and composition between modalities that never co-occurred in
training~\citep{imagebind}.
Each payoff today exists only on the few modality pairs its prototype
happens to span. Giving them a common geometry is what the rest of this
paper sets out to do.

\subsection{Our Position and Roadmap}
\label{sec:intro:roadmap}

Looking past the silos, four convergent trends are quietly unifying all the
modalities.
\textbf{Contrastive learning}~\citep{vandenoord2018cpc} has become the universal training
objective;
\textbf{Transformers}~\citep{vaswani2017attention}, increasingly repurposed from generative
LLMs and MLLMs, serve as universal encoders;
the two-stage \textbf{pretrain--finetune paradigm} standardizes the training
workflow;
and a \textbf{granularity shift} across all modalities moves the favored representational
object upward from local parts (tokens, pixels, frames, nodes) to complete semantic
units (sentences, images, utterances, graphs, series).

These convergences are not coincidental.
The Platonic Representation Hypothesis~\citep{huh2024platonic}
suggests that representations across
modalities may be converging toward a shared statistical model of reality---a
``platonic'' representation space.
Empirically, \citet{uniembedgeom2025} demonstrate a universal latent semantic
structure across text embedding spaces, enabling unsupervised translation
of embeddings from one vector space to another without paired data, with
preliminary extensions toward multimodal spaces---evidence
that the convergence is representational, not merely architectural.
Recent omni-/multi-modal embedding models such as
e5-omni~\citep{e5omni2026}, jina-embeddings-v5-omni~\citep{jinav5omni},
MM-Embed~\citep{lin2025mmembed}, and UniME~\citep{gu2025unime} are
early realizations of this vision, mapping text, image, video, and audio
into a shared space, whether by explicit cross-modal alignment or by
fusing modalities inside a single encoder.

Our central claim can thus be stated in one sentence:
\begin{quote}
\emph{Embedding research is converging along shared principles (\Cref{part:principles}),
which makes it possible to proceed from transformation and alignment to true
fusion (\Cref{part:path}); but to realize an omnimodal joint embedding we must
confront each modality's unique challenges and borrow each other's
hard-won techniques (\Cref{part:road}).}
\end{quote}

This claim, illustrated in \Cref{fig:overall-convergence-map}, organizes the paper into four parts.

\emph{\Cref{part:foundation}---The Unified Foundation} (\Cref{sec:framework}) 
establishes a modality-agnostic anatomy of embedding models (components, data
units, and the exposed representational objects), and introduces the analytical lens
(intersection, difference, reuse) used throughout this paper.

\emph{\Cref{part:principles}---Convergent Principles} (\Cref{sec:objectives,sec:arch,sec:data})
makes the case for convergence concrete: training objectives are becoming a shared
language, with the pretrain--finetune recipe assembling them into
the same two-stage workflow across modalities (\Cref{sec:objectives});
the Transformer, with emerging state-space
alternatives in contention, is becoming a shared encoder architecture
(\Cref{sec:arch}); and the representational object is shifting from parts to whole
semantic units across every modality~(\Cref{sec:data}).

\emph{\Cref{part:path}---The Path to Unification} (\Cref{sec:transformation,sec:alignment,sec:fusion})
surveys the three-step progression toward omnimodal embeddings: transforming
modalities into one another via rendering, serialization, or discretization
(\Cref{sec:transformation}), aligning independently encoded modalities
(\Cref{sec:alignment}), and ultimately fusing them through
deep interaction, in which joint embedding unifies and surpasses
alignment (\Cref{sec:fusion}).
To grade what each path actually achieves, we introduce the
\emph{3C lens}, which evaluates multimodal embedders across three
progressively harder capabilities:
\emph{coverage}, \emph{comparability}, and \emph{compositionality}.
Concretely, coverage asks which modalities the system can
encode and test; comparability asks whether separately encoded
inputs can be matched in a shared metric space (e.g., whether text can
retrieve relevant audio); and compositionality asks whether a joint
representation built from several inputs genuinely depends on all of them (e.g.,
whether an image and an edit instruction jointly define a query).
The three paths satisfy these
gates in turn: transformation broadens coverage, alignment enables
comparability, and joint embedding targets compositionality.

\emph{\Cref{part:road}---The Road Ahead} (\Cref{sec:frontier:properties,sec:frontier:pollination,sec:evaluation,sec:conclusion})
turns to what remains: the modality-specific properties a true joint embedding
must respect (\Cref{sec:frontier:properties}), a cross-pollination blueprint for
reusing one modality's methods in another (\Cref{sec:frontier:pollination}),
the unified evaluation that we cannot yet build, organized around the 3C
lens (\Cref{sec:evaluation}), and a closing outlook (\Cref{sec:conclusion}).
Per-modality cheat sheets are deferred to Appendix~\ref{app:cheatsheets}
for readers who want a quick history of a single modality without
interrupting the concept-first main thread.

To our knowledge, this paper is the first to provide a unified treatment of
embedding-model research across all major modalities from this cross-cutting,
position-driven perspective.

\section{A Unified Framework for Embedding Models}
\label{sec:framework}

Before tracing how the field is converging, we first establish a
modality-agnostic anatomy that decomposes any embedding model
into five recurring components~(\Cref{sec:framework:anatomy}),
and introduce the analytical lens of \emph{intersection}, \emph{difference},
and \emph{reuse} that guides our cross-modal
synthesis throughout the rest of the paper~(\Cref{sec:framework:lens}).

\subsection{The Anatomy of an Embedding Model}
\label{sec:framework:anatomy}

Let $\sX$ denote a data domain (e.g., the set of all English sentences, all
RGB images, all audio waveforms, all graphs, all time series).
An embedding model is a parameterized function
\begin{equation}
  f_\theta : \sX \longrightarrow \R^d
\end{equation}
that maps each input $x \in \sX$ to a $d$-dimensional vector
$\ve = f_\theta(x)$, referred to as an embedding or representation.
The goal is for the geometry of the embedding space to reflect the semantic
structure of the data:
\begin{equation}
  \simfunc(\ve_i, \ve_j) \;\approx\; \text{semantic similarity}(x_i, x_j),
\end{equation}
where $\simfunc$ is a similarity function, typically cosine similarity
$\simfunc(\vu,\vv) = \frac{\vu \cdot \vv}{\|\vu\|\,\|\vv\|}$ or negative
Euclidean distance $\operatorname{dist}(\vu,\vv) = -\|\vu - \vv\|$.
When this map is well-learned, semantically similar inputs cluster together
while dissimilar ones are pushed apart, enabling nearest-neighbor search,
clustering, classification, and cross-modal alignment through simple vector
operations.
Beyond Euclidean and cosine metrics, alternatives include geodesic
distances in hyperbolic geometries that embed hierarchies with low
distortion for knowledge taxonomies, graphs, and
images~\citep{nickel2017poincare,lin2023hyperbolic,khrulkov2020hyperbolicimage},
and dynamic time warping (DTW), the long-standing native measure for time series~\citep{dtw}.

\begin{figure*}[t]
\centering
\begin{adjustbox}{max width=\textwidth}
\begin{tikzpicture}[
  font=\sffamily,
  node distance=4.5mm,
  box/.style={
    draw=decorBlueBorder, rounded corners=4pt, align=center, minimum height=14mm,
    text width=21.5mm, inner sep=2pt, fill=decorBlueBg, font=\scriptsize\sffamily, line width=0.85pt,
    text=paperDark
  },
  decision/.style={
    draw=decorOrangeBorder, rounded corners=4pt, align=center, minimum height=14mm,
    text width=21.5mm, inner sep=2pt, fill=decorOrangeBg, font=\scriptsize\sffamily, line width=0.85pt,
    text=paperDark
  },
  arr/.style={-{Latex[length=2.5mm,width=1.6mm]}, draw=paperArrow, line width=1.1pt}
]
\node[box] (object) {{\small\bfseries Data Unit}\\[1.5pt]\textcolor{paperSubText}{raw data + metadata}};
\node[box,right=of object] (token) {{\small\bfseries Tokenizer}\\[1.5pt]\textcolor{paperSubText}{subword, patch, node}};
\node[box,right=of token] (encoder) {{\small\bfseries Encoder}\\[1.5pt]\textcolor{paperSubText}{interaction pattern}};
\node[box,right=of encoder] (pool) {{\small\bfseries Aggregation}\\[1.5pt]\textcolor{paperSubText}{\texttt{[CLS]}, mean, attn}};
\node[box,right=of pool] (reprobj) {{\small\bfseries Repr.~Object}\\[1.5pt]\textcolor{paperSubText}{vector, set, codes}};

\node[decision,right=of reprobj] (objective) {{\small\bfseries Objective}\\[1.5pt]\textcolor{paperSubText}{task + supervision}};
\node[decision,right=of objective] (eval) {{\small\bfseries Evaluation}\\[1.5pt]\textcolor{paperSubText}{claim + test}};

\draw[arr] (object)--(token);
\draw[arr] (token)--(encoder);
\draw[arr] (encoder)--(pool);
\draw[arr] (pool)--(reprobj);
\draw[arr] (reprobj)--(objective);
\draw[arr] (objective)--(eval);

\coordinate (feedback-right) at ([yshift=-8mm]eval.south);
\coordinate (feedback-left) at ([yshift=-8mm]object.south);
\draw[arr, rounded corners=5pt] (eval.south) -- (feedback-right) --
  node[below=1.5pt, font=\scriptsize\bfseries\sffamily, text=paperSubText]{the intended tests determine what the embedding must retain}
  (feedback-left) -- (object.south);
\end{tikzpicture}
\end{adjustbox}
\caption{The anatomy of an embedding model. The intended use should determine what is embedded, how inputs interact, how outputs are pooled, and which information must be retained. Structural components (data unit, tokenizer, encoder, aggregation, representational object) are distinguished from the modeling decisions (objective, evaluation).}
\label{fig:anatomy}
\end{figure*}
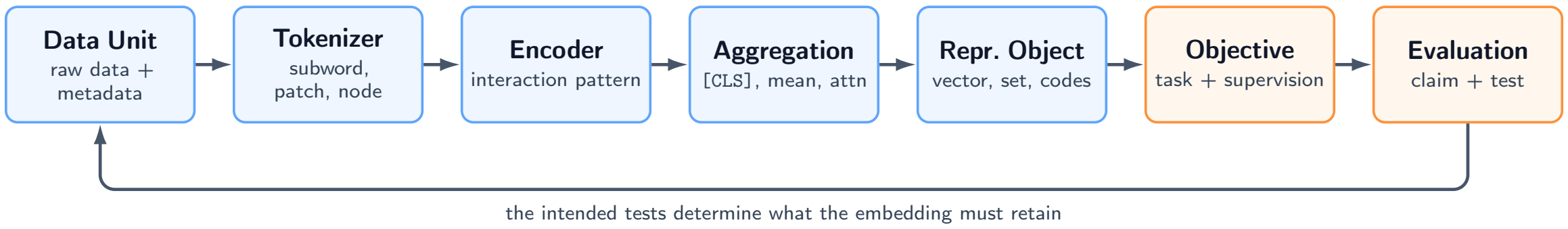

Whatever the modality, almost every modern embedding model can be
decomposed into five components.
The components are defined by their role, not by any particular
implementation, which is what makes the decomposition modality-agnostic.
\Cref{tab:anatomy} summarizes them, and \Cref{fig:anatomy} shows the role each
part plays in the life cycle of an embedding model.

\begin{table*}[t]
\centering
\small
\caption{The anatomy of an embedding model. Each row is a model component or
data object; each column instantiates that row in a
given modality. The same skeleton recurs across modalities, and
the substance that varies is the implementation of each role.}
\label{tab:anatomy}
\adjustbox{width=\textwidth,center}{
\rowcolors{2}{gray!10}{white}
\begin{tabular}{@{}l m{2.35cm} m{2.35cm} m{2.35cm} m{2.35cm} m{2.35cm} m{2.35cm}@{}}
\toprule
\textbf{Component} & \textbf{Text / Code} & \textbf{Image / Video} & \textbf{Audio} & \textbf{Time Series} & \textbf{Graph} & \textbf{Table} \\
\midrule
\textbf{Data unit \& tokenization}
  & character, subword, word
  & pixel, patch, region, frame
  & waveform, frame, spectrogram patch
  & timestamp, window, patch, channel
  & node, edge, subgraph, triple
  & cell, header, row, column \\
\textbf{Encoder}
  & Transformer; decoder-only LLM
  & CNN, ViT, video Transformer
  & CNN, Conformer, Transformer, SSM
  & RNN, TCN, SSM, Transformer, MLP
  & GNN, Graph Transformer, KG scorer
  & row--column Transformer, feature tokenizer \\
\textbf{Pooling / aggregation}
  & \texttt{[CLS]}, mean, last-token
  & \texttt{[CLS]}, dense patch set
  & statistics pooling, layer-weighted
  & mean, patch pooling
  & node readout, readout over graph
  & cell/row/column readout \\
\textbf{Representational object}
  & global vector, token sequence
  & global vector, patch set, discrete codes
  & clip vector, frame sequence, codec codes
  & magnitude-preserving vector, patch sequence
  & node/graph vector, distribution
  & cell/row/column vectors, table vector \\
\textbf{Training objective}
  & contrastive, MLM, next-token prediction
  & contrastive, reconstruction, distillation
  & masked prediction, speaker/content contrast
  & forecasting, reconstruction, contrastive
  & proximity, link/triple reconstruction, contrastive
  & cell corruption, entity recovery, table--text contrast \\
\textbf{Evaluation paradigm}
  & frozen-encoder retrieval/STS/classification
  & frozen-encoder retrieval/classification/detection
  & frozen-encoder recognition/retrieval
  & forecasting, anomaly, classification
  & node classification, link prediction
  & table QA, fact verification, text-to-SQL \\
\bottomrule
\end{tabular}
}
\end{table*}

\begin{itemize}[leftmargin=2em]
\item\textbf{(1) Data unit and tokenization.}
Raw data must be partitioned into units (e.g., subwords, patches, frames,
nodes, or cells) and mapped to tokens before encoding.
Here ``token'' denotes a unit's role in a sequence rather than membership in a
finite discrete vocabulary~(\Cref{sec:data:units}).
\item\textbf{(2) Encoder architecture.}
The encoder maps input tokens into contextualized hidden states.
While modality-specific backbones hard-code strong domain
symmetries (e.g., translation equivariance of CNNs, permutation equivariance
of GNNs, and temporal recurrence of RNNs;~\citealp{bronstein2021geometric}),
the field is rapidly converging on the
Transformer as the default backbone~(\Cref{sec:arch}).
\item\textbf{(3) Aggregation and representational object.}
Downstream consumers read out a specific representational object from the
hidden states $(\vh_1, \ldots, \vh_L)$ via pooling (e.g., \texttt{[CLS]},
mean pooling, attention pooling, or statistics pooling).
The choice of representational object trades off computational efficiency
against fine-grained structural correspondence~(\Cref{sec:data}).
\item\textbf{(4) Training objective.}
The loss function shapes the latent geometry.
Four families recur across modalities: contrastive objectives pull
positives together and push negatives apart; reconstructive
objectives mask part of the input and predict it back; generative
objectives train a decoder and reuse
its latent; and hybrid objectives jointly optimize for
representation and generation~(\Cref{sec:objectives}).
\item\textbf{(5) Evaluation paradigm.}
Embedding quality is measured by freezing the encoder and probing it on a
diverse suite of downstream tasks.
However, current benchmarks remain siloed by modality,
with each modality $X$ developing its
own M$X$EB (MTEB for text, MIEB for image, MAEB for audio)---a
limitation we confront in \Cref{sec:evaluation}.
\end{itemize}

The point of this anatomy is that the five components are orthogonal
axes: one can swap the encoder while holding the objective fixed, or change
the representational object while holding the encoder fixed.
Cross-modal convergence, the theme of \Cref{part:principles}, is largely the observation
that on four of these five axes the same choices are increasingly being made
in every modality.

\subsection{Our Analytical Lens: Intersection, Difference, and Reuse}
\label{sec:framework:lens}

To reason across modalities without collapsing their distinctions, we adopt
a three-part lens, applied uniformly in every subsequent chapter.
\begin{itemize}[leftmargin=2em]
  \item \textbf{Intersection}---the shared core: the mathematical principles,
    architectures, and paradigms that recur across modalities.
    The InfoNCE loss, the Transformer, the pretrain--finetune recipe, and
    the frozen-encoder evaluation protocol are intersections. \Cref{part:principles} is,
    in effect, a catalog of intersections.
  \item \textbf{Difference}---the modality-specific core: the challenges,
    data properties, and methods that belong to one modality because they
    derive from the structure of its inputs.
    Audio's superposition of factors, graph's explicit relational
    structure, time series' continuous axis and non-standardized units
    are differences. \Cref{sec:frontier:properties} enumerates them.
  \item \textbf{Reuse}---the directional transfer: how a methodology born
    in one modality is borrowed and adapted by another.
    Contrastive learning moving from vision to text to audio represents a reuse.
    The cross-pollination map of \Cref{sec:frontier:pollination} is a
    deliberate exercise in planning reuse.
\end{itemize}
These three categories are not a partition---a single finding often sits on
the boundary. Audio's superposition of factors is a difference, but
the disentanglement methods it provoked are a candidate reuse target
for other modalities; the Transformer is an intersection, yet the
question of which positional encoding to inject is a difference.
The lens is what keeps the comparison honest: it lets us state, for each
technique, not merely that it exists but which modality owns the
problem and which is borrowing the solution.


\clearpage
\part{Convergent Principles}
\label{part:principles}

\begin{wrapfigure}{r}{0.57\textwidth}
\centering
\vspace{-11pt}
\includegraphics[width=0.57\textwidth]{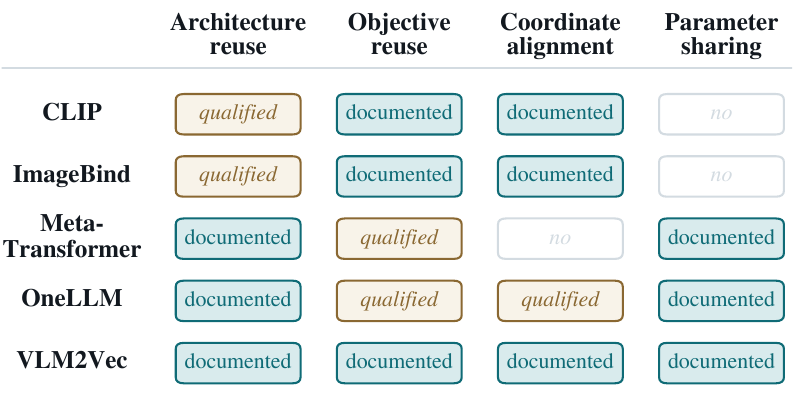}
\caption{Four meanings of convergence shown separately for illustrative
systems. ``Qualified''
marks partial architectural reuse or interfaces whose status depends on the
component examined.}
\label{fig:convergence}
\vspace{-7pt}
\end{wrapfigure}

Building on the anatomy of \Cref{sec:framework}, \Cref{part:principles} traces
the evolution across training objectives, encoder architectures,
and representational granularity.
Because ``convergence'' spans several independent design choices,
we separate four distinct meanings in \Cref{fig:convergence}: architecture reuse,
objective reuse, coordinate alignment, and parameter sharing. A shared trunk
does not entail aligned retrieval vectors (as in
Meta-Transformer;~\citealp{metatransformer2023}), nor vice versa (as in
ImageBind;~\citealp{imagebind}), while systems such as
OneLLM~\citep{han2024onellm} and VLM2Vec~\citep{jiang2025vlm2vec} combine both.

\section{Training Objectives: The Lingua Franca}
\label{sec:objectives}

Across modalities, a small set of losses---contrastive, reconstructive,
and increasingly generative--hybrid---now account for the overwhelming majority
of state-of-the-art embedding models.
Viewed through our analytical lens, the mathematical form of each loss
is a cross-modal intersection, whereas the construction of prediction
targets---what is pulled together, what is masked, what is reconstructed---is
a difference that reflects modality structures.
The convergence, moreover, is not only in the losses but in the workflow that
stages them: pretrain first, finetune second is now the common recipe in
every modality.

\subsection{Contrastive Objectives}
\label{sec:objectives:contrastive}

The contrastive paradigm is the dominant force in modern embedding research.
Its core is simple: given an anchor $x$, attract a positive $x^+$
(a semantically equivalent or related sample) and repel negatives
$x^-$ (unrelated samples) in the embedding space.

\paragraph{The shared mechanism.}
The most widely used instantiation is the InfoNCE
loss~\citep{vandenoord2018cpc}:
\begin{equation}
  \mathcal{L}_{\text{InfoNCE}}
  = -\log \frac{\exp\!\bigl(\simfunc(\ve, \ve^+)/\tau\bigr)}
                {\sum_{j=1}^{N} \exp\!\bigl(\simfunc(\ve, \ve_j)/\tau\bigr)},
  \label{eq:infonce}
\end{equation}
where $\ve = f_\theta(x)$, $\ve^+ = f_\theta(x^+)$, $\tau$ is a temperature, and the
denominator sums one positive with $N{-}1$ negatives (often in-batch).
InfoNCE is a categorical cross-entropy in which the positive is the correct
``class'' among $N$ candidates; under standard positive- and negative-sampling assumptions, $\log N - \mathcal{L}_{\text{InfoNCE}}$ lower-bounds the mutual information between views of the input~\citep{vandenoord2018cpc, wu2018infonce}.
Its normalized variant, NT-Xent~\citep{chen2020simclr}, projects embeddings to
the unit hypersphere before scoring to encourage uniformity, while the earlier
triplet loss~\citep{schroff2015facenet} directly enforces a margin between the
positive and negative distances. These forms are modality-agnostic: the loss
does not know whether the input comes from an image, a sentence, or a graph.
What differs across modalities is entirely in how
the positive $x^+$ and negatives $x^-$ are constructed, which is where each
modality's structure enters.

\begin{table*}[t]
\centering
\small
\caption{The contrastive objective is shared across modalities, but the choice
of positive pair defines what the embedding should treat as unchanged---the
declared invariance. The mechanism (InfoNCE) is an intersection; the positive
construction and its characteristic failure are the difference.}
\label{tab:positive-relations}
\adjustbox{width=\textwidth,center}{
\rowcolors{2}{gray!10}{white}
\begin{tabular}{@{}m{0.12\textwidth}m{0.24\textwidth}m{0.26\textwidth}m{0.30\textwidth}@{}}
\toprule
\textbf{Modality} & \textbf{Typical positive} & \textbf{Intended invariant} & \textbf{Characteristic failure} \\
\midrule
Text & Dropout views, paraphrase, query--relevant passage & Surface variation under semantic equivalence & False negatives and lexical shortcuts~\citep{gao2021simcse,xiong2021ance} \\
Vision & Crops or photometric views of one image & Viewpoint and appearance change & Cropping removes the discriminating object~\citep{he2020moco} \\
Audio & Two utterance views or an audio--caption pair & Content, speaker, event, or scene, depending on task & A desired factor for one task is a nuisance for another~\citep{clap} \\
Graph & Augmented views, nearby nodes, valid triples & Topological proximity or relation compatibility & Node/edge dropout changes graph semantics~\citep{you2020graphcl,wei2023boosting} \\
Time series & Windows, scales, or time--frequency views & Shape, regime, or instance identity & Jitter, scaling and warping are domain-dependent~\citep{ts2vec} \\
Cross-modal & Co-observed or captioned objects & Shared semantic content & Co-occurrence is partial or asynchronous~\citep{radford2021clip,imagebind} \\
\bottomrule
\end{tabular}
}
\end{table*}

\paragraph{Difference: how each modality builds its pairs.}
Contrastive learning is a machine for encoding a chosen
invariance~\citep{chen2020simclr}, and the augmentation set declares what
transformations should leave the semantics unchanged.
The modalities differ in how easy that declaration is.
\begin{itemize}[leftmargin=2em]
  \item \textbf{Text---from discrete edits to implicit augmentation.}
    Text has no natural geometric augmentation, so its positives were initially
    manufactured by discrete edits: EDA composes synonym replacement, random
    insertion, swap, and deletion~\citep{wei2019eda}; AEDA reduces this to
    random insertion of punctuation marks~\citep{karimi2021aeda}; UDA builds a
    consistency objective out of such augmented views~\citep{xie2019uda}; and
    TMix interpolates the hidden states of two sentences, moving the
    augmentation into the encoder's latent space~\citep{chen2020mixtext}.
    SimCSE~\citep{gao2021simcse} then made the augmentation entirely
    implicit: the same sentence passes the encoder twice, and the
    stochasticity of independent dropout masks in the encoder produces two slightly
    different embeddings of the same input---an elegant, data-free positive
    sampling method.
  \item \textbf{Vision---the paradigm's birthplace.} Contrastive learning
    originated and matured in vision, spanning from the Siamese network of signature
    verification~\citep{bromley1993siamese} to the modern instantiations.
    Random crops, color jitter, and flips preserve the semantic identity of an
    object~\citep{chen2020simclr}, with MoCo's momentum encoder maintaining a
    large, consistent negative queue~\citep{he2020moco}.
    The family's maturity has a landmark: self-supervised pretraining in
    vision now matches or beats supervised training on standard
    benchmarks~\citep{caron2021dino, he2022mae}---the outcome that made this
    recipe the template the other modalities would follow.
  \item \textbf{Vision--language---naturally paired data.} Where single
    modalities must manufacture positives, the web naturally provides
    abundant paired data for vision--language: a caption and its image are a positive by
    construction~\citep{radford2021clip}.
  \item \textbf{Audio---from waveform corruption to
    quantized targets and language pairing.}
    Analogous to vision's crop-and-jitter family, 
    audio manufactured positives by waveform-level
    augmentations: pitch shifting and time stretching, additive noise
    injected at a controlled SNR, and reverberation via convolution with
    measured or simulated room impulse responses~\citep{ko2015audioaug,
    salamon2017soundaug, allenberg1979image}.
    The newer constructions are less hand-crafted: wav2vec 2.0 masks latent
    frames and contrasts them against quantized targets derived from the
    same audio~\citep{baevski2020wav2vec2};
    CLAP ports the CLIP recipe to (audio, caption) pairs~\citep{clap},
    with successors manufacturing the semantic spread that closed-vocabulary
    tag captions lack~\citep{laionclap, glap, humanclap}.
  \item \textbf{Code---views as positives.}
    Code can have many different forms that describe the same behavior.
    Its positive includes a semantically equivalent view
    of the same program: its docstring, abstract syntax tree (AST),
    data-flow graph, or a semantics-preserving
    refactor~\citep{syncobert2021, coderetriever2022, codemvp2022,
    unixcoder2022}, which is what turns contrastive learning into a probe of
    executable semantics rather than surface form.
  \item \textbf{Graph---view construction over structure.} Graph contrastive
    methods define positive and negative graphs through structural augmentations
    such as node/edge dropout and feature masking; the contrastive and
    generative families are distinguished by the operation that creates the
    learning signal~\citep{you2020graphcl,wei2023boosting,jin2022empowering,
    chen2024tokenized_graph_transformer_contrastive}.
  \item \textbf{Time series---the augmentation problem.} Here the construction
    is hardest: jittering, scaling, permutation, and warping each encode an
    invariance that holds in some domains and fails in others---augmentations
    that are beneficial elsewhere can actively hurt here~\citep{ts2vec},
    which is why TF-C pools a bank of augmentations rather than trusting any
    single transform~\citep{tfc}.
    Time series confronts most starkly the truth that an augmentation is a
    declared invariance; the principled responses either declare the invariance
    architecturally~\citep{noether, hardcoded} or contrast across time-frequency
    views~\citep{tfc, trep} and multi-rate sampling hierarchies~\citep{mfclr}.
\end{itemize}

The recurring insight is that the InfoNCE mechanism remains constant, while each modality supplies its own structural prior to declare which transformations are meaning-preserving. Looking across the list above, positive-pair construction divides along a fundamental axis: geometric and signal perturbations on continuous grids versus structural rewrites or implicit views on discrete symbols. Continuous-grid modalities (vision, audio) naturally adopt spatial, photometric, or waveform perturbations, though such jittering hits a sharp boundary in time series due to heterogeneous domain semantics. Conversely, discrete and structured modalities (text, code, graphs) lack continuous coordinate grids and must rely on independent dropouts, syntax-tree/data-flow views, or topological masking.
Table~\ref{tab:positive-relations} makes this mechanism-versus-difference split explicit: the same equation means different things in each modality because the positive pair declares a different invariance, and the recurring failure is the augmentation that abuses it.

\paragraph{Hard negatives and the alignment--uniformity decomposition.}
Two further shared refinements recur across modalities.
Random in-batch negatives are often too easy; hard negative
mining---selecting negatives close to the anchor but semantically distinct---is
critical for dense retrieval~\citep{karpukhin2020dpr, xiong2021ance} and reappears
as batch construction in modern LLM embedders~\citep{lee2025nvembed,f2llmv2}.
And for the standard normalized contrastive loss, the asymptotic analysis
of \citet{wang2020alignment} motivates two measurable quantities---the
expected closeness of positives (\emph{alignment}) and the spread of normalized
embeddings on the hypersphere (\emph{uniformity})---which give a
modality-agnostic design lens and diagnostic used from
text~\citep{gao2021simcse} to time series~\citep{timehut}.

\paragraph{Negative-free contrastive learning.}
The same decomposition suggests that repulsion admits a substitute: what
negatives prevent is collapse, and the negative-free branches of the vision
lineage show two ways to prevent it without negatives.
The first is architectural bootstrapping: BYOL learns by predicting an
exponential-moving-average copy of itself~\citep{grill2020byol}, and SimSiam
reduces this to a plain Siamese pair whose predictor-and-stop-gradient
asymmetry alone prevents collapse~\citep{chen2021simsiam}.
The second is redundancy reduction: the learning signal lives in the
statistics of the embedding batch rather than in pairwise comparisons.
SwAV enforces consistency of cluster assignments across views,
spreading the batch against learned prototypes~\citep{caron2020swav};
Barlow Twins drives the cross-correlation matrix of the two views' embeddings
toward the identity~\citep{zbontar2021barlow}, so each embedding dimension is
stable across views yet non-redundant with the others;
VICReg states the same goal as explicit variance, invariance, and covariance
terms~\citep{bardes2022vicreg}, with the variance floor alone preventing
collapse.
Both routes keep the objective negative-free while still filling the
uniformity half of the decomposition: the former places the anti-collapse
pressure in an architectural asymmetry, the latter in the batch's own moments.

\subsection{Reconstructive Objectives}
\label{sec:objectives:reconstructive}

While contrastive learning operates on relative relationships,
reconstructive objectives learn representations by recovering masked or
corrupted portions of the input.
The shared mechanism is ``mask and predict''; the difference is \emph{what is
masked and what constitutes a correct prediction}.

\paragraph{The shared mechanism.}
A representative reconstructive loss masks a set $\sM$ of positions and trains
the model to predict the masked content from the remainder $x_{\setminus \sM}$:
\begin{equation}
  \mathcal{L}_{\text{rec}}
  = \E_{\sM}\!\left[\sum_{i \in \sM} \ell(\hat{x}_i, x_i)\right],
\end{equation}
where $\ell$ is typically cross-entropy for discrete targets and MSE for continuous targets. The encoder, forced to summarize enough of the input to fill in the gaps,
learns rich contextual representations that are then finetuned with
contrastive or supervised objectives for downstream embedding tasks.

\paragraph{Difference: what is masked and predicted.}
\begin{itemize}[leftmargin=2em]
  \item \textbf{Text---discrete token IDs.} BERT~\citep{devlin2019bert}
    masks tokens and predicts a discrete token ID from a finite
    vocabulary---a semantically loaded discrete symbol.
  \item \textbf{Code---structure-aware
    targets.} Code inherited MLM from BERT but quickly made the target richer than token
    IDs: CodeBERT~\citep{codebert2020} adds replaced-token detection over
    (NL, code) pairs, GraphCodeBERT~\citep{graphcodebert2021} predicts
    data-flow edges, and CodeT5~\citep{codet5_wang2021} masks identifiers so
    the model must infer what an obfuscated variable does. The masked target
    thus encodes ever deeper structure: surface token $\to$ syntax $\to$
    semantics.
  \item \textbf{Vision---high-ratio continuous pixels.} MAE~\citep{he2022mae}
    masks a large fraction (75\%) of image patches and reconstructs the raw
    continuous pixels; the asymmetry between deep encoder on visible patches
    and shallow decoder for all forces the encoder to capture semantics.
  \item \textbf{Audio---cluster IDs, spectrogram patches, and learned targets.}
    HuBERT~\citep{hsu2021hubert} masks latent features and predicts offline
    cluster assignments, refined across rounds; AudioMAE~\citep{audiomae} masks
    spectrogram patches and regresses toward the raw spectrogram; data2vec
    instead regresses toward contextualized representations built by an
    exponential-moving-average teacher~\citep{data2vec}; and BEATs predicts
    discrete labels generated by a learned acoustic tokenizer~\citep{beats}.
  \item \textbf{Graph---hidden structure and attributes.} Graph reconstructive
    methods reconstruct hidden structure or attributes: masked edges~\citep{li2023s}, or node features~\citep{hou2022graphmae} from partially observed graphs. KG completion methods score entity--relation triples against corrupted candidates~\citep{bordes2013transe, sun2019rotate}.
  \item \textbf{Table---cells, entities, and structure.} Among table encoders,
    TURL recovers the entity behind a masked cell~\citep{turl2020},
    TABBIE detects a
    corrupted cell among intact ones~\citep{tabbie2021}, and table-based
    QA models such as TaPas/TaBERT recover table-conditioned
    text~\citep{tapas2020, tabert2020}.
    The masked target is thus neither a token ID nor a pixel but
    typed cell content---a unit whose meaning comes jointly from its
    value, its header, and its row--column position.
  \item \textbf{Time series---the interpolation pitfall.} Masked patch
    reconstruction is the continuous analogue of MLM, but where the language
    version predicts a semantically loaded symbol, the continuous version
    regresses toward a conditional mean that is
    dominated by interpolation for a smooth signal,
    recovering trend and seasonality that can be extracted by classical
    decomposition without learning~\citep{nie2023times, tsfmintervention}.
\end{itemize}

The contrast is revealing: reconstructive pretraining succeeds when the masked target resists trivial local interpolation, either naturally via discrete semantic units or artificially via extreme masking ratios.
In symbolic and structured domains (text, code, graphs, tables), reconstruction targets semantic or topological integrity directly by masking tokens, identifiers, edges, or typed cells---a paradigm that audio similarly joins via discretization~(explained in detail in \Cref{sec:transformation:tokens}).
By contrast, reconstructing raw continuous signals confronts an interpolation pitfall: because adjacent samples are smoothly correlated, point-wise regression can easily collapse into trivial local interpolation rather than high-level semantic abstraction. Vision and spectrogram-based audio overcome this through aggressive spatial masking, while in time series the effects of reconstruction remain limited.

\subsection{Generative and Hybrid Objectives}
\label{sec:objectives:generative}

A third family learns representations as a byproduct of \emph{generating} the
data, and a recent, increasingly important trend unifies generative and
representational objectives.

\paragraph{Generative objectives for representation.}
The ``generative'' label fixes what must be predicted during training,
whereas the form of the exposed representation remains a separate design
choice---just as MAE trains under a generative pretext loss yet yields a
transferable continuous encoder~\citep{he2022mae}, generative models can
expose latents that are distributional, continuous, or discrete.
Variational autoencoders~\citep{kingma2014vae} expose a distributional latent
$p(\vz \mid x)$ that captures multiplicity and uncertainty, which is useful
when an input corresponds to several semantic readings, as an image matches
many captions~\citep{chun2021pcme}.
GAN-based encoders~\citep{donahue2017bigan, song2018bgan} learn a
data-to-latent mapping that inverts the generator, and its output doubles as
an embedding.
Diffusion models contribute latents too: latent diffusion generates in a
compressed autoencoder space~\citep{rombach2022ldm}, diffusion autoencoders
factor a semantic code from stochastic detail~\citep{preechakul2022diffae},
and DIFT shows that intermediate diffusion features yield strong zero-shot
correspondence~\citep{tang2023dift}.
A parallel line vector-quantizes continuous features into discrete
codebook indices for reconstruction, represented by VQ-VAE in
vision~\citep{vandenOord2017vqvae}, and SoundStream \& EnCodec
in audio~\citep{soundstream, defossez2022encodec}.

\paragraph{Hybrid: the unification of representation and generation.}
This unification began in the text world when UniLM jointly optimized bidirectional,
unidirectional, and sequence-to-sequence language modeling objectives within a
single Transformer via varying attention masks~\citep{dong2019unilm}.
Shortly after, T5 unified representation and generation tasks by casting them
both into a sequence-to-sequence modeling format~\citep{raffel2020t5},
and Unified-IO extended it into the multimodal setting, committing text, images,
and structured outputs to one shared discrete-token
vocabulary~\citep{lu2022unifiedio, lu2024unifiedio2}.
In the LLM era, GRIT~\citep{muennighoff2024grit} represents an early signal of
convergence between representation and generation, jointly training a single LLM
with a contrastive loss for embedding and an autoregressive loss for
generation, using instruction prefixes to disambiguate the tasks.
The unification is spreading: latent diffusion transfers self-supervised
geometry into diffusion-transformer states~\citep{yu2025repa}, MergeVQ jointly
optimizes representation and generation with token merging and
quantization~\citep{li2025mergevq}, early-fusion models interleave text and
visual tokens in a single autoregressive stream~\citep{chameleon2024},
and VladVA demonstrates that an autoregressive loss on long captions helps
a vision-language embedder capture fine-grained details~\citep{ouali2024vladva}.
Audio--language models add a scaling law of their own: continual generative
pretraining before contrastive training raises the ceiling on representation
quality~\citep{lcoembedding}.
The unification also runs in reverse, where pretrained encoders extend LLMs
rather than fuse into them: E2LLM feeds an LLM soft prompts
compressed from long contexts by a text
encoder~\citep{liao2025e2llm}, and GALLa aligns a GNN's structural embeddings
into a code LLM as an auxiliary training objective~\citep{zhang2025galla}.
What is emerging is a view in which representation and generation are not
opposed but two heads of the same objective---a unification that the joint
embedding models of \Cref{sec:fusion} will push to its logical conclusion.

\paragraph{Knowledge distillation.}
A practical hybrid across all modalities is distillation, in which a compact
student mimics a powerful teacher's embedding space.
Self-distillation underlies DINO in vision, where the teacher is an
exponential moving average of the student~\citep{caron2021dino}, and
increasingly text, where BGE-M3 distills from a teacher signal built as the
weighted integration of its dense, sparse, and multi-vector relevance
scores~\citep{chen2024m3embed}; cross-encoder-to-bi-encoder distillation is
standard in training dense retrievers~\citep{qu2021rocketqav2}, and extends
into the LLM regime~\citep{liao2024d2llm}.
Where a strong teacher exists, distillation is the most economic route to a
modality-agnostic embedding, and it is the mechanism behind many modern
multimodal embedders~\citep{gu2025unime}.

\subsection{The Pretrain--Finetune Recipe}
\label{sec:objectives:pretrain}

The three loss families above, for all their differences, are rarely applied
standalone: across modalities they are assembled into a common pipeline.
The dominant recipe is
to pretrain on large-scale unlabeled or weakly-supervised data, then finetune
on task-specific or contrastive data.
This recipe is integral to the discussion
of training objectives, and like them it is an intersection,
whose differences lie only in how the stages are filled.

\paragraph{Intersection: one recipe, every modality.}
The two-stage framework recurs everywhere.
In text, BERT pretrains with MLM and then finetunes on downstream
tasks~\citep{devlin2019bert}; SimCSE shows the second stage can itself be
contrastive and unlabeled, adapting a pretrained encoder with
dropout-induced positives~\citep{gao2021simcse}; and E5 shows the first stage
can be contrastive too, mining weakly-supervised pairs from the web and
relying on supervised finetuning for the strongest
results~\citep{wang2022e5}.
In vision, MAE and DINO pretrain by reconstruction and self-distillation,
respectively, and transfer by finetuning or linear
probing~\citep{caron2021dino, he2022mae}.
In audio, wav2vec 2.0 pretrains on unlabeled speech and then finetunes on as
little as ten minutes of transcribed speech~\citep{baevski2020wav2vec2}.
In graphs, pretraining must cover both grains of the data at once: combining
node-level and graph-level pretraining avoids
the negative transfer that either alone can cause~\citep{hu2020gnnpretrain,kong2025gofa}.
In tables, TaBERT and TaPas pretrain on tables and accompanying text and
then finetune on downstream tasks such as table question
answering~\citep{tabert2020, tapas2020}.
The recipe even survives the shift from encoders to LLM embedders, where
stage one is typically generative pretraining and stage two contrastive,
instruction-conditioned finetuning~\citep{su2022instructor,
wang2024e5mistral, lee2025nvembed}; audio--language embedders document the
same swap of objectives into the same slots, with a generative stage one
raising the ceiling of the contrastive stage two
(\Cref{sec:objectives:generative}).
What persists across all of these is not any particular loss but the
staging: accessible universal signal first, expensive task-specific signal second.

\paragraph{Difference: what fills each stage.}
What the modality determines is what counts as pretraining material---raw
text, unlabeled speech, web image--caption pairs, contrastively mined
sentence pairs---and what the finetuning signal is: labels, pairs, or
instructions.
The boundary cases show the recipe is a considered choice rather than a law.
CLIP-style models collapse the pipeline: where the web-scale paired data
is readily available, the contrastive stage becomes the pretraining, and zero-shot
transfer replaces finetuning~\citep{radford2021clip}.
And the graph case above warns that the pipeline can even hurt when its first
stage is mismatched to the modality's structure, restating this section's
mechanism-versus-difference split at the level of the workflow.

\highlightbox{tred}{Takeaway}{%
The core takeaway of this section is convergence-with-difference.
The InfoNCE loss, the mask-and-predict recipe, the representation--generation
unification, and the pretrain--finetune workflow are intersections---the same
machinery running in every modality.
What differs is the raw material each modality gives that machinery: a
dropout-induced positive for text, a cropped view for vision, a quantized
target for audio, a structural augmentation for graphs, and a declared
symmetry for time series.
The objective is the lingua franca; the data unit decides what can be said in
it.%
}

\section{Architectures: The Universal Encoder}
\label{sec:arch}

The second axis of convergence is architectural: the
Transformer~\citep{vaswani2017attention} is becoming the default encoder across
modalities---ViT for images~\citep{dosovitskiy2021vit}, audio Transformers for
speech~\citep{baevski2020wav2vec2}, Graph Transformers for structured
data~\citep{geisler2023transformers}, patch Transformers for time
series~\citep{nie2023times}, and decoder-only LLMs for
text~\citep{wang2024e5mistral, lee2025nvembed}.
This section explains why a single architecture can serve as a universal
backbone across all modalities~(\Cref{sec:arch:transformer}), and, equally important,
when a modality-specific architecture still remains justified~(\Cref{sec:arch:specific}).
We also compare Transformer's most prominent challengers---state space models and
linear-attention variants---in the context of embedding models~(\Cref{sec:arch:ssm}),
and step beyond the trunk to examine the system-level interaction architectures and
pooling mechanisms that turn hidden states into usable embeddings~(\Cref{sec:arch:pooling}).

\subsection{The Transformer as a Universal Backbone}
\label{sec:arch:transformer}

We argue the Transformer converges as a universal encoder because four
interlocking properties make it the most flexible architecture that remains
practically trainable at scale.

\paragraph{Weak inductive bias: self-attention as a set operator.}
The Geometric Deep Learning framework~\citep{bronstein2021geometric} unifies
architectures by the symmetry assumptions they encode: CNNs assume translation
equivariance and locality; RNNs assume temporal equivariance and Markovian
structure; GNNs assume permutation equivariance over a prescribed graph
topology.
The Transformer's core mechanism assumes none of these: self-attention is a
\emph{permutation-equivariant set operation}.
Each token issues a query, compares it against all keys, and receives a
weighted sum of all values; the operation treats its input as an unordered
token set in which every element may interact with every other, and it presumes
no particular spatial or temporal structure---neither the locality of
CNNs nor the sequentiality of RNNs.
Concretely, absent positional encodings and causal masks, permuting the input
tokens permutes the token-level outputs in step~\citep{yun2019universal}, and
a pooled readout such as \texttt{[CLS]} or mean pooling then aggregates the
equivariant token outputs into a permutation-invariant embedding.
From the GNN perspective, self-attention is message-passing on a
complete graph with learned edge weights, rather than a topology
prescribed by domain structure~\citep{bronstein2021geometric,
mijangos2025relational}.
By the criteria of \citet{battaglia2018relational}, this all-to-all
interaction carries a correspondingly weak relational inductive bias: it
makes no assumption about which elements should interact, only that
any interaction is possible.
The trade-off is a data tax: strong priors enable fast convergence in
low-data regimes but constrain the solution space, whereas the Transformer's
weak bias requires more data but can learn representations less constrained
by hand-designed priors~\citep{dosovitskiy2021vit}.
Empirically, under matched training, Transformers exhibit stronger
out-of-distribution generalization than CNNs~\citep{bai2021robust}, and ViT
representations are markedly more uniform across layers than CNNs, because
self-attention enables early global information
aggregation~\citep{raghu2021vitlike}.
And the mechanism is expressive enough to be a universal approximator of
continuous permutation-equivariant sequence-to-sequence
functions~\citep{yun2019universal}, with even two-layer attention-only
architectures sufficing~\citep{universalapproxattention2025}, while
cross-attention extends it to a universal bipartite relational operator for
multimodal fusion (\Cref{sec:fusion}).
This is why it is universal: any modality that can be tokenized into
a set can be processed by a Transformer.

\begin{figure*}[t]
\centering
\definecolor{modText}{HTML}{3B82F6}   \definecolor{modTextBg}{HTML}{DBEAFE}
\definecolor{modVision}{HTML}{DB2777} \definecolor{modVisionBg}{HTML}{FCE7F3}
\definecolor{modAudio}{HTML}{7C3AED}  \definecolor{modAudioBg}{HTML}{EDE9FE}
\definecolor{modGraph}{HTML}{EA580C}  \definecolor{modGraphBg}{HTML}{FFEDD5}
\definecolor{modTS}{HTML}{4B5563}     \definecolor{modTSBg}{HTML}{E5E7EB}
\begin{adjustbox}{max width=\textwidth}
\begin{tikzpicture}[
  font=\sffamily,
  node distance=3.5mm and 8mm,
  input/.style={
    draw=paperBorder, rounded corners=4pt, align=center, text width=22mm,
    minimum height=8mm, fill=bgNeutral, font=\scriptsize\bfseries\sffamily, line width=0.75pt, text=paperDark
  },
  token/.style={
    draw=decorBlueBorder, rounded corners=4pt, align=center, text width=26mm,
    minimum height=8mm, fill=decorBlueBg, font=\scriptsize\sffamily, line width=0.75pt, text=paperDark
  },
  core/.style={
    draw=coreBorder, line width=1.3pt, rounded corners=6pt, align=center, text width=38mm,
    minimum height=20mm, fill=coreBg, font=\small\sffamily, text=coreText
  },
  outbox/.style={
    draw=decorOrangeBorder, rounded corners=4pt, align=center, text width=33mm,
    minimum height=8.5mm, fill=decorOrangeBg, font=\scriptsize\sffamily, line width=0.75pt, text=paperDark
  },
  arr/.style={-{Latex[length=2.2mm,width=1.5mm]}, draw=paperArrow, line width=1pt},
  inputText/.style={input, draw=modText, fill=modTextBg, text=modText},
  inputVision/.style={input, draw=modVision, fill=modVisionBg, text=modVision},
  inputAudio/.style={input, draw=modAudio, fill=modAudioBg, text=modAudio},
  inputGraph/.style={input, draw=modGraph, fill=modGraphBg, text=modGraph},
  inputTS/.style={input, draw=modTS, fill=modTSBg, text=modTS},
  tokenText/.style={token, draw=modText!55, fill=modTextBg},
  tokenVision/.style={token, draw=modVision!55, fill=modVisionBg},
  tokenAudio/.style={token, draw=modAudio!55, fill=modAudioBg},
  tokenGraph/.style={token, draw=modGraph!55, fill=modGraphBg},
  tokenTS/.style={token, draw=modTS!55, fill=modTSBg},
]
\node[inputText] (text) {Text / Code};
\node[inputVision,below=of text] (vision) {Image / Video};
\node[inputAudio,below=of vision] (audio) {Audio};
\node[inputGraph,below=of audio] (graph) {Graph};
\node[inputTS,below=of graph] (ts) {Time Series};

\node[tokenText,right=of text] (tt) {subwords + order};
\node[tokenVision,right=of vision] (tv) {patches/tubelets + $t,x,y$};
\node[tokenAudio,right=of audio] (ta) {frames/codes + rate};
\node[tokenGraph,right=of graph] (tg) {nodes/edges + topology};
\node[tokenTS,right=of ts] (tts) {windows + time/var ID};

\node[core,right=14mm of ta] (core) {{\bfseries Shared processing core}\\[2.5pt]\footnotesize\color{paperSubText} Transformer / SSM / Experts};

\node[outbox,right=14mm of core,yshift=14mm] (global) {\textbf{global metric vector}};
\node[outbox,right=14mm of core] (local) {\textbf{local / structured vectors}};
\node[outbox,right=14mm of core,yshift=-14mm] (private) {\textbf{input-specific info}\\[0.5pt]+ uncertainty};

\foreach \a/\b in {text/tt,vision/tv,audio/ta,graph/tg,ts/tts}{\draw[arr] (\a)--(\b);}
\draw[arr] (tt.east) -- ([yshift=7mm]core.west);
\draw[arr] (tv.east) -- ([yshift=3.5mm]core.west);
\draw[arr] (ta.east) -- (core.west);
\draw[arr] (tg.east) -- ([yshift=-3.5mm]core.west);
\draw[arr] (tts.east) -- ([yshift=-7mm]core.west);

\draw[arr] (core.east) -- (global.west);
\draw[arr] (core.east) -- (local.west);
\draw[arr] (core.east) -- (private.west);
\end{tikzpicture}
\end{adjustbox}
\caption{A shared processing core with modality-specific tokenizers and external structural information. Self-attention is permutation-equivariant and order-agnostic, while state space models impose a strictly one-dimensional scan. In either case, order, topology, rate, and variable identity must be re-supplied through positional encodings, scan orders, attention biases, edge features, or timestamped tokens.}
\label{fig:token-interface}
\end{figure*}
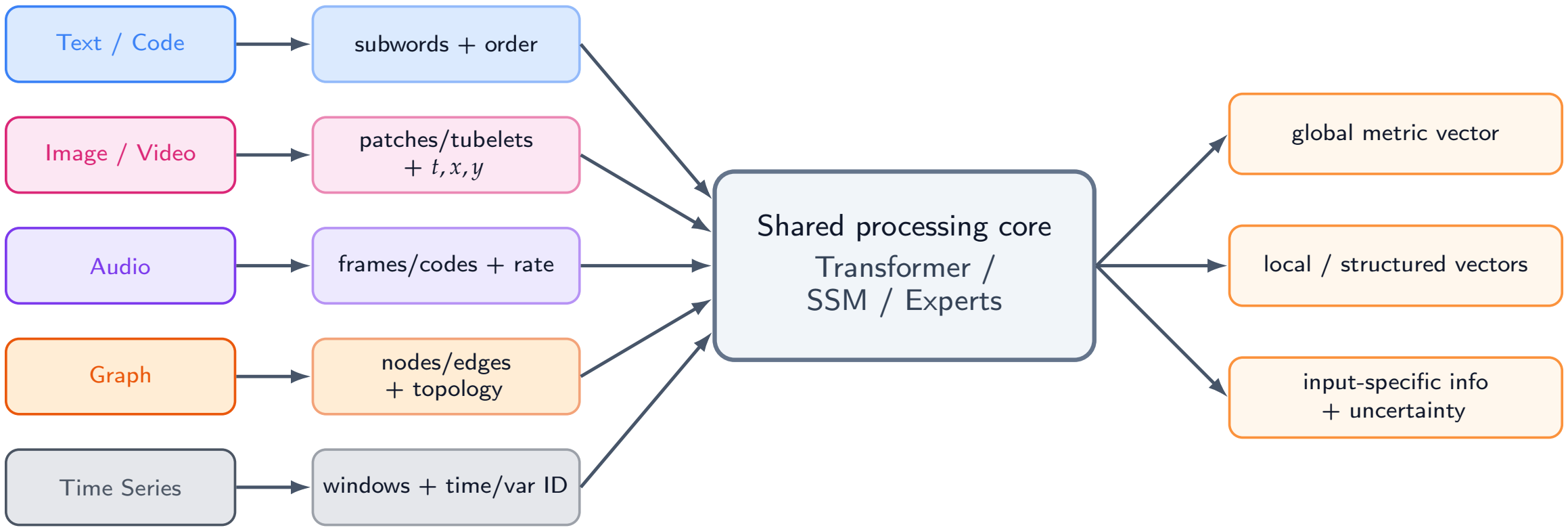

\paragraph{Tokenization: the modality interface.}
Tokenization---converting raw data into a sequence of discrete or continuous
tokens---is the abstraction that makes the Transformer modality-agnostic:
subword units for text, patches for images~\citep{dosovitskiy2021vit},
spectrogram patches for audio, continuous waveform features for
speech~\citep{baevski2020wav2vec2}, codec tokens for
codec-based speech models, node features for graphs, time-window segments
for time series~\citep{nie2023times}, and cells for tables~\citep{tapas2020,
tabbie2021}.
Meta-Transformer~\citep{metatransformer2023} demonstrates this explicitly:
the same frozen encoder, with modality-specific tokenizers and lightweight
heads, processes 12 modalities without any paired cross-modal data.
data2vec~\citep{data2vec} pushes the point further: one self-supervised
reconstruction objective serves speech, vision,
and language alike, with only the input processing, masking scheme, and
target normalization adapting to each modality.
\Cref{fig:token-interface} shows the resulting picture: a modality-specific
tokenizer is the prerequisite for interfacing with the shared processing core, and
\Cref{tab:tokenizers} records what the tokenizer emits for each named
modality and which symmetry its index structure carries (e.g., translation
equivariance over a regular lattice or permutation invariance over an
unordered set). This structure decides how the missing prior should
be re-injected: through a positional encoding over the lattice indices,
or through features supplied by the tokenizer or attention bias.

\begin{table*}[t]
\centering
\small
\caption{The tokenizer, rather than the Transformer trunk, is where named
modalities remain distinct. We group modalities by the symmetry their index
structure carries. \emph{Translation-symmetry} modalities are indexed by a
regular lattice (image, video, spectrogram, time index), so each token has a
fixed address that a positional encoding can name directly; \emph{Permutation-symmetry}
modalities (point cloud, table, graph) live on an unordered set with no
natural lattice; here self-attention's own permutation equivariance matches
the domain, but relational structure must be supplied through the
tokenizer or attention bias.}
\label{tab:tokenizers}
\adjustbox{width=\textwidth,center}{
\rowcolors{2}{white}{gray!10}
\begin{tabular}{@{}m{0.18\textwidth}m{0.30\textwidth}m{0.24\textwidth}m{0.26\textwidth}@{}}
\toprule
\textbf{Named modality} & \textbf{Tokenizer} & \textbf{Unit it emits} & \textbf{Symmetry of index set} \\
\midrule
\rowcolor{gray!30}\multicolumn{4}{@{}l}{\textit{Translation symmetry (regular lattice)}} \\
Natural language / code & WordPiece or BPE segmentation and vocabulary lookup~\citep{devlin2019bert} & Embedded subword & 1D translation equivariance \\
2D RGB image & Flatten fixed 2D patches and linearly project them~\citep{dosovitskiy2021vit} & Spatial-patch token & 2D translation equivariance \\
Video & Replace the image projection by a 3D convolutional patch embedding~\citep{tong2022videomae} & Spatiotemporal-patch token & 3D translation equivariance \\
Audio (spectrogram-based) & Log-mel spectrogram patched by a 2D convolution~\citep{imagebind} & Time--frequency patch token & 2D translation equivariance \\
Speech (waveform-based) & Multi-layer 1D temporal convolutions over the raw waveform~\citep{baevski2020wav2vec2} & Continuous latent-frame token & 1D translation equivariance \\
Depth / thermal / hyperspectral & Single- or multi-channel patch projection; spectral bands as tokens & Spatial/spectral-band token & 2D translation equivariance \\
Time series & Split each channel into fixed-length subseries patches and linearly project them~\citep{nie2023times} & Channel-local window token & 1D translation equivariance \\
\addlinespace
\rowcolor{gray!30}\multicolumn{4}{@{}l}{\textit{Permutation symmetry (unordered set)}} \\
3D point cloud & Farthest-point sampling, $k$-NN grouping, and learned local aggregation~\citep{qi2017pointnetpp} & Local point-group token & Permutation invariance over a metric set \\
Table & Column-keyed categorical/numerical embeddings contextualized across fields~\citep{turl2020,tabert2020} & One token per categorical/numerical field & Permutation invariance over keyed fields \\
Graph & Node-feature tokens plus learned centrality, shortest-path and edge encodings~\citep{khoshraftar2024survey} & Node token with structural attention biases & Permutation equivariance over a relational set \\
\bottomrule
\end{tabular}
}
\end{table*}

\paragraph{Adding structure back: positional encoding as a plug-in prior.}
At first glance, the same order-agnosticism that makes the Transformer
universal is a defect for modalities in which order matters: text is
sequential, time series is ordered in time, a graph's edges carry fixed
relations.
The resolution is architectural: for any modality where order is meaningful,
\emph{positional encodings} explicitly reintroduce the structural prior.
Text and time series use absolute or relative positional encodings over their
indices; graphs inject structural encodings, attention biases, or edge
features that expose topology to the otherwise structure-free
attention~\citep{tao2026code,yuan2026survey}; audio may attach timestamps to codec tokens;
and table encoders keep the 2D grid by attaching explicit row and column
index embeddings to each token~\citep{tapas2020}, or
applying vertical self-attention across the rows aligned to a given
column~\citep{tabert2020}.
This ``core mechanism plus pluggable priors'' design is the source of the
Transformer's flexibility: one operator with weak spatial and temporal
priors, combined with a modality-specific positional scheme, spans the whole
range from unordered graphs to strictly ordered time series.
In \citet{battaglia2018relational}'s terms, the weak relational bias is the
base layer on which modality priors are stacked as needed.

\paragraph{Scalability and parallelizability.}
Unlike RNNs, Transformers process all tokens in parallel after computing
attention scores, making training on hundreds of billions of tokens
feasible in wall-clock time;
and unlike MLPs~\citep{2003nnlm}---which likewise impose weak
structural priors---they exchange information across tokens with
a parameter count independent of context length.
Scaling laws~\citep{kaplan2020scaling, hoffmann2022chinchilla} give predictable
improvement as a function of size, data, and compute, and the Transformer is
uniquely amenable to exploiting these laws at scale.
Platonic-representation evidence~\citep{huh2024platonic} that larger models
across modalities grow increasingly aligned comports with the idea that the
Transformer's architectural generality both enables and accelerates the
convergence.

These four properties, taken together, reach their natural limit in one
further arrangement: a single shared trunk, front-ended by
modality-specific tokenizers and lightweight heads, serving every
modality at once.
The feasibility of such modality-agnostic processing was first sketched by
MultiModel's shared body over image, speech, and
text~\citep{kaiser2017multimodel} and most prominently established by
Perceiver and its
successors~\citep{jaegle2021perceiver, jaegle2022perceiverio,
zhu2022uniperceiver, li2023uniperceiverv2}, which showed that one Transformer
can process inputs as diverse as images,
audio, and point clouds with minimal modality-specific assumptions.
We name this arrangement the \emph{shared
backbone}. It is instantiated later by Meta-Transformer across twelve
modalities~\citep{metatransformer2023}, 4M-21 across
twenty-one~\citep{bachmann20244m21}, and OneLLM binding additional
modalities into one language-centered trunk through a universal projection
module~\citep{han2024onellm}.
We keep it conceptually separate from the three bridges of \Cref{part:path} because
what it shares is parameters, not representations: sharing parameters
alone leaves the per-modality outputs tied by no learned cross-modal metric,
and cross-modal retrieval requires at least an alignment objective on top of the
trunk---added, for instance, by Uni-Perceiver's shared cosine-similarity
loss~\citep{zhu2022uniperceiver}.
The shared backbone is therefore best read not as a route toward unifying
representations but as the culminating evidence of the architectural
convergence chronicled in this Part.

\subsection{Modality-Specific Architectures as Inductive Biases}
\label{sec:arch:specific}

Reading the above as ``the Transformer wins everywhere'' would be premature.
The modality-specific architectures---CNNs, GNNs, RNNs---are best understood
not as inferior predecessors but as encodings of strong domain priors the
Transformer omits, and there are regimes where those priors remain advantageous.
\begin{itemize}[leftmargin=2em]
  \item \textbf{CNNs encode translation equivariance and locality.} A
    convolution's shared weights and local receptive field are the
    prior that helps vision in low-data regimes~\citep{krizhevsky2012alexnet,
    long2015fcn}; group-equivariant and steerable CNNs make those symmetries
    explicit~\citep{cohen2016gcnn, cohen2017steerable, cohen2018spherical}.
  \item \textbf{GNNs encode a prescribed topology.} Message-passing over a
    fixed graph is the inductive bias that lets GNNs respect graph structure
    by construction~\citep{kipf2016semisupervised,velikovi2017graph,zhang2026llm}.
    Graph Transformers reuse attention as an operator but are ``graph
    mechanisms only when relational structure is encoded into tokenization,
    attention bias, positional signal, edge representation, or training
    task''---they are alternative routes for exposing structure to
    sequence-style architectures, not structure-free replacements for message
    passing~\citep{ying2021do,yuan2026survey}.
  \item \textbf{Code encoders inject program structure as a pluggable prior.}
    Code demonstrates the plug-in pattern at the level of the input
    and the attention mask: the same Transformer backbone becomes
    structure-aware once its inputs and masks carry data-flow or AST
    structure with them.
    GraphCodeBERT~\citep{graphcodebert2021} augments code and text tokens
    with data-flow nodes guided by graph attention masks constructed
    from data-flow and source-token relations; and
    UniXcoder~\citep{unixcoder2022} linearizes the AST into the token
    stream.
    Code thus supplies the sharpest worked example of the ``core mechanism
    plus pluggable structural priors'' recipe: the backbone is untouched,
    but its inputs and attention masks carry a program's syntax and
    semantics instead of raw surface tokens.
  \item \textbf{RNNs and TCNs encode temporal recurrence.} Recurrence and
    causal convolutions carry the sequential prior useful for time series in
    small-data regimes; the asynchronous (point-process) regime still relies on
    continuous-time LSTMs and attention over event sets~\citep{nhp, attnhp}.
    The state space family of \Cref{sec:arch:ssm} is the modern descendant of
    this lineage: the recurrence prior made parallelizable, and scaled until it
    too could claim backbone status.
\end{itemize}
The right framing is therefore a trade-off, not a verdict: strong
inductive biases buy data efficiency and low-data performance but limit
expressiveness, while the Transformer's weak bias buys generality and
scalability at the cost of being data-hungry.
Nor is the contest one-dimensional---a counter-current even within time series
argues that attention is not always the right primitive: TSLANet's adaptive
spectral filtering at $O(N\log N)$ beats Transformers in small-data, high-noise
regimes~\citep{tslanet}, and a single linear layer matches elaborate
Transformers on long-horizon forecasting~\citep{dlinear}.
Both are the expected consequence of a trade-off in which the modality's data,
scale, and priors jointly determine which architecture dominates.
The convergence is toward a \emph{default}: the Transformer is the first choice
when data and compute are ample, and modality-specific architectures remain the
right tool when priors are strong and data are scarce.

\subsection{The Rival Universal Encoder: State Space Models and Linear Attention}
\label{sec:arch:ssm}

Before accepting the Transformer as the universal encoder, we must also examine
its most prominent rival---and the rival's own trajectory
turns out to confirm, rather than contradict, the convergence this Part
documents.

\paragraph{A second family born cross-modal.}
The structured state space line begins with S4, proposed explicitly as a step
toward a ``general-purpose sequence model'' and evaluated at birth on
text, image, speech, and time series~\citep{gu2022s4}; it is streamlined by S5~\citep{smith2023s5} and
culminates in Mamba, whose selective SSM (named S6) lets the recurrence's
parameters depend on the input, so that the model chooses token by token
what its fixed-size state keeps---with state-of-the-art results reported
across language, audio, and genomics~\citep{gu2023mamba}.
The family has since swept the modalities at a speed rivaling the
Transformer's own: Vim and VMamba for images~\citep{zhu2024vim,
liu2024vmamba}, Audio Mamba for spectrograms~\citep{erol2024audio}, S-Mamba
for time-series~\citep{smamba2024}, Graph Mamba for
graphs~\citep{behrouz2024graphmamba}, and long-range DNA sequence
models---spanning implicit-convolutional relatives such as HyenaDNA and
bidirectional Mamba backbones such as Caduceus---with
context lengths up to a million nucleotides~\citep{nguyen2023hyenadna,
schiff2024caduceus}.
And the sweep reuses the recipe of \Cref{sec:arch:transformer} rather than
replacing it: SSMs consume the same tokenization interface (Vim simply takes
ViT patches), and the structure that attention recovers through positional
encodings must be re-injected through the same pluggable priors---position
embeddings plus bidirectional scans for Vim's image sequences, a cross-scan
traversing the spatial axes for VMamba, neighborhood tokenization and
explicit token ordering for Graph Mamba~\citep{zhu2024vim, liu2024vmamba,
behrouz2024graphmamba}.
Even here, modality differences reassert themselves. Mamba's own
ablations find input-dependent selection (the S4-to-S6 change) decisive for
discrete, information-dense sequences like language, while on continuous,
smoothly varying signals such as raw audio waveforms the non-selective LTI
recurrence (S4) remains preferable, its time-invariance matching the
signal's structure~\citep{gu2023mamba}. This reinforces the core lesson of
this Part: the mechanism is the intersection, and the data decides.

\paragraph{What the fixed state yields, and what it sacrifices.}
Where attention keeps every token available to every other, an SSM compresses
the past into a fixed-size state updated sequentially, achieving linear-cost
computation and constant-memory inference at the expense of random access to
its input.
Analyses from both directions agree that the toll is real: generalized SSMs
are provably weaker than attention at copying and retrieving from
context~\citep{jelassi2024repeat}, and despite their recurrent form they
inherit expressiveness limits similar to attention's
own~\citep{merrill2024illusion}.
For embedding models, this trade-off is particularly acute: rerankers
bind query tokens to
candidate tokens, instruction-conditioned encoders retain the prompt information
until the readout, and dense encoders must
settle token-level correspondence through all-to-all contextual
disambiguation before pooling---all content-addressable lookups that a
fixed-size state approximates poorly.
Linear attention marks the middle ground: rewriting attention with kernel
feature maps turns the quadratic operator into a constant-memory, linear-time
recurrence~\citep{katharopoulos2020linear}, a lineage that runs through RWKV
and RetNet's retention~\citep{rwkv2023, retnet2023}; and the structured
state-space duality result shows SSMs and attention variants to be closely
related decompositions of one matrix class~\citep{dao2024mamba2}, so the
practical question is less \emph{which family wins} than \emph{where each
belongs}.
For the models this survey tracks, the answer still favors the Transformer:
the interaction and recall on which embedding quality leans are exactly what
the state space family sacrifices to achieve its efficiency, and that efficiency is
decisive only where sequence length rather than association is the binding
constraint---raw audio, video, and genomic sequences at lengths attention
cannot afford.
The SSM wave is therefore best read not as a counterexample but as a second
demonstration of this section's argument: the convergence runs deeper than
either backbone, toward modality-agnostic token processors that leverage
per-modality tokenizers and injected structural priors.

\subsection{Beyond the Backbone: Interaction and Aggregation}
\label{sec:arch:pooling}

While the backbone maps input tokens to contextualized hidden states, turning those states into a deployable embedding requires two further architectural commitments that recur across all modalities: how multiple inputs interact, and how token sequences are aggregated.

\paragraph{Interaction architectures: dual-encoders versus cross-encoders.}
When a task pairs queries with candidates or combines multiple modalities, two interaction patterns dominate, directly setting up the unification paths of \Cref{sec:alignment,sec:fusion}.
The \emph{dual-encoder} (bi-encoder) encodes query and candidate independently and compares them only in the output metric space, enabling approximate nearest-neighbor search over pre-computed indexes at scale and serving as the canonical architecture of cross-modal alignment~(\Cref{sec:alignment}).
Conversely, the \emph{cross-encoder} concatenates query and candidate tokens for all-to-all attention, achieving finer token-level correspondence at the cost of one joint forward pass per query--candidate pair.
In unimodal stacks, cross-encoder typically serves as a reranker over a dual-encoder shortlist~\citep{monot5_2020, rankt5_2023}. When generalized across modalities, it forms the architectural basis for multimodal joint fusion~(\Cref{sec:fusion}).

\paragraph{Pooling and aggregation.}
Given an encoder's hidden states $(\vh_1,\ldots,\vh_L)$, an aggregation operator must reduce the sequence into the exposed representational object, which governs the granularity trade-offs of \Cref{sec:data}.
Bidirectional encoders typically rely on a dedicated \texttt{[CLS]} token, mean pooling, or learned attention pooling~\citep{devlin2019bert,lee2025nvembed}, whereas causal LLM embedders read out the last \texttt{[EOS]} token~\citep{qwen3embed,f2llm}.
Crucially, the pooling choice must respect the requirements of downstream tasks: dense vision and document models bypass global reduction to retain a set of patch vectors for late-interaction retrieval~\citep{faysse2024colpali}, while audio speaker models use statistics pooling (concatenating the frame-wise mean and standard deviation), deliberately preserving the second-order temporal variability that standard text-side global pooling would erase~\citep{xvector}.


\highlightbox{tred}{Takeaway}{%
The Transformer converges as the default universal encoder not by
eliminating inductive biases, but by adopting a minimal, set-based
operator whose missing structural priors (order, lattice, topology)
are re-injected via modality-specific tokenizers and positional
encodings. While emerging state space models share this tokenized
interface, their fixed-size states sacrifice the all-to-all contextual
resolution and recall essential for fine-grained retrieval representations.
Meanwhile, modality-specific architectures (CNNs, GNNs) remain principled
choices rather than outdated predecessors, offering vital data efficiency
in strong-prior, data-scarce regimes.%
%
}

\section{Data and Representational Objects: Units, Granularity, and the Upward Shift}
\label{sec:data}

The third axis of convergence is an upward shift in representational granularity:
across all modalities, the favored embedding target is moving from local
parts (tokens, pixels, frames, nodes) to complete semantic
units (sentences, images, utterances, graphs, series).
The consequences of this shift are the focus of this section: what it preserves,
what it discards, and where it is reversed.
We unpack it by first distinguishing the input data unit from the
output representational object~(\Cref{sec:data:units}), and then analyzing
the granularity shift itself, alongside the tasks that require reversing
it~(\Cref{sec:data:shift}).

\subsection{Units and Representational Objects}
\label{sec:data:units}


A subtle but pervasive confusion in the literature conflates the model
input (the \emph{data unit}) with the extracted output (the \emph{representational object}).
The data unit is the atom of the input: the subword, the patch, the
spectrogram frame, the node, the cell, and the timestamp at which an encoder
first sees the signal.
The representational object is the atom of the output---the entity at
which an embedding is actually assigned and which downstream consumers
retrieve, compare, and combine.
Distinguishing the two is essential because
input units such as pixels are physical measurements, while the choice
of representational object is a modeling commitment to which information
must survive the interface and which may be discarded~\citep{nutime},
determining what an embedding can do.

The input data unit reflects the physical structure of each modality,
and the granularity of that unit trades off resolution against
sequence length: finer partition preserves more detail but lengthens the
sequence the encoder must process.
The families supply different candidate units, as cataloged in \Cref{tab:anatomy}.
Text and code have characters, subword pieces, and words,
with boundaries partly supplied by a symbolic system.
Vision offers pixels, patches, regions, and frames.
Audio has waveform samples, spectrogram frames, and continuous
latent frames.
Time series provides timestamped samples, channels, windows, and patches.
Graph methods ingest nodes, edges, typed triples,
paths, or subgraphs~\citep{bordes2013transe, perozzi2014deepwalk,
ying2021do,khoshraftar2024survey}.
Tables offer cells, rows, and columns.
A recurring difficulty is that the unit is often a manufactured
choice, not a given one:
a pixel in an image or a sampled value in a time series carries
no standalone meaning~\citep{nie2023times}, and an audio frame is not
an object the way a word is in text.
This difficulty has
led to an extreme line of work that bypasses manual partitioning
altogether by taking raw bytes directly as inputs, including
ByT5~\citep{xue2022byt5} for text and ByteFormer~\citep{horton2024bytes}
for multiple modalities.

\begin{table*}[t]
\centering
\small
\caption{Different representational objects support different uses and
entail different trade-offs. The representational object is exposed
to downstream consumers; this choice---not the encoder---decides what
an embedding can do.}
\label{tab:repr_objects}
\adjustbox{width=\textwidth,center}{
\rowcolors{2}{gray!10}{white}
\begin{tabular}{@{}m{0.16\textwidth}m{0.16\textwidth}m{0.32\textwidth}m{0.30\textwidth}@{}}
\toprule
\textbf{Output format} & \textbf{Form} & \textbf{Typical uses} & \textbf{Main trade-off} \\
\midrule
Global dense vector & $\ve\in\R^{d}$ & ANN retrieval, clustering, linear probes & Loss of fine spatial, temporal and relational evidence \\
Local vector sequence/set & $\mE\in\R^{N\times d}$ & Grounding, late interaction, reranking & Larger indexes and less predictable storage \\
Discrete codes & $\vc\in\{1,\ldots,K\}^{L}$ & Compression, generation, sequence modeling & Quantization loss and potentially unstable code semantics \\
Probability distribution & $p(\vz\mid x)$ & Ambiguity, calibration, risk-aware retrieval & More complex indexing than point vectors \\
Magnitude-preserving vector & $\ve\in\R^{d}$ (scale retained) & Time-series retrieval where amplitude \emph{is} the semantics & Must balance scale sensitivity against the shape invariance that aids cross-domain transfer \\
\bottomrule
\end{tabular}
}
\end{table*}

Representational objects, by contrast, represent the output semantic
entities. They are independent of how inputs are partitioned, but instead
reflect the requirements of downstream applications, ranging from atomic
components for tagging to macroscopic wholes for search.
Five forms of representational objects recur across modalities, and
\Cref{tab:repr_objects} summarizes the trade-off each entails. A \textbf{global
vector} $\ve \in \R^{d}$ is one point per input, ideal for retrieval and
classification, but it entangles every piece of evidence into a single
coordinate and discards location, multiplicity, and correspondence. A
\textbf{sequence of local vectors} $\mE \in \R^{N \times d}$---the ViT patch
features, the wav2vec frame features, the tabular cell states---preserves a
spatial or temporal tessellation for dense prediction and late interaction,
at the price of storage and comparison cost~\citep{akbik2018contextual,
tabbie2021, faysse2024colpali}. A \textbf{set of discrete codes}
$\vc \in \{1, \ldots, K\}^{L}$---VQ-VAE indices, audio codec tokens, HuBERT
cluster assignments---is the bridge by which audio and part of vision
entered the language-model ecosystem~\citep{vandenOord2017vqvae,
hsu2021hubert, defossez2022encodec}. A \textbf{probability distribution}
$p(\vz \mid x)$---the latent of a VAE or a probabilistic cross-modal
retriever---represents the multiplicity and uncertainty that a single point
cannot~\citep{kingma2014vae, chun2021pcme}. And \textbf{magnitude-preserving
vectors} retain the numerical level of the signal, so that a temperature of
$30^\circ$ and $3^\circ$ remain distinguishable---a property time series needs and
no vocabulary can enumerate~\citep{nutime}.
No object is universally best: the representational object is the principal
axis along which modalities differ even as their encoders and
objectives converge, and it is why a ``stronger encoder'' is not the same
thing as a ``better embedding''.

\subsection{The Granularity Shift---and Why It Is Partially Reversible}
\label{sec:data:shift}

The convergence of this section is that each modality's preferred
representational object is climbing upward, from local parts to whole
semantic units,
driven by the need for invariance, fixed-size indexing, and efficient comparison.

The shift is most legible in text, where it is also the oldest.
NLP embeddings climbed the unit inventory one rung at a time: from
token-level static embeddings that placed each word at a single point
independent of context~\citep{mikolov2013distributed, pennington2014glove}
to sentence-level contextual and contrastive embeddings that pool a
whole sentence into one vector~\citep{reimers2019sbert, gao2021simcse}, and on
to retrieval-level dense embeddings designed so that an entire
query or passage is a single searchable point~\citep{karpukhin2020dpr,
izacard2022contriever, wang2022e5}.
Each step pooled finer units into a coarser, semantics-bearing object, and
each step was motivated by the same insight: the most directly comparable
representation encodes a whole semantic unit, not its parts.
Vision traces the same arc: local descriptors and patch-level objectives
give way to whole-image readouts---image-level contrast~\citep{chen2020simclr}
and masked patch reconstruction~\citep{he2022mae}---and now to
region- and set-level readouts that retain correspondence when the
task needs it~\citep{faysse2024colpali}.
Audio moved from frame-level masked prediction~\citep{baevski2020wav2vec2}
to utterance- and clip-level discrimination and cross-modal
alignment~\citep{clap}.
Time series moved from point-level classification to window/patch-level
modeling~\citep{nie2023times} and series-level universal
representations~\citep{ts2vec, moment}, and
graph embeddings from node-level proximity to subgraph- and graph-level
readouts~\citep{surveygraphembed2018,khoshraftar2024survey}.
The pattern is general enough to read as a law: as a literature
matures, the granularity at which it embeds climbs from a measurable atom toward
a semantic whole.

The shift is not, however, monotonic toward a universally better unit.
The pooling operation that drives it upward can discard the
locations, counts, ordering, uncertainty, factor identity, or relations
needed by the next task, so modern systems move in both directions.
Dense and region-level vision features preserve
correspondence~\citep{faysse2024colpali}; video systems retain temporally
aligned clip representations for localization and
retrieval~\citep{videoclip2021}; audio diarization requires frame-wise
speaker identity through overlapped speech~\citep{eend2019}; graph systems
expose nodes or subgraphs alongside whole-graph
summaries~\citep{perozzi2014deepwalk, xu2018how,sun2020infograph}; and temporal models expose
windows or event histories~\citep{nhp, ts2vec, easytpp}.
The result is best read as a lattice of permissible readouts rather
than a historical ladder: an encoder may expose a compact global vector for
first-stage search, a local sequence or set for verification and grounding,
discrete codes for reconstruction, and calibrated uncertainty for ambiguous
items.
Global vectors favor latency and storage; local or set-valued outputs favor
correspondence and attribution; discrete objects favor rate-controlled
generation~\citep{defossez2022encodec}; distributions favor calibrated
ambiguity~\citep{chun2021pcme}.
Evaluation should therefore report both the information an embedding
preserves and the invariances it induces.

The representational object and chosen granularity established here decide
what can be aligned, fused, or reused in
\Cref{sec:transformation,sec:alignment,sec:fusion}: a global vector, a local set,
a code sequence, or a distribution each exposes a different cross-modal
contract, and the three paths of \Cref{part:path} are each predicated on a different
assumption about which contract prevails.

\highlightbox{tred}{Takeaway}{%
The third axis of convergence is the upward shift of the representational
object: every modality has moved from embedding the measurable atom (token,
pixel, frame, sample) toward embedding a whole semantic unit (sentence,
image, utterance, series), driven by the same insight---the most directly
comparable representation encodes a whole, not its parts.
The shift is not a one-way ladder but a bounded trade: pooling into a coarser
object buys invariance and efficient indexing yet discards the location,
multiplicity, and correspondence some tasks need, so a mature encoder
exposes a lattice of readouts and moves up or down it as the task
demands.
That lattice, not any single object, is what the alignment and fusion paths
of \Cref{part:path} must operate on.%
}

\clearpage
\part{The Path to Unification}
\label{part:path}

\begin{figure*}[t]
    \centering
    \includegraphics[width=\textwidth]{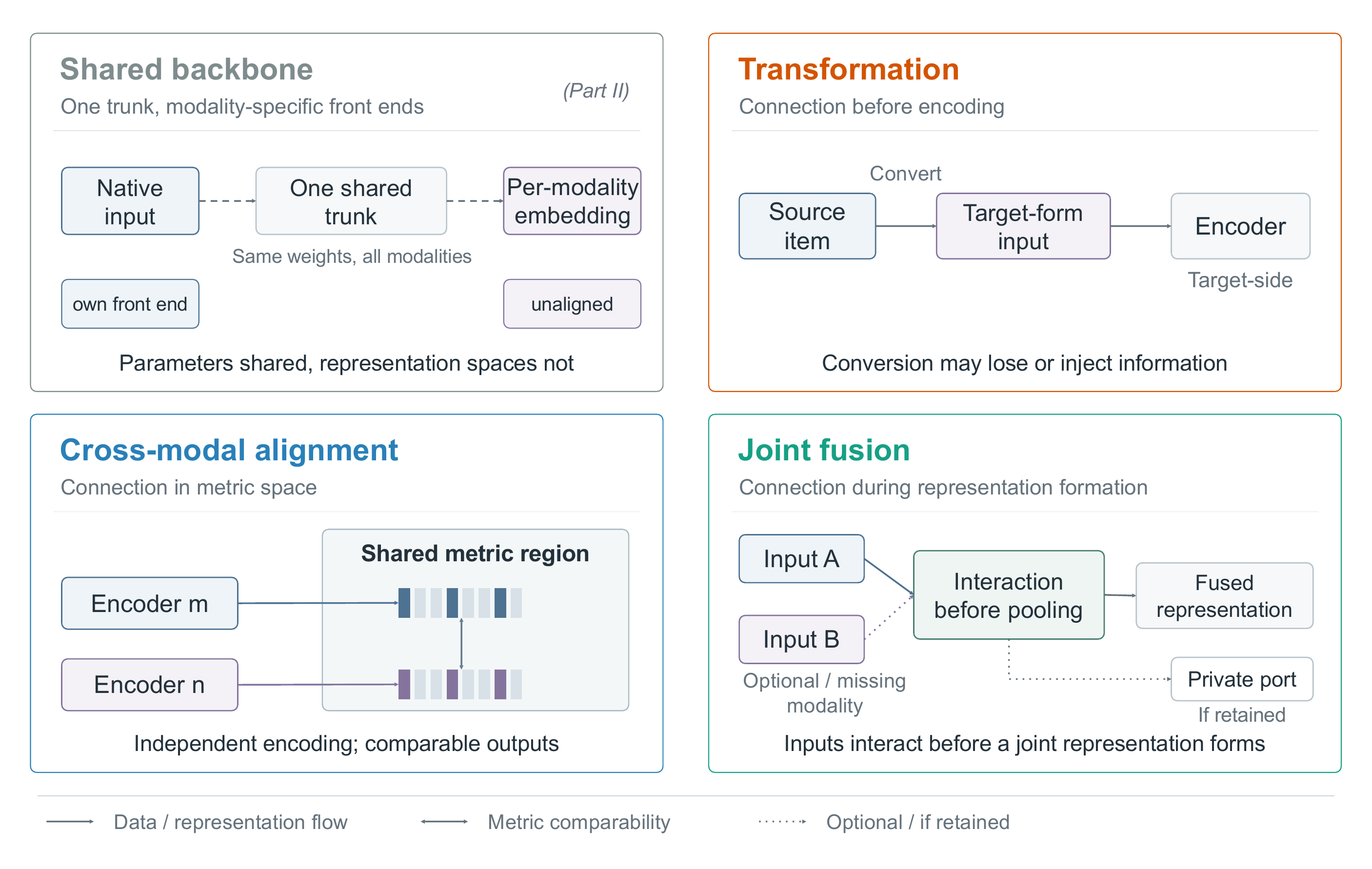}
    \caption{Four ways of sharing work across modalities. The leading cell is the shared backbone (\Cref{sec:arch}): one model serves every modality through modality-specific front ends, reusing parameters without necessarily producing a shared representation space. The remaining three are the three bridges of \Cref{part:path}, in order of increasing cross-modal interaction: Transformation (\Cref{sec:transformation}) converts a source item into a target-form input before encoding, potentially losing or injecting information. Cross-modal alignment (\Cref{sec:alignment}) makes independently encoded representations comparable under a shared metric. Joint fusion (\Cref{sec:fusion}) lets modalities interact before pooling, with optional inputs and a private port retained, subsuming the alignment case (interaction set to zero).}
    \label{fig:cross-modal-paradigm-map}
\end{figure*}

We now trace the three-step progression toward omnimodal embeddings.
\Cref{fig:cross-modal-paradigm-map} sets the shared-backbone case and the
three paths side by side in a single picture before we unpack them in turn:
transformation~(\Cref{sec:transformation}), alignment~(\Cref{sec:alignment}),
and joint fusion~(\Cref{sec:fusion}), in order of increasing cross-modal interaction.
To grade what each path actually achieves, we recall the 3C lens introduced
in \Cref{sec:intro:roadmap}, which evaluates a system by three progressively
harder capabilities: \emph{coverage}~(C1: which modalities the system can
ingest and test), \emph{comparability}~(C2: whether separately encoded modalities
can be matched in a shared metric space), and \emph{compositionality}~(C3: whether
a fused representation of multiple inputs genuinely depends on all of them
rather than their simple sum). As we will see across \Cref{part:path}, the three
paths satisfy these gates in turn.

\section{Path 1: Unification by Transformation}
\label{sec:transformation}

The first and most pragmatic path avoids building new multimodal architectures
altogether but reuses existing models by transforming non-native inputs into a
modality that is already well-served by mature systems:
rendering non-visual data as images~(\Cref{sec:transformation:rendering}),
serializing non-text structures into text~(\Cref{sec:transformation:serialization}),
or quantizing continuous signals into discrete token
vocabularies~(\Cref{sec:transformation:tokens}).
All are ingenious, and all suffer from information loss.
Viewed through the 3C lens, transformation primarily provides coverage: it
lets an encoder trained on one modality accept another, but by itself guarantees
neither comparability nor compositionality---which is why transforming
modalities into one another so often loses the structure that the next two paths
have to put back.

\subsection{Rendering as a Unifying Canvas}
\label{sec:transformation:rendering}

The first transformation treats pixels as a common computational medium: a
deterministic or learned renderer $R_m$ maps an input $x_m$ to an image, after
which a visual backbone produces $\ve_m = f_{\mathrm{vis}}(R_m(x_m))$.
The appeal is obvious: it reuses the mature, heavily-pretrained image
encoders and preserves the visual layout. But a
caveat is that comparability is not free: rendering confers a shared shape, not
shared semantics, and the renderer becomes part of the representation function.

Rendering has been tried across an instructively wide range of modalities:
\begin{itemize}[leftmargin=2em]
  \item \textbf{Text $\to$ image.} PIXEL~\citep{rust2023pixel} renders sentences
    as images and performs masked-patch pretraining, replacing a finite subword
    inventory with glyph-level visual input; CLIPPO~\citep{tschannen2023clippo}
    trains a single ViT under a contrastive objective over natural images and
    rendered text, extending the same pixel interface to image--text
    retrieval and language tasks.
    (CLIPPO thereby straddles two paths: its interface is pure
    transformation, its objective is paired alignment. We group it
    throughout by its primary mechanism, as \Cref{fig:modality-trajectories}
    makes explicit.)
  \item \textbf{Documents and screenshots $\to$ image.} The rendering family's
    most consequential success is where the modality's semantics \emph{is}
    visual layout: Pix2Struct pretrains on rendered web pages to parse
    screenshots~\citep{lee2023pix2struct}; DSE directly embeds page screenshots
    as single dense vectors~\citep{ma2024dse}; and ColPali shows that
    late-interaction retrieval over document images outperforms
    OCR-then-text pipelines on visually rich documents~\citep{faysse2024colpali}.
    The LayoutLM line takes a middle route by parsing the page first,
    and then injecting the recovered layout back into the token stream:
    two-dimensional positional embeddings over OCR tokens~\citep{xu2020layoutlm},
    joined by the rendered image as an extra input channel
    in v2~\citep{xu2021layoutlmv2} and by unified text--image masking with
    word--patch alignment in v3~\citep{huang2022layoutlmv3}.
    Here rendering is not a lossy approximation of text but the more
    faithful representation: a flat text serialization of a document already
    discards the two-dimensional layout that carries its meaning.
  \item \textbf{Text $\to$ image, again---but for efficiency.} A 2025 line
    inverts PIXEL's motivation: rather than rendering text for
    coverage, it renders text for \emph{compression}.
    Glyph renders long contexts into images whose visual tokens are
    3--4$\times$ fewer than the original text tokens~\citep{cheng2025glyph},
    and DeepSeek-OCR pushes optical compression past 10$\times$ at usable OCR
    accuracy~\citep{wei2025deepseekocr}.
    Transformation, in this second guise, is no longer a compromise made for an
    unserved modality---it is a claim that the pixel channel is a
    denser code for text than text tokens themselves.
  \item \textbf{Graph $\to$ image.} GITA~\citep{wei2024gita} renders a graph as
    a drawing paired with a textual description for vision-language graph
    reasoning. The strategy is clever but lossy: one drawing encodes a
    graph's topology only as well as its labels and conventions allow, and
    the negative evidence is that
    multimodal LLMs still struggle with basic visual network
    analysis~\citep{williams2024vna}.
  \item \textbf{Point cloud $\to$ image.} PointCLIP projects a 3D point cloud
    into multi-view depth maps and feeds a pretrained CLIP without any 3D
    pretraining~\citep{zhang2022pointclip}---the transformation route by which the
    3D point cloud family first reached a shared-space
    model.
  \item \textbf{Time series $\to$ image.} Gramian Angular Fields and Markov
    Transition Fields convert a sequence into an image whose entries encode
    pairwise angular geometry or transition statistics before CNN
    processing~\citep{wang2015imagingtimeseries}; VisionTS recasts forecasting
    as image reconstruction and reuses a visual masked autoencoder without any
    time-series pretraining~\citep{chen2024visionts}; and ViTST renders
    irregularly sampled series as line-chart images for a
    ViT~\citep{li2023vitst}, the drawing conventions of the chart substituting
    for the interpolation a numeric pipeline would have to invent.
\end{itemize}

The successes and failures of rendering hinge on a single question: whether
a two-dimensional projection preserves the structure the task actually queries.
It works when the visual mapping captures the semantics that matter, such as
glyph shapes of text or recurrence patterns of time series. And it excels in
layout-bearing modalities: for documents, slides, web pages, and charts, the
screenshot is not a lossy projection of the semantics but \emph{is} the
semantics~\citep{ma2024dse,faysse2024colpali}.
Conversely, rendering fails when it flattens non-Euclidean topology or continuous
magnitudes into pixels and forces a vision encoder to awkwardly re-learn domain
invariants that were explicit in the raw data: a 2D drawing can leave a graph's
edges implicit, and a chart that rescales its axes discards the magnitude that
time series treats as semantics~\citep{nutime}.

\subsection{Serialization as a Unifying Language}
\label{sec:transformation:serialization}

The second transformation converts non-text data
toward the mature text stack, either as strings that off-the-shelf text
models consume unchanged, or as continuous inputs that
are projected directly into a language model's token embedding space.

\begin{itemize}[leftmargin=2em]
  \item \textbf{Time series $\to$ string.} Long before LLMs, SAX
    symbolized a z-normalized, piecewise-aggregated window into a word over a
    small alphabet~\citep{sax}, and symbolic successors such as
    SAX-VSM~\citep{saxvsm} and BOSS~\citep{boss} built classifiers
    and retrievers on the resulting words.
    The LLM era revived the same interface at scale: LLMTime shows that
    GPT-3, given a string of number sequence, is a
    competitive zero-shot forecaster, leveraging the unchanged next-token
    prior~\citep{gruver2023llmtime};
    PromptCast recasts forecasting entirely as natural-language
    prompting~\citep{xue2023promptcast}; and TimeRAG retrieves numerical
    neighbors and injects them into the forecaster LLM as text prompts~\citep{timerag}.
    LETS-C~\citep{letsc2025} serializes each series for a general text
    embedding model and fuses that embedding with the preprocessed numerical
    series by element-wise addition, outperforming 27 baselines on a
    ten-dataset UEA benchmark. The headline is provocative but not a clean
    control: either input alone trails the full model, so whether the text
    embedding itself---rather than the fused raw series---matches
    purpose-built architectures is a question left open.
  \item \textbf{Time series $\to$ LLM embedding space.} A parallel
    stem bypasses the text interface and reuses the
    pretrained weights directly: TEST aligns
    series embeddings with text-prototype axes in a frozen LLM's embedding
    space~\citep{test}, Time-LLM reprograms patched series into mixtures
    of the LLM's own word-embedding prototypes~\citep{jin2024timellm},
    GPT4TS/OFA lightly finetunes the positional embeddings and layer norms
    of a frozen pretrained Transformer backbone for any time-series
    task~\citep{zhou2023gpt4ts}, and aLLM4TS adapts LLM backbones via
    multi-patch prediction~\citep{allm4ts}.
    Unlike the discrete tokenizers of \Cref{sec:transformation:tokens},
    this family does not introduce new time-series tokens, but repurposes the
    LLMs' sequence-processing weights by directly accepting pseudo tokens of
    continuous patches. And while this family is mechanically
    similar to the visual projectors used in MLLM
    embedders (\Cref{sec:fusion:paradigm}), we group it under transformation
    because the works in this family primarily repurpose a frozen text backbone
    to process a single non-text modality, rather than jointly pooling multiple
    interacting modalities into a unified representation.
  \item \textbf{Graph $\to$ text.} Serializing a graph is not one operation
    but a design space: ``Talk like a Graph'' shows that the choice of
    verbalization---node list, edge list, adjacency, or incident
    description---changes LLM performance by double digits~\citep{fatemi2024talklikeagraph};
    GraphText serializes node attributes and inter-node relationships by traversing a graph-syntax tree~\citep{zhao2023graphtext}; LLaGA organizes nodes into structure-aware sequences and projects their representations into the LLM's token embedding space~\citep{chen2024llaga}.
    The same lesson GITA taught by drawing---one projection does not automatically preserve all of a graph's topology---reappears here in linguistic form: which edges survive serialization is a modeling decision, not a property of the data.
  \item \textbf{Tables $\to$ serialized schema and cells.}
    For tables, the serialization-format sensitivity is well-documented:
    TAPEX established that a shared text decoder can execute SQL over a
    linearized table~\citep{liu2022tapex}; TableLlama trains a language model over serialized tables~\citep{zhang2024tablellama}; and the
    ``Table meets LLM'' benchmark measures how markdown, JSON, and other
    formats alter LLM accuracy, finding none uniformly
    dominant~\citep{sui2024tablemeetllm}.
    TableRAG embeds a table by serializing its schema (column names,
    types, example values) and its cells (column--value pairs) with an
    off-the-shelf text encoder, then retrieves only the top-$k$ fragments into
    the LM~\citep{tablerag2024}.
    The approach makes million-token tables tractable, and is the clearest
    instance of transformation at work today. However, it is also a cautionary tale:
    a serialized ``column\:=\:value'' string has already discarded the
    row--column geometry that structure-aware table encoders
    exist to preserve~\citep{turl2020, tabbie2021}, so the transformation pays for
    scale with the very structure that made the table a distinct modality.
  \item \textbf{Non-text modalities $\to$ text via a cascade.} The most common
    transformation in production systems is also the least discussed in the
    literature: transcribe, caption, then embed as text.
    Speech enters a text embedder through ASR (Whisper's outputs are embedded
    as if they were written language~\citep{radford2023whisper}), and images or
    videos enter through captioning (BLIP-style captioners~\citep{li2022blip}
    feed a text retriever in the default multimodal RAG pipeline).
    This cascade is ``text-as-hub'' in its most literal form: it works
    only to the extent that speech and vision are already
    verbalizable, and it fails, by construction, on content that never
    fits in words---music, scene timbre, texture, and layout.
\end{itemize}
Serialization is appealing because it is convenient and leverages the scaling
advantages of LLMs, but it faces a structural problem: a string is a
1D sequence, so whatever the modality's native structure, the serialization
must spell it out linearly---and the formats in wide use do not spell it all
out, discarding the channel structure of multivariate
series, the topology of graphs, the row--column geometry of tables,
and the magnitude that time series
treats as semantics.
The loss is not uniform, but most pronounced where the format leaves the
native structure implicit, which is why tables and graphs
suffer most under serialization while univariate forecasting suffers least.

\subsection{Discrete Tokens: Leveraging the LM Interface}
\label{sec:transformation:tokens}

Rather than mapping data into natural language or pixels, the third
transformation targets the language model interface itself:
vector quantization~\citep{gray1984vq,vandenOord2017vqvae}
maps continuous signals onto learned codebooks,
yielding discrete token streams drawn from a finite vocabulary
and thereby bridging continuous signals to the language model stack.
It is distinct from serialization because the ``words'' here are
not attested in a language, but rather invented by a codec.

\begin{itemize}[leftmargin=2em]
  \item \textbf{Audio $\to$ semantic and codec tokens.}
    Audio entered the discrete-token ecosystem in two distinct waves. The first discretized continuous speech to unlock BERT-style MLM reconstruction for representation learning, bypassing the interpolation pitfall of regressing raw continuous frames~(\Cref{sec:objectives:reconstructive}): wav2vec 2.0 quantizes latent frames for contrastive targets~\citep{baevski2020wav2vec2}, HuBERT predicts offline cluster assignments~\citep{hsu2021hubert}, and BEATs trains an iterative acoustic tokenizer to supply discrete supervision for general audio encoders~\citep{beats}.
    The second wave originated in neural audio compression: neural codecs such as SoundStream and EnCodec~\citep{soundstream, defossez2022encodec} use residual vector quantization (RVQ) to compress waveforms into discrete codebook indices for low-bitrate transmission and high-fidelity reconstruction.
    Recent work bridges the two waves by uniting representation learning and quantization, capturing both acoustic details and high-level semantics: SpeechTokenizer~\citep{zhang2024speechtokenizer} and HASRD~\citep{hasrd} distill SSL representations (e.g., HuBERT) into the first RVQ layer while leaving subsequent layers to capture acoustic residuals; RepCodec vector-quantizes pretrained speech representations directly into discrete semantic tokens~\citep{repcodec}; Language-Codec compresses information across fewer codebooks to ease language-model consumption~\citep{languagecodec}; and EnCodecMAE closes the loop by using neural codec tokens as the prediction targets for universal masked autoencoding~\citep{encodecmae}.
  \item \textbf{Time series $\to$ tokenized vocabularies.}
    Time series similarly borrow the discrete LM interface to reuse
    language backbones and token-level objectives:
    Chronos tokenizes a
    series by scaling and rounding its values onto a fixed integer
    vocabulary and forecasts with
    a language model over that vocabulary~\citep{chronos}; TOTEM learns a
    discrete codebook that yields transferable embeddings across diverse
    domains with no shared sampling rate or sensors~\citep{totem},
    and byte-pair encoding over quantized values
    builds a data-specific vocabulary~\citep{tsbpe}.
\end{itemize}

Compared to text serialization, learned discrete tokens can preserve
signal structure in analog details while still unlocking the LM ecosystem's
training objectives and vocabulary interfaces.
However, this strategy faces a fundamental boundary condition:
speech has a natural inventory of recurring phonetic units that
discrete clusters readily capture~\citep{hsu2021hubert},
whereas arbitrary continuous signals may not.
While a learned codebook can manufacture a domain-general vocabulary,
as demonstrated by BEATs in general audio~\citep{beats} and
TOTEM in time series~\citep{totem},
its tokens lack a cross-corpus convention,
making vocabulary stability across domains an empirical precondition that
we formalize as a reuse rule in \Cref{sec:frontier:pollination}.


In summary, while transformation provides an immediate route to cross-modal
coverage by borrowing mature text and vision pipelines, this convenience comes
at the cost of structural loss---flattening the very geometric or relational
properties native to the source data. To preserve these native structures
without lossy projections, the field turns to alignment, which leaves
unimodal encoders intact and knits their coordinate spaces together.

\highlightbox{tred}{Takeaway}{%
Transformation is a pragmatic shortcut that provides coverage at the
expense of native geometry. Rendering preserves 2D layout but flattens
magnitude and graph topology; serialization borrows LLM infrastructure
but linearizes multidimensional relations; discretization enables LM
sequence modeling but hinges on recurring semantic units.
When a modality’s intrinsic semantics \emph{is} visual layout (e.g., documents),
transformation excels; elsewhere, its lossy projections expose exactly
what a genuine multimodal joint embedding must be engineered to preserve.%
}

\section{Path 2: Cross-Modal Alignment}
\label{sec:alignment}

To avoid the structural loss of transformation, the second path
keeps each modality's native encoder and stitches their output
spaces together via contrastive pairing.
Alignment earns comparability in terms of the 3C lens,
and scales to many modalities via hub-and-spoke expansion without
pairwise retraining~(\Cref{sec:alignment:hub}),
leading to its domination of cross-modal retrieval today.
However, alignment hits a dual ceiling:
architecturally, separately encoded inputs never interact
inside the model to achieve compositionality;
geometrically, aligned modalities remain segregated
by the modality gap and fail to occupy a truly unified region
in the space~(\Cref{sec:alignment:gap}).

\subsection{The Alignment Paradigm}
\label{sec:alignment:paradigm}

\paragraph{Dual towers plus contrast---and its pre-web prehistory.}
The alignment recipe is architecturally simple: two (or more) independent
encoders, one per modality, trained so that paired items map to nearby vectors
and unpaired ones do not.
The recipe is also old: DeViSE aligned visual embeddings into a word-vector
space with a ranking loss back in 2013~\citep{frome2013devise}, deep CCA
optimized the same correspondence objective with learned nonlinear
projections~\citep{andrew2013deepcca}, VSE++
showed that the crucial ingredient is hard negatives in the ranking
loss~\citep{faghri2018vsepp}, and stacked cross attention (SCAN) added
token-level interaction on top of aligned towers~\citep{lee2018scan}.
On the audio side, SoundNet
learned sound representations by transferring category structure from a
frozen visual classifier into an audio encoder~\citep{aytar2016soundnet},
and L3-Net~\citep{arandjelovic2017look} learned an audio--visual
embedding from raw, unlabeled video by asking nothing more than whether
a frame and a sound co-occur---self-supervised correspondence in place
of the paired web captions the recipe would later run on.
What CLIP~\citep{radford2021clip} changed was therefore neither the dual-tower
architecture nor the contrastive loss~\citep{vandenoord2018cpc} but the
supervision corpus: natural-language supervision at web scale, with ALIGN's
billion noisy pairs~\citep{jia2021align} turning a well-established ranking
objective into a general-purpose representation learner.
CLAP then ported the same recipe wholesale to
audio--language~\citep{clap, laionclap}.
The architecture is the dual-encoder of \Cref{sec:arch:pooling}:
each modality is encoded on its own, and comparison happens only in the shared
space, which is what makes the resulting embeddings usable for efficient
cross-modal nearest-neighbor retrieval at scale.
This is why alignment, more than transformation, dominates production cross-modal
retrieval: it preserves each modality's strong native encoder while providing a
shared index.

\paragraph{The web-scale family: losses, towers, and above all data.}
The recipe has since been refined along three axes that matter for what
follows.
In the \emph{loss}, SigLIP replaces the softmax-normalized contrastive
objective with a pairwise sigmoid loss that does not normalize inside the
batch~\citep{zhai2023siglip}, decoupling quality from huge global batches and
demonstrating robustness at essentially every scale; its successors extend this to
multilingual and dense objectives~\citep{tschannen2025siglip2,zooclaw2026}.
SigLIP towers are now the default visual front end of most open multimodal
LLMs---a quiet but consequential reuse that the convergence trend in
\Cref{part:principles} predicted.
In the \emph{towers}, LiT showed the image tower can be frozen
while only the text encoder trains onto it~\citep{zhai2022lit}---%
the direct ancestor of the frozen-tower composition that
jina-embeddings-v5-omni reprises (\Cref{sec:alignment:gap})---and CoCa showed
a captioning head can be added so that one model serves both contrastive and
generative consumers~\citep{yu2022coca}.
In the \emph{data}, the decisive lever turned out to be curation rather than
architecture: systematic studies of open-data
scaling~\citep{ilharco2021openclip,cherti2023openclip},
explicit curation of already-filtered pools (DataComp;~\citealp{gadre2023datacomp}),
and end-to-end transparent filtering of the raw pool
(MetaCLIP;~\citealp{xu2024metaclip}) each produced sizable gains that
architectural tweaks did not, and EVA-CLIP pushed improved training
techniques to the 18B-parameter regime~\citep{sun2023evaclip,sun2024evaclip18b}.
LLM2CLIP even shows the text tower can be replaced by an LLM that has
seen no image but is only contrastively tuned on caption pairs~\citep{huang2024llm2clip}.
The alignment paradigm, refined for four years, is thus the mature middle of
the field, because its three dials (loss, towers, data) have all
been optimized for web-scale regimes.

\paragraph{The recipe ports to every pair.}
The strongest evidence that alignment is a generalizable method rather than a
vision--language accident is how routinely the CLIP recipe has been
transplanted.
Video: CLIP4Clip adapted CLIP's towers with framewise pooling and temporal
attention for text-to-video retrieval~\citep{luo2022clip4clip}, VideoCLIP
pretrained video--text pairs directly~\citep{videoclip2021}, and
InternVideo2 scaled joint video--text pretraining to billion-parameter
regimes~\citep{wang2024internvideo2}.
Audio: LAION-CLAP~\citep{laionclap} and WavCaps~\citep{mei2024wavcaps} align
general sound and caption embeddings, with WavCaps showing caption-like text
can be manufactured (by ChatGPT) where it does not naturally occur.
3D: ULIP aligns point clouds to the CLIP image--text
space~\citep{xue2023ulip}, OpenShape scales 3D encoders toward open-world
zero-shot recognition over natural-language descriptions~\citep{liu2023openshape},
and Uni3D scales the 3D encoder itself to a billion parameters with
reinstallation of visual and textual priors~\citep{zhou2024uni3d}.
Medicine: ConVIRT aligned radiology images with their
reports~\citep{zhang2022convirt}, MedCLIP showed pairing can be learned from
unpaired image--text data~\citep{wang2022medclip}, BiomedCLIP scaled the
recipe to the full biomedical literature~\citep{zhang2025biomedclip}, and
CONCH pretrained a pathology vision--language model on
image--caption pairs with joint contrastive and captioning
objectives~\citep{lu2024conch}.
Molecules: MoleculeSTM aligns molecular graphs with natural
language~\citep{liu2023moleculestm}, and CLAMP pretrains molecules contrastively
against assay descriptions~\citep{seidl2023clamp}.
Time series: TS-CLIP aligns numerical time-series representations with textual captions
under a CLIP-shaped objective~\citep{tsclip}---the most recent modality family to
adopt this paradigm, and correspondingly the least settled.
The pattern is this paper's reuse axis made concrete: one objective, one
training recipe, and one interface template---paired data plus two
towers---transplanted, unchanged in its essentials, to every modality family
that could assemble a paired corpus.

\paragraph{What alignment buys.}
Alignment achieves true semantic correspondence---searching images with
text, describing audio with language---and it does so while letting each
modality keep its strongest encoder.
It is the foundation on which every current cross-modal retrieval system is
built, and the early omnimodal models such as e5-omni still lean on explicit
cross-modal alignment to map text, image, video, and audio into one
space~\citep{e5omni2026}.

\subsection{The Hub Choice: Which Modality Anchors the Space?}
\label{sec:alignment:hub}

Pairwise alignment does not scale combinatorially, and every multimodal
system in this section therefore confronts the design axis that this
subsection makes explicit: which modality should anchor the shared
space?
The anchor can play three distinct roles: a \emph{hub modality} that every
spoke aligns to; a \emph{coordinate system} defined by a fixed set of
reference items rather than by any encoder at all; or a \emph{frozen tower}
whose geometry new modalities are trained into rather than allowed to
disturb.

\paragraph{Binding into a hub modality.}
That a pretrained contrastive space can be extended with further
modalities was demonstrated early on the audio side: AudioCLIP integrated a
third, audio tower into the CLIP image--text space so that image, text,
and audio embed into one geometry through joint contrastive
training~\citep{guzhov2022audioclip}.
ImageBind turned this one-off extension into a general binding
strategy, the hub-modality role: rather than aligning
every modality to every other, it aligns each modality to a single
hub (image) using naturally available pairs~\citep{imagebind}.
Because alignment to a common anchor is approximately transitive, audio, depth,
thermal, and text all end up in one shared space without ever being explicitly
paired with one another.
VIT-LENS generalizes this toward omnimodal representations through a unified
lens interface~\citep{vitlens}.
The hub, however, is a design axis, and the field's subsequent history can be
read as a debate over the choice of this anchor modality:
LanguageBind answers ``text,'' aligning all modalities to the language
tower~\citep{languagebind}; OmniBind goes pluralist, binding fourteen
pretrained specialist spaces into one routed
space~\citep{wang2024omnibind}, with FreeBind showing the binding itself is a
cheap post-hoc operation on frozen encoders~\citep{wang2024freebind}; and
C-MCR connects separately pretrained contrastive spaces---new modalities
and new datasets alike---into one shared space without retraining the
underlying towers~\citep{wang2023cmcr}.
The trend of the past two years has been largely language-centric, for a reason
\Cref{sec:intro:why} prefigured: language is the only modality in which every
other modality already has a verbalized description, making natural pairs
easily available---even though it remains a lossy hub that
discards whatever was never put into words.
Viewed through our analytical lens (\Cref{sec:framework:lens}), this
hub-and-spoke paradigm neatly balances cross-modal intersection (a
single shared anchor geometry) with modality-specific difference, as
each spoke retains its own native encoder.
Yet whichever hub is chosen, binding leaves a residual offset between the
towers---the geometric remainder that the methods of
\Cref{sec:alignment:gap} explicitly set out to correct.

\paragraph{Stitching: comparability retrofitted on frozen encoders.}
At the lightest end of this path sit methods that retrofit comparability
onto frozen encoders without retraining.
Training-free approaches define shared coordinates through reference items:
relative representations~\citep{moschella2023relative} re-express every
embedding as its similarity to a fixed anchor set,
while ASIF~\citep{norelli2023asif} bridges unimodal models efficiently
through similarity retrieval over paired anchors.
Similarly, model stitching~\citep{bansal2021stitching} reconnects
independently trained networks by optimizing only a single layer between them.
Conceptually, whether spaces are knitted together through training-free reference
anchors or gradient-stitched layers, the fact that
post-hoc alignment succeeds at all provides compelling evidence for the
shared latent geometry.

\subsection{The Limits of Alignment: The Modality Gap}
\label{sec:alignment:gap}

Alignment's ceiling is a striking geometric fact: even after training, the
embeddings of different modalities do not occupy the same region of the shared
space.
\citet{liang2022modalitygap} named this the \emph{modality gap}: in a model
like CLIP, image and text embeddings remain spatially segregated in two separate
cones. This phenomenon persists across modalities (text, natural images,
video, medical images, amino-acid sequences), and even manifests in models
with random weights.

\paragraph{Where the gap comes from.}
The modality gap stems from both architectural dynamics and data asymmetry.
Geometrically, deep neural networks exhibit a ``cone effect'':
the inductive bias of deep networks with nonlinear activations shrinks
the angle between any pair of embeddings, confining each encoder's output
to a narrow cone~\citep{ethayarajh2019anisotropy,liang2022modalitygap}.
And because dual towers start from independent random initializations,
they inhabit separate cones from the outset.
During training, the contrastive loss under a low temperature treats this
separation as a local minimum, while the inherent information imbalance
between dense signals (e.g., images) and sparse captions actively drives
the separation along a handful of dominant embedding
dimensions~\citep{schrodi2025twoeffects}.
Being a joint consequence of initialization, activation geometry,
loss temperature, and cross-modal information asymmetry, the gap
cannot be removed simply by training longer.

\paragraph{Why it matters.}
Because different modalities inhabit separate cones, cross-modal nearest-neighbor search must bridge two distinct distributions, which causes a few items near the cone boundary to absorb disproportionate retrieval traffic, aggravating the high-dimensional hubness
problem~\citep{radovanovic2010hubs,dinu2014hubness,bogolin2022querybank}.
In standard cross-modal retrieval and zero-shot classification where all candidates belong to a single target modality, simply collapsing the gap to zero is not always optimal: the offset modulates logit entropy~\citep{schrodi2025twoeffects}, and shifting the gap distance nontrivially alters downstream accuracy and fairness~\citep{liang2022modalitygap}.
However, the gap turns into an unambiguous obstacle in mixed-modality retrieval where a corpus of mixed text-only, image-only, and multimodal documents causes queries to systematically favor candidates from the same cone~\citep{li2025grclip}, and in generation or captioning tasks where embeddings from different towers must be used interchangeably~\citep{zhang2024c3}.

\paragraph{Mitigation and deeper convergence.}
When cross-modal interchangeability or mixed-corpus search is required,
recent methods close the gap from several angles.
I0T~\citep{an2025i0t} and $C^3$~\citep{zhang2024c3}
standardize aligned embeddings to push toward zero modality gap, and
jina-embeddings-v5-omni preserves the frozen text geometry while adding
modalities through locked towers and trainable connecting components~\citep{jinav5omni}.
Post hoc methods apply training-free per-modality centering,
which boosts NDCG@10 by up to 26 points in GR-CLIP~\citep{li2025grclip}.
More fundamentally, \citet{uniembedgeom2025} demonstrate that a universal latent
geometry underlies distinct text embedding spaces as well as a small number of
cross-modal model pairs, enabling unsupervised vector translation without paired data.
Together with the training-free stitching methods
above~\citep{moschella2023relative,norelli2023asif},
this operationalizes the Platonic Representation Hypothesis~\citep{huh2024platonic}:
if a shared geometry already exists beneath the modality cones, the gap is
merely a surface phenomenon---a coordinate artifact superimposed on the representation,
which motivates architectures that expose the unified space directly through joint fusion.

\paragraph{Alignment is still relevant in the MLLM era.}
One might expect the alignment paradigm to have been obsoleted by
single-stream multimodal LLMs, yet the 2025--26 SOTA says otherwise.
Qwen3-VL-Embedding's retrieval stack keeps an indexable bi-encoder embedding
model paired with, but separate from, an explicit
cross-encoder reranker~\citep{qwen3vlembedding}.
jina-embeddings-v5-omni builds its omni space by locking a frozen text tower
and aligning new modality towers onto its geometry~\citep{jinav5omni}.
Most strikingly, Omni-Embed-Nemotron reports that for audio--video retrieval,
encoding the two streams separately---without the token-level
interleaving that generative omni models use for synchronized
understanding---outperforms the interleaved fusion of its own backbone~\citep{omniembednemotron}.
And LCO-Emb argues the deeper reason: latent cross-modal alignment is
already present inside MLLMs as a by-product of generative pretraining,
so contrastive tuning is a refinement of an existing alignment rather than the
creation of one~\citep{lcoembedding}.
The boundary between ``alignment'' and ``fusion'' is thus blurring into a
question of where inside the pipeline modality-specific streams meet---%
but the design lesson survives: indexable, separately encoded modalities
remain the anchor of most deployed retrieval systems, which is why the next
path must treat alignment as something to subsume, not abolish.

\highlightbox{tred}{Takeaway}{%
Alignment is the pragmatic path that earns comparability: it stitches
independently encoded modalities into a shared retrieval index at scale
without pairwise retraining. Yet it hits a fundamental ceiling.
Contrastive dual encoders inevitably leave modalities segregated in
separate cones---a persistent modality gap driven by initialization
and informational asymmetry. Because inputs are related only by contrastive
pairs rather than joint computation, alignment coordinates spaces without
forming a truly unified semantic object, falling inherently short of
compositionality and necessitating deep joint fusion.%
%
}
\section{Path 3: Multimodal Joint Embedding}
\label{sec:fusion}

While transformation flattens modalities and alignment
compares separately encoded vectors,
joint embedding, our advocated terminus, fuses modalities inside the model
to capture their fine-grained interaction and target compositionality.
This section traces the evolution of the fusion
paradigm~(\Cref{sec:fusion:paradigm}), argues that joint embedding unifies
and surpasses alignment as a more general framework~(\Cref{sec:fusion:joint}),
and distills five design desiderata for overcoming optimization
pathologies in joint embedding models~(\Cref{sec:fusion:desiderata}).
\Cref{tab:fusion-mechanisms} summarizes the spectrum of fusion mechanisms,
ranging from score combination to full token-level interaction.

\subsection{The Fusion Paradigm}
\label{sec:fusion:paradigm}

\paragraph{Which streams attend to which: two kinds of ``fusion''.}
The literature labels both of the following as fusion, and they must be kept
apart.
\emph{(A) Intra-item fusion}: the streams being fused are the modalities
of one item (e.g., an image--text pair that constitutes a query, a
video--audio document that constitutes a candidate) and the output is a
single embedding of that item.  This is what VLM2Vec, GME, MM-Embed, and the
omni-embedders of this section do; it remains a bi-encoder over items,
so items can still be encoded and indexed offline.
\emph{(B) Query--candidate cross-encoding}: the streams being fused are the
query and the candidate themselves, feeding both through one
model, producing a reranking score only and no embedding of the item at all.
The two share the mechanics of cross-attention but not the economics: intra-item
fusion keeps offline indexability at the price of a heavier encoder,
while query--candidate fusion forfeits it and pays a
forward pass per (query, candidate) pair.
In this paper, ``joint embedding'' always refers to intra-item fusion,
while the other is only mentioned (explicitly as ``rerankers'') when
deployed stacks pair both in pursuit of accuracy,
with an indexable intra-item-fusion embedding first and
a cross-encoder reranker over its top-$k$ list.

\paragraph{From vision--language pretraining to fusion.}
The fusion paradigm predates MLLMs.
Early vision--language pretraining explored both multi-stream architectures
that process the two modalities in separate streams and combine them in
co-attention layers~\citep{lu2019vilbert, tan2019lxmert}, and
single-stream architectures that build one Transformer over concatenated
visual and textual token streams~\citep{chen2020uniter,
kim2021vilt}, with FLAVA and BEiT-3 later unifying image, text, and
image--text objectives at scale~\citep{singh2022flava, wang2023beit3}.
Pretraining objectives varied across these models: masked multimodal
modeling and alignment prediction in ViLBERT, and LXMERT's five tasks from
masked language modeling to cross-modality matching and image question
answering.
The pairing that would become standard---a dual-encoder image--text
contrastive (ITC) head plus an image--text matching
(ITM) cross-encoder head---was crystallized by ALBEF, whose recipe is
literally ``align before fuse'': the ITC loss acts on unimodal
representations before fusion, jointly optimized with ITM and
masked language modeling~\citep{li2021albef}; BLIP bootstraps both from
synthetic captions~\citep{li2022blip}, and BLIP-2 projects the Q-Former's
output query embeddings into the input space of a
frozen language model as soft visual prompts~\citep{li2023blip2}, which
later led to the MLLM embedders below.
This design has since come under audit: DeCo argues that
compressive projectors such as the Q-Former abstract the visual stream
prematurely, and that compressing patch tokens while leaving semantic abstraction
to the decoder is both cheaper and better~\citep{yao2024deco}, while
Cross-Layer Injection replaces the single static connection with a dynamic
bridge through which the decoder draws on every vision layer on
demand~\citep{chen2026cli}.
This early pretraining era is thus, in retrospect, where the field
discovered that alignment and interaction are two operating modes of
one architecture, a discovery central to \Cref{sec:fusion:joint}.  In
audio--visual learning the same arc runs from VATT~\citep{akbari2021vatt} to
attention bottlenecks, which show that restricting cross-modal attention to a
few tokens both regularizes and improves fusion~\citep{nagrani2021mbt}.

\begin{table*}[t]
\centering
\small
\caption{Ways to combine inputs, ordered by how early and at what resolution the
inputs interact. Letting inputs interact earlier and at a finer resolution can
preserve more detail, but leaves less computation to cache: the deeper the interaction, the more capability shifts
from comparability toward compositionality.}
\label{tab:fusion-mechanisms}
\adjustbox{width=\textwidth,center}{
\rowcolors{2}{gray!10}{white}
\begin{tabular}{@{}m{0.14\textwidth}m{0.195\textwidth}m{0.285\textwidth}m{0.19\textwidth}m{0.2\textwidth}@{}}
\toprule
\textbf{Mechanism} & \textbf{Construction} & \textbf{Representative system} & \textbf{Strength} & \textbf{Main trade-off} \\
\midrule
Score fusion & $s=\sum_m w_m s_m$ & Late-fusion rerankers; UniIR score-fusion variants~\citep{wei2024uniir} & Reuses unimodal indices; handles missing inputs & May not produce a combined embedding \\
Vector fusion & $\ve=\operatorname{norm}(\sum_m w_m \ve_m)$ & Additive cross-modal baselines & Low-cost, cacheable, modular & Fixed addition cannot model correspondence or conflict \\
Learned composition & $\ve=h(\ve_1,\ldots,\ve_k,r)$ & TIRG~\citep{vo2019tirg} (gated composition of global image and text features) & Models interactions among input vectors at low cost & Compresses local evidence before interaction \\
Pseudo-token composition & Translate one modality into another token space & VISTA~\citep{zhou2024vista} (visual tokens injected into a text encoder) & Reuses a pretrained model's ability to combine tokens & Loses information the target token space cannot express \\
Token/feature fusion & $\ve=\operatorname{pool}(F(\mH_{1:k},r))$ & Everything at Once~\citep{shvetsova2022everything}, VLM2Vec~\citep{jiang2025vlm2vec}, UniIR feature-fusion variants~\citep{wei2024uniir}, LCO-Embedding~\citep{lcoembedding}, MM-Embed~\citep{lin2025mmembed} & Models fine-grained interaction before pooling & Requires re-encoding; cost grows with tokens \\
Multi-part output & Global plus local/input-specific components & Multi-vector / late-interaction retrieval~\citep{faysse2024colpali} & Supports retrieval, attribution and fallback & More complex storage, APIs and evaluation \\
\bottomrule
\end{tabular}
}
\end{table*}

\paragraph{Cross-attention as deep interaction.}
Where alignment uses an independent dual-encoder, intra-item fusion uses
cross-attention: queries from one modality attend to keys and values from
another, so that each modality's representation is conditioned on the other's
content.
Recall from \Cref{sec:arch:transformer} that self-attention is a
permutation-equivariant all-to-all operator; cross-attention is its
bipartite counterpart, a universal relational operator between two
token sets.
Where contrastive alignment learns only that a pair ``corresponds,'' fusion
learns how they correspond---which parts of a caption describe which
regions of an image, which audio moment matches which frame~\citep{shvetsova2022everything}.
For symmetric multimodal-to-multimodal retrieval, SOLAR learns an intersection
mask over each fused image--text pair so that one joint vector keeps both the
shared semantics and the complementary difference, showing that the
interaction can reach into the construction of the embedding
itself~\citep{yang2026solar}.
The interaction is not free: attention is quadratic in the token counts of the
streams being combined, so fused encoders are strictly more expensive than
their separately encoded parts---but the encoding stays per item and
therefore remains pre-computable.

\paragraph{From fusion to omnimodal embedders.}

The fusion paradigm now drives a wave of multimodal embedding models that repurpose
generative MLLMs to treat embeddings as a first-class output rather than the byproduct
of two aligned towers.
Pioneered by VISTA injecting visual tokens into text encoders~\citep{zhou2024vista},
Everything at Once fusing multimodal transformers for video
retrieval~\citep{shvetsova2022everything},
and VLM2Vec~\citep{jiang2025vlm2vec} introducing instruction conditioning into
multimodal representations,
recent MLLM embedders such as MM-Embed~\citep{lin2025mmembed},
UniME~\citep{gu2025unime}, VLM2Vec-V2~\citep{meng2025vlm2vecv2}, GME~\citep{zhang2025gme},
and DME~\citep{chen2026dme} have converged on several core design improvements:

\begin{itemize}[leftmargin=2em]
  \item \textbf{Architectural adaptation and latent alignment.}
    Because causal MLLMs are suboptimal encoders out of the box, MoCa restores
    bidirectional attention via modality-aware continual pretraining~\citep{chen2025moca},
    while LCO-Emb prepends generative pretraining before lightweight
    contrastive tuning~\citep{lcoembedding} to raise the representation ceiling.
    Strikingly, E5-V~\citep{jiang2024e5v} and LCO-Emb~\citep{lcoembedding} show
    that finetuning an MLLM on text pairs alone generalizes zero-shot to image and
    composed retrieval, proving that cross-modal alignment already exists
    latently inside generative MLLMs.
  \item \textbf{Synthetic data and modality-aware hard negatives.}
    To overcome the scarcity of complex multimodal retrieval pairs, recent works
    such as mmE5 and MegaPairs synthesize massive instruction-conditioned triplets and
    hard negatives~\citep{chen2025mme5, zhou2024megapairs}, while LLaVE applies
    hardness-weighted contrastive losses~\citep{lan2025llave}, and UniME-V2 leverages
    MLLM-as-a-judge to refine query-candidate relevance~\citep{gu2026unimev2}.
  \item \textbf{Reranker integration.}
    Because cross-encoder rerankers provide accuracy that a precomputable embedding
    must leave behind, production stacks increasingly couple them with bi-encoder
    embedders, either as dual LoRA modules on one MLLM in LamRA~\citep{liu2025lamra},
    via reranker-to-embedder distillation in Qwen3-VL-Embedding~\citep{qwen3vlembedding},
    or by training reasoning rerankers with reinforcement learning in
    MM-R5~\citep{xu2025mmr5}.
  \item \textbf{Reasoning-driven embedding.}
    Pushing the representation--generation unification of \Cref{sec:objectives} to its
    limit, a latest fork of the fusion path generates explicit reasoning rationales
    before reading out the embedding, represented by ReasonEmbed \& O1 Embedder in
    text~\citep{reasonembed2025,letretrieversthink2026}
    and Think-Then-Embed in multimodal domains~\citep{cui2025thinkthenembed}.
    UME-R1 further optimizes the model with reinforcement learning over retrieval
    rewards~\citep{lan2025umer1}, while Retrv-R1 balances reasoning gains against
    latency costs with thinking-mode schedules~\citep{zhu2025retrvr1}.
  \item \textbf{Composed and multi-vector readouts.}
    When queries are themselves multimodal, fusion is the task rather than the
    architecture. This is operationalized by the composed reference images and
    modification instructions in CIRR and FashionIQ~\citep{liu2021cirr, wu2021fashioniq}.
    Targeting such scenarios, TIRG composes the queries within a Siamese
    network~\citep{vo2019tirg}, while Pic2Word and SEARLE learn textual
    inversion, mapping images into the word embedding space of a
    frozen CLIP without training the image and text
    encoders~\citep{saito2023pic2word, baldrati2023searle}.
    More recently, MagicLens and CoVR have extended the instructions to open-ended
    domains by constructing self-supervised query-instruction-target triplets for
    images and videos from web data pairs~\citep{zhang2024magiclens, ventura2024covr}.
    Meanwhile, on the output end, multi-vector readouts such as
    ColPali~\citep{faysse2024colpali} and jina-embeddings-v4~\citep{guenther2025jinav4}
    preserve fine-grained patch correspondence and allow late interaction,
    and MetaEmbed even enables test-time scaling of interaction
    vectors~\citep{xiao2025metaembed}, turning the readout lattice of \Cref{sec:data}
    into an inference-time budget choice alongside Matryoshka width
    truncations~\citep{kusupati2022mrl}.
\end{itemize}

Armed with these techniques, the frontier has expanded beyond vision--language
to full omnimodal encoders that jointly ingest text, image, video, and audio
within a single backbone rather than an aligned set of towers.
Recent realizations explore distinct architectural recipes for this expansion:
e5-omni combines explicit cross-modal alignment with token-level
fusion~\citep{e5omni2026}, WAVE targets general audio--visual and temporal
correspondence~\citep{wave2025}, Omni-Embed-Nemotron processes audio and video
as deliberately non-interleaved streams~\citep{omniembednemotron}, and
Ovis-Embedding scales unified omni-modal training to lead the MMEB-V3 benchmark
in late 2026~\citep{ovisemb}.
Fusion, in short, is no longer a specialized architecture for a particular
bimodal task; it has become the default paradigm for building universal
omnimodal embeddings.

\subsection{Joint Embedding as a General and Ultimate Framework}
\label{sec:fusion:joint}

We now make the argument that elevates fusion from ``one path among three''
to the terminus of the journey.

\paragraph{The core insight: fusion subsumes alignment.}
The key observation is that a joint-embedding (fusion) model is a
strict generalization of an alignment model.
A dual-encoder alignment model is simply a joint-embedding model in which the
cross-modal interaction has been set to zero---an architecture that fuses
nothing and falls back to comparing independently encoded vectors.
Conversely, exposing the interaction (cross-attention) can only add
information the alignment model lacks: the relative arrangement of tokens
across modalities that no contrastive pair can capture.
The vision-language pretraining era supplies the cleanest architectural
witness: ALBEF and BLIP carried both regimes on one backbone---an
image--text contrastive (alignment) head and an image--text matching
(interaction) head---and the interaction side won precisely on the tasks
where correspondence matters~\citep{li2021albef, li2022blip}, so
interaction behaves as a dial on one architecture rather than a separate
paradigm.
This observation, together with the theory of partial information
decomposition~\citep{liang2023quantifying}, motivates our ``compositional gain''
argument: a genuinely joint vector should surpass the strongest unimodal encoder
as well as score- and vector-addition baselines by capturing cross-modal
interactions, which we formalize as $\Delta_{\mathrm{comp}}$ in \Cref{sec:evaluation}.
By this logic we conjecture that in the capacity-unconstrained limit,
fusion weakly dominates alignment: there would be no regime in
which the architecture is at a principled disadvantage, only regimes in which
fusion's interaction term is unnecessary and its compute cost is wasted, in
which case the two coincide.
The limit is also where the conjecture must stop: model capacity, data mixture,
and modality imbalance each break it in practice~\citep{omniembednemotron,
lin2025mmembed,qwen3vlembedding, huang2026mmebv3}.
Whether fusion's extra capacity pays off at a given compute and data budget is
therefore an empirical question, and is what the C3 compositional-gain
controls of \Cref{sec:evaluation} are designed to test.

\paragraph{One model, two operating modes.}
This generalization yields a striking consequence.
A single joint-embedding model has two operating modes determined by the input:
\begin{itemize}[leftmargin=2em]
  \item \textbf{Multimodal in $\Rightarrow$ fusion.}
    When the input spans modalities, the streams interact through explicit
    cross-attention or the shared self-attention of a single-stream backbone over
    concatenated tokens, and the model produces a representation of their interaction.
  \item \textbf{Unimodal in $\Rightarrow$ a stronger single-/cross-modal model.}
    When only one modality is presented, single-stream backbones degrade
    gracefully to within-modality attention, while
    explicit cross-attention designs need a defined policy for
    the missing stream, such as skipping or substituting the missing keys and values.
    Under either design, the same parameters can serve as a strong unimodal
    encoder that benefits from the auxiliary signal and richer supervision of
    cross-modal training.
\end{itemize}

In the capacity-unconstrained limit, the model forfeits no capability
by being general.
This is the operational reading of the Platonic convergence~\citep{huh2024platonic}
and the universal-geometry result~\citep{uniembedgeom2025}: if a single
geometry underlies all modalities, a single model trained to expose it should
be at least as good on any one modality as a specialist, and strictly better on
their combination.
E5-V's text-only training is this prediction run in reverse: finetuned on
text pairs alone, the model transfers to image and composed-image
retrieval~\citep{jiang2024e5v}.

\paragraph{Why this is the terminus.}
The three paths therefore form a progression rather than parallel options.
Transformation homogenizes inputs at the cost of structural loss;
alignment shares a metric space but hits the geometric ceiling of the modality gap;
and joint embedding represents the terminus, computing within a unified
space to preserve modality-specific structure while capturing cross-modal
interaction.
Notably, this terminus is not merely an engineering convenience, but mirrors
the hub-and-spoke organization of human semantic memory, where a
modality-invariant convergence zone~\citep{damasio1989timelocked,patterson2007where}
integrates multisensory cues within a single reliability-weighted
computation~\citep{ernst2002humans}.

In practice, however, naive fusion is neither free of computational cost
nor immune to optimization pathologies: at finite capacity, joint training
can trigger modality collapse, gradient interference, and unimodal degradation.
Turning joint embedding into a reliable terminus requires confronting these
open costs directly, which we next formalize as five concrete design desiderata.

\subsection{Desiderata for a Multimodal Joint Embedding Model}
\label{sec:fusion:desiderata}

If joint embedding is to be the terminus, progress toward it needs a
definition: what separates an effective joint embedding model from a naively
fused one?
We distill the lessons from the failure modes of joint training and
the remedies that resolve them into five explicit design criteria,
turning the framework argued above into a concrete checklist that any
claimed joint embedding can be held to, and giving \Cref{sec:evaluation}
testable properties to measure.

\begin{itemize}[leftmargin=2em]
  \item \textbf{Preserve each modality's own interface and parameters
  to prevent collapse and competition.} 
    Fusion is not a license to homogenize the periphery.
    Sharing all feed-forward pathways across modalities triggers
    gradient competition~\citep{peng2022ogm} and
    modality collapse, where latent representations ignore weaker
    modalities~\citep{javaloy2022modalitycollapse}.
    This erases the modality-unique signals formalized by partial
    information decomposition~\citep{liang2023quantifying}
    and results in modality competition, where joint training provably
    underperforms unimodal training on every
    constituent modality~\citep{huang2021makes,huang2023modalitycompetition}.
    The architectural remedy is to retain modality-specific
    tokenizers~(\Cref{sec:arch:transformer}) and allocate dedicated
    expert capacity, as in VLMo's Mixture-of-Modality-Experts~\citep{bao2022vlmo}
    and MoE-LLaVA's sparsely activated experts~\citep{moellava2024}.
  \item \textbf{Let each modality deepen before fusion, and bound the interaction cost.} 
    Fusion should be a late-stage act.
    Mixing streams in shallow layers averages away modality-specific
    structure before it can form, resulting in the modality collapse warned above.
    Layerwise analyses such as DeepInsert confirm that deferring cross-modal
    interaction to middle and deep layers allows multimodal tokens to bypass
    early language layers with negligible loss~\citep{deepinsert2025}.
    This schedule is also mirrored by training recipes that stage unimodal or generative
    pretraining before cross-modal fusion: LCO-Emb raises the representation ceiling
    through continual generative pretraining before contrastive
    finetuning~\citep{lcoembedding}, e5-omni couples an
    explicit cross-modal alignment stage with fusion~\citep{e5omni2026}, and
    VLMo trains its experts on image-only and text-only data before image-text
    pairs~\citep{bao2022vlmo}.
    Furthermore, to keep fusion computationally tractable under the quadratic cost
    of attention upon cross-modal interaction,
    sequence lengths should be bounded beforehand via compressive projectors or
    attention bottlenecks~\citep{yao2024deco,nagrani2021mbt}.
  \item \textbf{Fuse for composition, not just consensus.}
    The point of joint embedding is capturing the synergistic cross-modal
    interaction, so the model must be optimized and held to it.
    Standard contrastive losses optimize only for cross-modal
    consensus---preserving shared redundancy while failing to remove
    modality-specific nuisance or protect modality-unique and synergistic
    signals~\citep{liang2023quantifying,liang2023factorcl,almudevar2025ib},
    which is acceptable for dual-tower alignment but insufficient for fusion.
    A joint model therefore earns its name only when its fused representation
    strictly outperforms the strongest unimodal encoder as well as additive
    score- and vector-fusion baselines---a requirement we operationalize as
    the compositional gain $\Delta_{\mathrm{comp}}$ in \Cref{sec:evaluation}.
  \item \textbf{Audit and enforce the unimodal mode; do not assume it.}
    At finite capacity, joint multimodal training routinely induces negative transfer
    on single-modality tasks, as seen in the text retrieval degradation of both
    MM-Embed~\citep{lin2025mmembed} and Qwen3-VL-Embedding~\citep{qwen3vlembedding}.
    Unimodal parity must therefore be actively enforced through multi-stage curricula,
    balanced data mixtures, or knowledge distillation.
  \item \textbf{Treat absent streams and flexible readouts as first-class design choices.}
    Deployed omni-encoders routinely encounter incomplete modality
    subsets (e.g., text-only queries against video--audio candidates) and
    diverse granularity demands~(\Cref{sec:data}).
    Rather than relying on ad hoc defaults or blindly inheriting a generative
    backbone's token-interleaving format~\citep{omniembednemotron}, a robust
    joint embedder must explicitly define its fallback policy for missing streams
    and expose flexible readouts across the granularity lattice to
    match downstream task demands.    
\end{itemize}

The five criteria compound: the first two say where modality-specific
structure should live (the periphery, the early depth), the last three say
what the fusion itself must deliver (composition, parity, robustness), and
all five are testable---making them the natural complement to the 3C lens
operationalized in \Cref{sec:evaluation}.

\begin{landscape}
\begin{figure*}[t]
\centering
\adjustbox{center}{%
\includegraphics[width=1.5\textwidth]{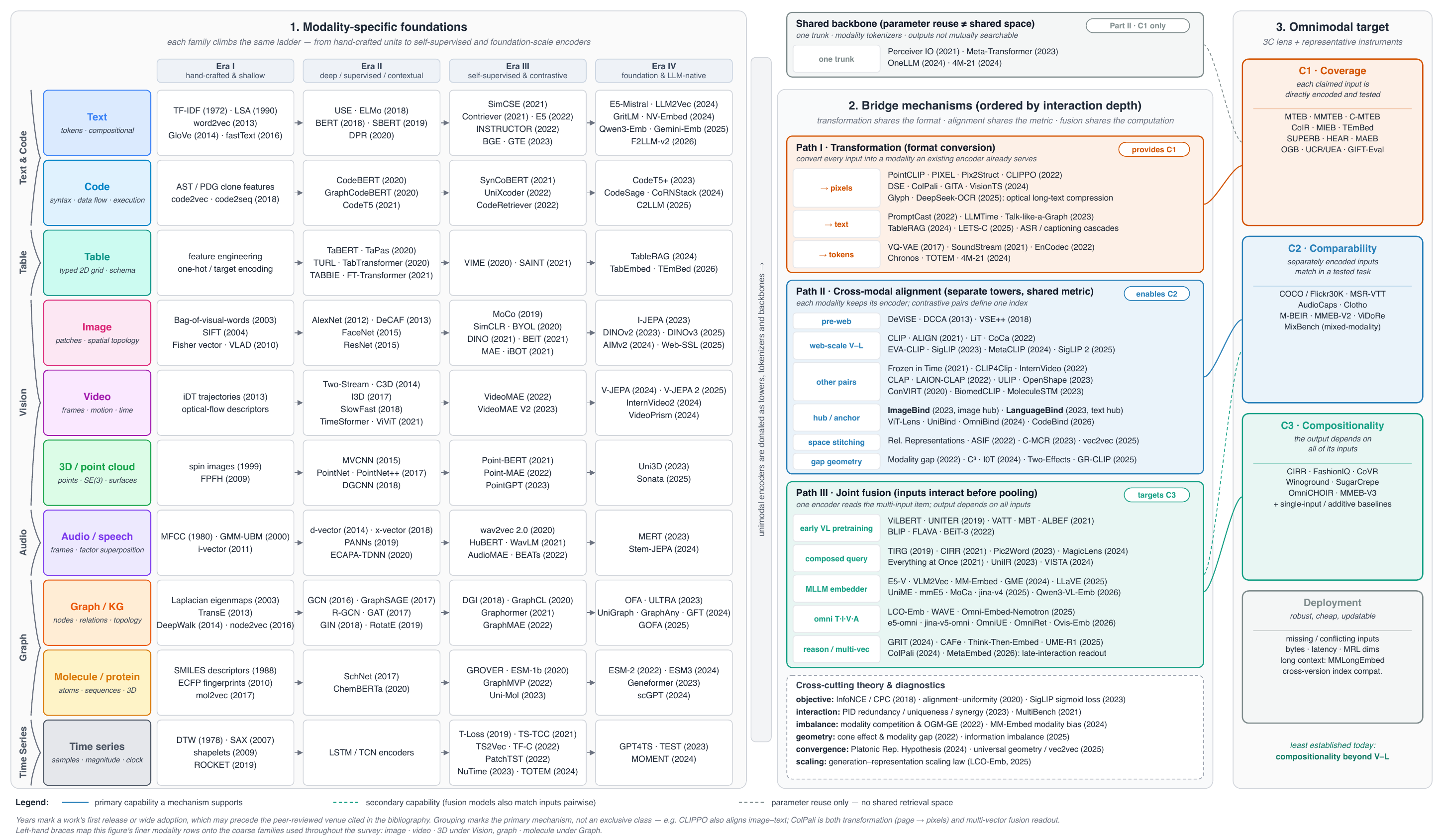}%
}
\caption{From modality-specific foundations to omnimodal embedding.
\textbf{Left:} ten modality families each climb the same four-era
ladder---hand-crafted units, deep supervised/contextual encoders,
self-supervised and contrastive pretraining, and foundation-scale or
LLM-native encoders---which is the per-modality face of the convergence
argued in \Cref{part:principles}.
\textbf{Middle:} shared backbone (\Cref{sec:arch}) and the three \Cref{part:path} bridges, ordered by interaction depth:
transformation (\Cref{sec:transformation}) converts inputs into a format an existing
encoder already serves; alignment (\Cref{sec:alignment}) keeps separate
towers and learns a shared metric; joint fusion (\Cref{sec:fusion}) lets
inputs interact before pooling.
\textbf{Right:} the 3C lens with representative instruments for each
capability (\Cref{sec:evaluation}).
Solid arrows mark the capability a mechanism most directly supports;
dashed arrows mark secondary support.
Groupings indicate the primary mechanism, not an exclusive class.}
\label{fig:modality-trajectories}
\end{figure*}
\end{landscape}

\paragraph{From mechanism to trajectory.}
\Cref{fig:cross-modal-paradigm-map} opened \Cref{part:path} by abstracting the three
bridge mechanisms in isolation; \Cref{fig:modality-trajectories} closes it by
situating those mechanisms against the per-modality histories that produced
them---from modality-specific foundations, through the transformation and
alignment bridges, to the joint-fusion models and the three 3C requirements
they must ultimately answer.
Where \Cref{fig:cross-modal-paradigm-map} fixed the mechanisms,
\Cref{fig:modality-trajectories} fixes the
milestones: taken together they frame the progression we argue the
field is in the middle of, from separate encoders to a single fusion model
that inherits them all.

\highlightbox{tred}{Takeaway}{%
Joint embedding unifies and surpasses alignment.
It is the omnimodal embedding model in its ultimate form: a single encoder
whose representational object is the joint semantics of whatever modalities it
is given, that reduces to a strong specialist when given a single modality---a
level that, at finite capacity, the training recipe must actively
maintain---and that closes the modality gap at its source
when paired with appropriate objectives and prompting.
Reaching this end, however, demands that we stop ignoring modality-specific
structure and instead learn to preserve and even borrow it---the subject of
\Cref{part:road}.%
}

\clearpage
\part{The Road Ahead}
\label{part:road}
\section{Frontier I: Mastering Modality-Specific Properties}
\label{sec:frontier:properties}

To build a joint embedding that truly fuses rather than collapses, we must
first respect what is modality-specific---the ``difference'' axis of our
analytical lens.
The convergence of \Cref{part:principles} tells us the encoders and objectives can be shared;
the difference axis tells us what each modality demands the shared
machinery preserve or invent.
A joint embedding model that ignores these properties discards the very
information that made each modality worth encoding.
This section inventories the defining difficulty of each modality and the
methods it has forced into being.
\Cref{tab:modality-ontology} frames the inventory as a practical input
taxonomy: for each family, which structure must be preserved and the
representative evidence, tagged by whether the input is a raw observation,
a derived view, a structured view, or a compound format.

\begin{table*}[t]
\centering
\fontsize{8.0}{9.4}\selectfont
\setlength{\tabcolsep}{3.5pt}
\renewcommand{\arraystretch}{1.02}
\rowcolors{2}{gray!10}{white}
\caption{A practical input taxonomy for the modality-specific challenges of
\Cref{sec:frontier:properties}. R denotes a raw observation, D a derived view,
S a structured view, and C a compound format. Tensor shape does not define a
modality: RGB and thermal are both arrays but measure different quantities. A
domain label such as ``medical'' is not itself a modality.}
\label{tab:modality-ontology}
\begin{tabular}{@{}m{0.18\textwidth}m{0.26\textwidth}m{0.28\textwidth}m{0.22\textwidth}@{}}
\toprule
\textbf{Family} & \textbf{Examples and type} & \textbf{Structure to preserve} & \textbf{Representative evidence} \\
\midrule
Language and symbols & Natural language and code (R); tables and markup (S) & Discrete sequence, hierarchy and references; code adds syntax, flow and execution & BERT, CodeBERT, GraphCodeBERT, Meta-Transformer~\citep{devlin2019bert,codebert2020,graphcodebert2021,metatransformer2023} \\
Visible and dynamic vision & RGB image, video and event streams (R); flow, segmentation, keypoints and boxes (D) & 2D topology; video/events add order, rate, sparse time and motion & VideoMAE, UniBind, 4M-21~\citep{tong2022videomae,lyu2024unibind,bachmann20244m21} \\
Geometry and 3D & Depth, LiDAR (R/D); point sets, meshes, voxels, poses and fields (S/D) & Metric coordinates, surface topology and SE(3) behavior & ImageBind, ViT-Lens, PointNet++, MeshCNN~\citep{imagebind,vitlens,qi2017pointnetpp,hanocka2019meshcnn} \\
Spectral and ranging sensors & Thermal, hyperspectral, SAR, radar, sonar, location streams (R) & Band identity, calibration, sparsity and sensor noise & SatMAE, BEVFusion~\citep{cong2022satmae,liu2023bevfusion} \\
Audio and acoustics & Speech, music, environmental sound, waveforms (R) & Sampling rate, phase, mixtures, content, speaker and environment & wav2vec 2.0, CLAP, MAEB~\citep{baevski2020wav2vec2,clap,maeb} \\
Embodied and physiological sensing & IMU, force, tactile, EEG, ECG, EMG, PPG (R) & Body/sensor topology, units, synchronization, privacy & ImageBind, CodeBind, MultiBench~\citep{imagebind,chen2026codebind,liang2021multibench} \\
Biomedical images and records & X-ray, CT, MRI, pathology (R); notes, labs, EHR events (R/C) & Anatomy, protocol, patient time, missingness, provenance & BiomedCLIP, RETFound~\citep{zhang2025biomedclip,zhou2023retfound} \\
Temporal and event data & Regular/irregular series, trajectories, point processes, logs (R/S) & Timestamps, magnitude, units, variable identity, event marks & TS2Vec, PatchTST, RMTPP~\citep{ts2vec,nie2023times,rmtpp} \\
Relational and structured data & Graphs, knowledge graphs, hypergraphs, relational tables (S) & Permutation symmetry, typed edges, schema, hierarchy, cardinality & GraphSAGE, Graphormer, Meta-Transformer~\citep{hamilton2017inductive,ying2021do,metatransformer2023} \\
Scientific sequences and measurements & DNA/RNA, proteins, omics, molecules, spectra (R/S) & Alphabet and long-range structure, assay context, physical units & Protein LM, Geneformer, MoleculeSTM~\citep{rives2021protein,theodoris2023geneformer,liu2023moleculestm} \\
Compound documents and interfaces & Visual documents, slides, charts, web pages, GUIs, maps (C) & Text, pixels, layout, reading order, interaction, provenance & ColPali, MMEB-style deployment tasks~\citep{faysse2024colpali} \\
\bottomrule
\end{tabular}
\end{table*}

\paragraph{Text and code: discreteness and compositionality.}
Text is inherently discrete, comprising a finite vocabulary governed by
syntax, and its semantics are compositional.
This is both an enabler and a burden.
Discreteness makes text amenable to tokenization and gives it an attested,
recurring unit inventory of words and subwords, which is why MLM, contrastive
disambiguation, and skip-gram-style objectives all work~\citep{mikolov2013distributed,
devlin2019bert, gao2021simcse}.
But compositionality breeds massive polysemy (e.g., ``bank'' denotes a financial
institution, a river edge, or a verb), and the contextual representations
that must resolve it tend to be anisotropic, collapsing into a narrow
cone~\citep{ethayarajh2019anisotropy};
contrastive finetuning is in part a mechanism for disambiguating the embedding
space.
Code is text but not just text: it adds executable semantics (two
functions may be syntactically dissimilar but semantically
equivalent;~\citealp{fuzzpretrain2024}) and cross-language
equivalence (implementations
of an algorithm in Python and Java share semantics despite divergent
lexicon;~\citealp{langagnostic2024}), and it carries a hierarchical
structure of tokens, expressions, statements, functions, modules, and
projects that current models flatten rather than
exploit~\citep{jin2023edgeformers, clever2025}.
The code encoders' own history shows how these layers were reclaimed one at
a time, progressing from surface tokens~\citep{codebert2020}, AST syntax, and
data-flow semantics~\citep{graphcodebert2021} to the compiler's language-independent
intermediate representation (IR;~\citealp{ir2vec2020,irgen2022}), which provides
evidence that compiler-transformed views are a productive axis of invariance a
joint embedding can declare.
A joint embedding must capture compositional meaning without losing
disambiguation, and must learn to represent code's behavioral equivalence, not
just its surface form~\citep{coreb2026}.

\paragraph{Vision: spatial topology and multi-scale composition.}
Vision has no canonical counterpart to word2vec: a pixel at coordinate $(i,j)$
is not the same entity across images, patch boundaries are imposed by
preprocessing, and object identity must be inferred across scale, viewpoint,
deformation, and occlusion.
Continuity is not the obstacle; choosing stable recurring units and a
meaningful context relation across scenes and resolutions
is~\citep{sivic2003videogoogle, vandenOord2017vqvae}.
Visual semantics are spatial, compositional, and multi-scale: edges
compose into textures and objects, and the same entity produces different
measurements under translation, illumination, and occlusion, while multiple
entities coexist in one frame.
The representation must therefore choose between invariance (recognition may
discard position and texture) and equivariance (correspondence, geometry, and
generation require them)~\citep{long2015fcn, caron2021dino}.
Video adds a temporal axis and a hierarchy of units (frame, clip, track,
event), so a strong image encoder is a necessary appearance model but not a
complete video representation---motion must be retained where retrieval is
motion-sensitive~\citep{tong2022videomae, devaraj2024motion}.
A joint embedding must preserve this spatial and temporal correspondence, not
collapse a scene to a single global point.

\paragraph{Audio: factor superposition.}
What distinguishes audio is not its geometry but its \emph{superposition of
factors}.
A three-second utterance simultaneously encodes what was said (content),
who said it (speaker), how they said it (prosody and emotion), in what
language, in what room, through what microphone, and increasingly the speaker's
health and affective state.
These factors are not a hierarchy of abstraction like edges, textures, and
objects in vision, but distinct factors coexisting in one waveform.
Different tasks require different factors preserved and the rest destroyed,
and whether a given factor is recoverable at all depends on the recording,
the supervision, and the evaluation protocol.
The defining consequence is that there is no such thing as the embedding
of an audio clip: a request to embed audio is incomplete until the factor is
named~\citep{xvector, baevski2020wav2vec2, ecapa}, which is why the SUPERB
protocol scores a learned weighted sum over layers rather than the final
one---the representation is not a layer but a choice over layers.
\Cref{tab:challenge-audio-factors} makes this concrete: it maps each audio
factor to a representative encoder family, the tasks that keep it, and the
metrics that score it, exposing the asymmetry that success in one row may
require suppressing information rewarded in another.
Audio confronted the task-dependence of similarity two decades before the other
modalities, and necessitated a distinct family of objectives that
certify a factor is absent---speaker removal, anonymization, and
privacy-preserving speech representations~\citep{voiceanon, etawavlm, privsem,
wavshape}---together with tokenizers that separate a content stream from an
acoustic one, so that a downstream generation model can keep the words while
changing the speaker~\citep{zhang2024speechtokenizer, defossez2024moshi}.
A joint embedding must name the factor it is encoding, or preserve them as a
superposition it can separate on demand rather than a single entangled point.
\begin{table*}[t]
\centering
\small
\rowcolors{2}{gray!10}{white}
\caption{Factors superposed in an audio signal, representative encoder families,
and factor-specific evaluation. The rows are task-conditioned views of the same
input rather than mutually exclusive latent variables. The table is the
empirical form of the audio-specific ``difference'' (\Cref{sec:frontier:properties}):
success in one row may require suppressing information rewarded in another---so
``the audio embedding'' is undefined until the factor is named.}
\label{tab:challenge-audio-factors}
\adjustbox{width=\textwidth,center}{
\begin{tabular}{@{}m{0.14\textwidth}m{0.28\textwidth}m{0.30\textwidth}m{0.22\textwidth}@{}}
\toprule
\textbf{Factor} & \textbf{Representative encoder} & \textbf{Kept by} & \textbf{Scored by} \\
\midrule
Linguistic content
  & SSL frame features~\citep{baevski2020wav2vec2,hsu2021hubert,wavlm};
    semantic codec tokens~\citep{repcodec}
  & Recognition, translation, keyword spotting, voice activity detection
  & Word/phone error rate for recognition; accuracy for keyword spotting \\
\addlinespace
Speaker identity
  & i-vector~\citep{ivector}, x-vector~\citep{xvector},
    ECAPA-TDNN~\citep{ecapa}, content-disentangled speaker
    embeddings~\citep{seqdisent}
  & Verification, diarization, target-speaker extraction, voice cloning
  & EER/minDCF for verification; DER for diarization \\
\addlinespace
Prosody \& emotion
  & Prosody/timbre embeddings~\citep{prosodytimbre}; emotion
    representations~\citep{emotion2vec}
  & Expressive synthesis, affect recognition, punctuation restoration
  & Classification accuracy, listener studies \\
\addlinespace
Language \& accent
  & Language embeddings~\citep{disentmultiling}
  & Language identification, multilingual routing
  & Identification accuracy \\
\addlinespace
Channel \& environment
  & Session or environment embeddings~\citep{envagnostic}
  & Enhancement and dereverberation; removed for robust verification
  & Downstream robustness \\
\addlinespace
Acoustic detail
  & Neural codec residuals~\citep{soundstream,defossez2022encodec}
  & Reconstruction and generative decoding
  & Perceptual quality, bitrate \\
\bottomrule
\end{tabular}
}
\end{table*}

\paragraph{Graph: explicit relational structure and non-Euclidean geometry.}
Graph embeddings differ from text, vision, audio, and time series because
relations are part of the represented object, not a side effect of an
ordered signal~\citep{bronstein2021geometric}.
A graph object begins from entities plus relations---edges, paths, types,
weights, directions, hyperedges, temporal links, schema constraints, or
higher-order substructures---and the embedding must preserve task-relevant
relations while remaining insensitive to arbitrary node naming, i.e.,
permutation equivariant.
Graph specificity may reside in arity, curvature, invariance, dynamics, or
topology rather than in vectors alone: higher-arity knowledge beyond binary
edges~\citep{xiong2023shrinking, kim2024hypeboy}, non-Euclidean
(hyperbolic/Lorentz) geometry for hierarchy and
curvature~\citep{lin2023hyperbolic}, and sign/basis-invariant spectral
embeddings so that a comparison is not an artifact of node
ordering~\citep{lim2023signbasis, ma2023laplacian}.
The same encoder can be trained under contrastive, masked, or proximity
objectives, but the learning signal is whatever operation preserves the
relational unit~\citep{chen2024tokenized_graph_transformer_contrastive,
yang2023vqgraph}, and a Graph Transformer is a graph mechanism only when
relational structure is explicitly encoded into its tokenization, attention
biases, or positional signal~\citep{yuan2026survey}.
A joint embedding must treat relations as a representational object to be
preserved, not a context to be projected away.

\paragraph{Tables: grid geometry, typed cells, and the two-readings problem.}
A table is not a bag of strings: its semantics live in a
two-dimensional, typed grid---cell meaning is jointly determined by
its value, its column's header and type, its row's identity, and its
positional relations (merges, hierarchies, captions)~\citep{turl2020,
tabbie2021}.
Three consequences follow.
First, the natural unit is heterogeneous and multi-granular---cell, row,
column, and whole table are all legitimate representational objects, and a
model must decide which to expose (TABBIE exposes all three;
TaPas aggregates evidence over cell selections)~\citep{tapas2020,
tabbie2021}.
Second, table semantics are often relational but implicit: entity
cells link to knowledge bases (masked entity recovery pretraining), and
headers carry the schema~\citep{turl2020, tabert2020}---closer to graph's
explicit relations than to text's linear order.
Third, and most distinctively, tables suffer a two-readings
polarity: the same artifact is either a semi-structured document (row and
column geometry as semantics, for QA and retrieval) or a matrix of
i.i.d.\ feature rows (for prediction), and the two readings demand
different invariances---the document reading requires sensitivity to
row/column identity; the ML reading requires insensitivity to row
order~\citep{fttransformer2021, tabpfn2023}.
A joint embedding must state which reading it serves, or the geometry it
learns will be wrong for the other.

\paragraph{Time series: continuous axis, no standard units, magnitude
semantics.}
Time series is indexed by continuous time and is the modality that
lacks an inherent, semantically meaningful atomic unit:
individual samples carry no independent meaning, making window lengths and
series boundaries modeling choices that must be manufactured~\citep{takens1981,
nie2023times}.
Although physically similar to audio as 1D sampled signals, time series differ
fundamentally in semantics: audio is perceptual and grounded in what is heard,
whereas time series capture statistical temporal dependencies across
observations like prices or telemetry, despite porous boundaries in
physiological waveforms~\citep{hubertecg, eegpt}.
Together, this continuous axis and missing unit produce three distinctive
properties.
First, the observation grid itself varies, spanning regular sampling,
irregular clinical panels, and point processes where timestamps are the
observations~\citep{nhp, mtan,easytpp}.
Second, recurring motifs (e.g., a QRS complex or weekday seasonality) are
domain-specific and operationally defined rather than fixed by community
convention like words in text and audio, leaving distributional word2vec-style
machinery without a ready-made vocabulary~\citep{totem, nutime}.
Third, and most distinctively, amplitude is semantics:
a $2^\circ$C versus $20^\circ$C excursion carries the core signal, yet the
instance normalization that eases cross-dataset training erases it,
reducing encoders to shape-only views that require scale re-injection at
the decoder~\citep{nutime, hardcoded} or explicit statistical retention (e.g.,
NuTime attaching window mean and standard deviation;~\citealp{nutime}).
A joint embedding must preserve the continuous axis and magnitude while
tolerating unstandardized units and channel permutations.

\paragraph{Signals share a second domain; symbols do not.}
One cross-cutting regularity the per-modality inventory above hides is that every
family built on a continuous signal---audio, time series, and vision---has found
it profitable to represent in the frequency domain.
Audio's default front end is the spectrogram, a time--frequency
image~\citep{audiomae}; time-series models locate their periodicity with the
FFT before modeling, unfolding series by their intrinsic
period~\citep{wu2023timesnet} or enforcing consistency between the
time-domain and frequency-domain views of the same
series~\citep{tfc}; vision in its coordinate-based form must contend with
spectral bias, and explicit Fourier-feature encodings are what let models
learn high-frequency image content~\citep{tancik2020fourier}. Graph
learning, though targeting discrete structures rather than continuous signals,
similarly has a long-standing spectral branch built on the Laplacian
eigenbasis~\citep{ma2023laplacian, lim2023signbasis}.
By contrast, for text, code, and tables, Fourier machinery
appears only as an imported generic
basis~\citep{leethorp2021fnet,sergazinov2025ntkfourier}, never as a data-native spectrum; plausibly because the additive,
shift-stationary structure that Fourier-type bases exploit is a property of
physical signals, which symbolic data lack.
For an omnimodal encoder, the observation has design force: it isolates a
genuine intersection (transform-based encoding shared by audio, image,
graph, and time series) that cuts across the family boundaries this section's
inventory works with---the intersection and difference axes of
our lens (\Cref{sec:framework:lens}) meeting in a single
regularity.

\highlightbox{tred}{Takeaway}{%
Each modality owns a structural property the convergence has not
abolished---and a future joint embedding must design mechanisms that respect
these properties rather than overlooking them.
The properties are not obstacles to be engineered around; they are the
information the modality carries, and the test of a true omnimodal
embedding is whether it preserves them while still fusing.
The next section argues that the fastest route to building those mechanisms is
not to invent each in isolation but to borrow across modalities---the
``reuse'' axis, where one modality's hard-won technique becomes another's
import.%
}

\section{Frontier II: The Cross-Pollination Blueprint}
\label{sec:frontier:pollination}

If \Cref{part:principles} documented the intersections and \Cref{sec:frontier:properties}
the differences, this section is the deliberate exercise of the third
axis: reuse.
We argue that the surest route to an omnimodal joint embedding is not for
each modality to reinvent its missing machinery in isolation, but to
cross-pollinate---to port a method hardened in one modality into another
where its preconditions happen to hold.
Our analysis of each modality suggests a concrete opportunity map:
source method $\to$ target modality $\to$ expected benefit.
\Cref{tab:oppmap} summarizes it, marking which transfers are already
documented and which remain a proposed experiment, with the expected benefit
spelled out for each; the rest of this section expands the most promising vectors, each
of which we frame as a conditional port whose preconditions must hold for
the benefit to materialize.

\begin{table*}[t]
\centering
\small
\setlength{\tabcolsep}{4pt}
\caption{Conditional cross-pollination opportunity map. \textbf{\yes{}} marks at least
one use documented in this survey; \textbf{\avail{}} a proposed cross-family
experiment without absence claim; \textbf{\na{}} no clear mapping proposed here.
Rows are grouped by the source family in which the method was
hardened, without claiming unique origin. The rightmost column states the
expected benefit of the transfer when the target's preconditions hold.}
\label{tab:oppmap}
\adjustbox{width=\textwidth,center}{
\begin{tabular}{@{}m{0.29\textwidth}c c c c c c m{0.35\textwidth}@{}}
\toprule
\textbf{Method (source family)} & \textbf{Text} & \textbf{Vision} & \textbf{Audio} & \textbf{Series} & \textbf{Graph} & \textbf{Tabular} & \textbf{Expected benefit} \\
\midrule
\rowcolor{gray!30}\multicolumn{8}{@{}l}{\textit{From text}} \\
Instruction-conditioned embedding~\citep{su2022instructor,wang2024e5mistral} & \yes & \avail & \avail & \avail & \avail & \avail & One amortized trunk replaces $N$ per-factor/per-task encoders; makes the requested geometry explicit and auditable \\
\rowcolor{gray!9}Nested (Matryoshka) representations~\citep{kusupati2022mrl} & \yes & \yes & \avail & \avail & \avail & \avail & Cheap storage--accuracy dial shared across modalities at serve time \\
Late interaction / multi-vector readout~\citep{faysse2024colpali} & \yes & \yes & \avail & \avail & \avail & \avail & Recovers localized evidence a global vector collapses \\
\rowcolor{gray!9}Relevance-judged contrastive supervision~\citep{xiong2021ance} & \yes & \avail & \avail & \avail & \avail & \avail & Sharpens distinctions coarse in-batch negatives miss \\
Frozen-trunk benchmark across heterogeneous tasks~\citep{conneau2018senteval,muennighoff2023mteb} & \yes & \yes & \yes & \avail & \avail & \avail & Makes ``a better embedding'' operationally meaningful, decoupled from a decoder \\
\addlinespace
\rowcolor{gray!30}\multicolumn{8}{@{}l}{\textit{From audio}} \\
Statistics pooling (second moment)~\citep{xvector} & \avail & \avail & \yes & \avail & \avail & \avail & Retains the variability that separates a steady signal from an erratic one \\
\rowcolor{gray!9}Attack-and-utility evaluation after factor suppression~\citep{etawavlm,privsem} & \yes & \avail & \yes & \avail & \avail & \avail & Turns ``this embedding does not contain $X$'' into an auditable property \\
Layer-wise probing as a pretraining diagnostic~\citep{superb} & \yes & \avail & \yes & \avail & \avail & \avail & Reveals which representation state a task actually consumes \\
\addlinespace
\rowcolor{gray!30}\multicolumn{8}{@{}l}{\textit{From graph}} \\
Declared symmetry enforced architecturally~\citep{hardcoded,ying2021do} & \na & \yes & \avail & \avail & \yes & \avail & States the invariance group in the model rather than guessing it via augmentation \\
\rowcolor{gray!9}Structural (non-positional) position encoding~\citep{ying2021do} & \na & \na & \na & \avail & \yes & \yes & Adds topology/centrality without imposing a false linear lattice \\
Permutation-equivariant treatment of a channel set~\citep{kipf2016semisupervised} & \na & \na & \na & \yes & \yes & \yes & Treats rows/variables as an unordered set, not a fixed axis \\
\addlinespace
\rowcolor{gray!30}\multicolumn{8}{@{}l}{\textit{From time series}} \\
Scale-factorized numeric embedding~\citep{nutime} & \avail & \na & \avail & \yes & \na & \avail & Places numerals (2 vs.\ 20) and absolute levels in a metric relation where current embeddings fail \\
\rowcolor{gray!9}Continuous-time index rather than integer positions~\citep{mtan,trep,nhp,xu2020tgat} & \na & \avail & \avail & \yes & \yes & \na & Handles variable-rate video and dynamic graphs natively \\
Entity-keyed positives or groups & \avail & \avail & \yes & \yes & \avail & \avail & Declares ``same instance/state'' rather than a generic augmentation \\
\addlinespace
\rowcolor{gray!30}\multicolumn{8}{@{}l}{\textit{From tabular / table}} \\
Cross-dataset column identity derived from descriptions~\citep{turl2020,tabert2020} & \na & \na & \na & \yes & \avail & \yes & Aligns a ``channel'' across datasets that share units but not schema \\
\bottomrule
\end{tabular}
}
\end{table*}

\paragraph{Text's instruction tuning $\to$ audio and time series.}
Text resolved the universality/specialization tension by conditioning the
embedding on a task instruction~\citep{su2022instructor}: one model produces
different embeddings of the same input depending on the instruction, amortizing
one trunk instead of $N$ specialist encoders and extending to factors not
anticipated at training.
Audio's answer to its factor superposition was instead a separate encoder per
factor plus a sprawling disentanglement
literature~\citep{neuralfa, seqdisent, prosodytimbre, etawavlm}.
The cross-pollination is direct: an audio encoder that accepts an
instruction---``embed for speaker identity, discarding content'' or ``embed for
acoustic scene, ignoring speech''---would subsume the per-factor table and much
of the disentanglement work, dominating on every axis (one trunk, explicit and
auditable geometry, no need for explicit factor-orthogonality
constraints)~\citep{su2022instructor, wang2024e5mistral}.
The same logic reaches time series, where an instruction-conditioned series
encoder is technically straightforward given the LLM-bridging work already in
place~\citep{test}, and would realize task-aware representations for a
modality that currently has no instruction-conditioned embedding
model.

\paragraph{Audio's codec discretization $\to$ code and time series.}
As detailed in \Cref{sec:transformation:tokens}, neural audio codecs
bridge continuous waveforms and discrete modeling interfaces.
The transferable idea is learned semantic codebooks, and it has two
natural targets.
For code, discretizing execution traces into token sequences could let
code embedding models jointly learn from source text and behavioral semantics,
aligning a function with its trace the way a codec aligns audio with its
content~\citep{fuzzpretrain2024}.
For time series, a learned codebook is the path to language-model
treatment---and indeed has been attempted with byte-pair encoding over
quantized values~\citep{tsbpe} and learned VQ~\citep{totem}, whose generalist
codebook transfers to domains that share no sampling rate or sensor with its
training data.
The precondition flagged by audio is that discretization requires recurring
units, whether attested phones for a speech codec, learned acoustic labels
for BEATs~\citep{beats}, or data-native codes for TOTEM~\citep{totem}.
What none of these cases settles is what the vocabulary means away from its
training corpus: tokens defined per corpus need not denote the same thing
across domains, so the import test is whether a vocabulary---attested or
learned---sustains stable recurring units on the target signal.
Reuse, here, comes with that precondition test,
because recurrence, not signal shape, is what the method was built on.
Where no stable units form, magnitude-preserving numeric
embedding~\citep{nutime} is the more faithful bridge.

\paragraph{Graph's structural modeling $\to$ text and code.}
Graph's export is relational abstraction as a structural layer.
Edgeformers, for example, couples edge text with network context for node and edge representation learning~\citep{jin2023edgeformers}.
Related ideas are already established in code: GraphCodeBERT incorporates data-flow structure~\citep{graphcodebert2021}, while CGM integrates repository-level code graphs into language modeling~\citep{tao2026code}. These examples motivate further study of when explicit relations, such as citation links or code dependencies, improve embeddings beyond content alone.
The transferable lesson here is the graph section's discipline: name the
represented unit and the relational objective before comparing
architectures~\citep{khoshraftar2024survey, wang2024llm_zero_shot_graph}.
Graph's structural modeling is thus a complementary relational layer over other
modalities.

\paragraph{Time series' continuous-time indexing $\to$ video and dynamic graphs.}
Every modality is acquiring a time axis---variable-rate video, continuous-time
dynamic graphs, time-scoped retrieval---and most still treat time as a
positional encoding over integers, the approximation time series abandoned a
decade ago for representations defined on $\R$~\citep{nhp, time2vec,
mtan, trep, lognde}.
For continuous-time dynamic graphs the crossing has already happened: TGAT
learns node embeddings as functions of continuous time, with a functional time
encoding derived from Bochner's theorem~\citep{xu2020tgat}.
Time series therefore exports its coordinate: the machinery for
continuous-time embeddings, multi-time attention, and learned time-embeddings
that handle irregular sampling natively.
The expected benefits are concrete: variable-speed video understanding without
frame re-sampling, and temporal knowledge-graph embeddings that move beyond
treating timestamps as isolated discrete identifiers~\citep{teast, stoke}
to full continuous-time functions.
The broader position is that in any real omnimodal setting (a patient with
waveforms, labs, imaging, and notes; a vehicle with telemetry, camera, and
events) modalities are co-observed on a common clock, and the binding relation
should be temporal co-occurrence, with time series supplying and encoding the
axis---not the hub, but the coordinate system the hub is expressed in.

\paragraph{Factor accounting $\to$ vision and text.}
Audio's most transferable contribution may be methodological rather than
architectural: a factor table that lists, for each factor over an input,
which embedding space keeps it, which removes it, and how each is
scored~\citep{xvector, baevski2020wav2vec2, ecapa, defossez2022encodec}, with
removal metrics that make ``this embedding does not contain $X$''
testable~\citep{etawavlm, privsem, wavshape}.
This template is not an audio artifact: any modality whose inputs superpose
distinct factors can build it, converting ``similarity is task-dependent''
into a per-factor list with metrics.
Vision's appearance-vs-identity factors and text's content-vs-style factors
(sarcasm, sentiment, register) both admit such a table, and importing audio's
removal-objective family would turn factor control from an ad hoc finetuning
trick into an auditable property of the embedding.

\highlightbox{tred}{Takeaway}{%
The pattern across these vectors is that reuse is directed and
conditional, not a free transplant.
Each port succeeds when the target modality satisfies the method's
preconditions---a recurring unit inventory for discretization, a relation as
the represented object for structural modeling, a continuous axis for
continuous-time indexing---and fails, predictably, when it does not.
The cross-pollination that respects these preconditions is, we argue, the most
efficient route to a joint embedding that masters each modality's distinctive
property: not by reinventing a solution per modality, but by routing each
modality to the community that has already solved its problem.%
}

\section{Evaluation: We Cannot Build What We Cannot Measure}
\label{sec:evaluation}

The third level of fragmentation besides research and
surveys is evaluation~(\Cref{sec:intro:silos}), and it is the binding
constraint on omnimodal embeddings: without a shared yardstick, claims
of cross-modal convergence and fusion remain unfalsifiable.
In this section, we apply the 3C lens of coverage, comparability, and
compositionality~(\Cref{tab:capability-axes}) to survey existing
modality-specific benchmarks, diagnose their blind spots, and specify
the requirements for a unified omnimodal evaluation protocol.
Crucially, \Cref{tab:capability-axes} decouples architectural mechanism
from demonstrated capability: a mechanism may architecturally support a
capability without necessarily demonstrating it.
Supporting an input format or sharing a backbone confers coverage,
but establishing comparability and compositionality requires
progressively stricter empirical evidence.

\begin{table*}[t]
\centering
\fontsize{8}{9.4}\selectfont
\caption{The 3C evaluation lens and its correspondence to the paths of
\Cref{sec:transformation,sec:alignment,sec:fusion}.
Each progressively stronger capability requires separate evidence and
cannot be collapsed into a single score.}
\label{tab:capability-axes}

\newcommand{\nl}{\newline\rule{0pt}{9pt}}

\adjustbox{width=\textwidth,center}{
\rowcolors{2}{gray!10}{white}
\begin{tabular}{@{}m{0.26\textwidth}m{0.23\textwidth}m{0.30\textwidth}m{0.22\textwidth}@{}}
\toprule
\textbf{Capability} & \textbf{Primary Supporting Mechanism} & \textbf{Minimum evidence} & \textbf{Common shortcuts (insufficient evidence)} \\
\midrule
\textbf{C1: Coverage} \nl Which input types are encoded and tested?
    & Transformation: \nl $x_m \xrightarrow{g_m} \tilde{x}_a \xrightarrow{f_a} \ve$ \nl {\scriptsize (Supports: C1 only)}
    & Direct processing of the original input, tests at each claimed scale and domain, and defined behavior for unsupported inputs
    & Counting tokenizers, channels, or modality names \\
\textbf{C2: Comparability} \nl Which separate embeddings can retrieve or match one another?
    & Alignment: \nl $\ve_m = f_m(x_m),\, \ve_m \approx \ve_n$ \nl {\scriptsize (Supports: C1, C2)}
    & Both retrieval directions, the same candidates, held-out pairs, and calibration/bias checks
    & A shared backbone, equal vector width, or a common loss \\
\textbf{C3: Compositionality} \nl Does the output depend on \emph{all} inputs?
    & Joint fusion: \nl $\ve_{\sS,r} = F(\{x_m\}, r)$ \nl {\scriptsize (Supports: C1, C2 (optional), C3)}
    & A reusable combined embedding with positive compositional gain $\Delta_{\mathrm{comp}}>0$ (\Cref{eq:compgain}) over the best single or subset input, score fusion, and vector addition
    & Co-occurrence, concatenation, or task accuracy alone \\
\bottomrule
\end{tabular}
}
\end{table*}

\paragraph{Islands of evaluation.}
Most modalities have each matured their own frozen-encoder benchmark.
Text has MTEB~\citep{muennighoff2023mteb} and its multilingual extension
MMTEB~\citep{enevoldsen2025mmteb}, with retrieval suites BEIR~\citep{beir},
AIR-Bench~\citep{airbench2025}, the enterprise-domain
RTEB~\citep{rteb2025}, and reasoning-intensive suites
like BRIGHT~\citep{bright2024} that expose a sharp gap (a model scoring
$59.0$ on MTEB can drop to $18.3$ on BRIGHT).
Code has CoIR~\citep{coir2025}, CPRet~\citep{cpret2025}, and
CoREB~\citep{coreb2026} for the full search pipeline.
Audio has SUPERB~\citep{superb}---established in 2021, three years after
text's SentEval standardized the same frozen-encoder
probing~\citep{conneau2018senteval}---and its successors HEAR for general
audio~\citep{hear}, MARBLE for
music~\citep{marble}, AV-SUPERB for audio-visual~\citep{avsuperb}, the unified
X-ARES~\citep{xares}, and the MTEB-style MSEB~\citep{mseb} and
MAEB~\citep{maeb}.
Vision has classification, retrieval, and detection
suites~\citep{russakovsky2015imagenet, radenovic2018revisiting}, now
aggregated in the MTEB-style MIEB~\citep{xiao2025mieb},
with the successively harder ViDoRe line covering visual document
retrieval~\citep{faysse2024colpali, mace2025vidorev2, loison2026vidorev3},
and MVEB extending the frozen-encoder discipline to
video~\citep{elassadi2026mveb}.
Graph evaluation spans node, link, and graph-property prediction in OGB~\citep{hu2020ogb}, structural stress tests such as LRGB~\citep{dwivedi2022lrgb}, and cross-dataset transfer benchmarks for graph foundation models~\citep{chen2024textspacegfm,yu2026gfmevaluation}.
Tables are scored through task heads rather than an embedding suite:
table QA, fact verification, and text-to-SQL over
WikiTableQuestions, TabFact, and Spider~\citep{wtq2015, spider2018,
tabfact2020,zhou2024dbgpthub}, with MTEB-style frozen-encoder analogs (TEmBed,
TabBench, TRL-Bench) only now appearing~\citep{tembed2026, tabembed2026,
trlbench2026}.
Time series is the outlier: encoders there are evaluated through end-to-end
forecasting metrics~\citep{gifteval, quitobench} or frozen-representation
classification probes on UCR/UEA~\citep{bagnall2018ueaarchive,
dau2019ucrarchive}, but neither strand has consolidated into a broadly
adopted MTEB analog.
Cross-modal retrieval is where the aggregation effort has concentrated most
recently: the M-BEIR and MMEB suites~\citep{wei2024uniir, jiang2025vlm2vec},
which assemble long-standing text--image retrieval benchmarks~\citep{lin2014mscoco},
have been extended by MMEB-V2
(adding video and visual documents;~\citealp{xu2016msrvtt, meng2025vlm2vecv2})
and MMEB-V3 (adding audio
and agent retrieval;~\citealp{huang2026mmebv3}), while MixBench targets the
mixed-modality corpus in which texts, images, and text-plus-image documents
must be retrieved against one another~\citep{li2025grclip} and MMLongEmbed
probes how embedding quality degrades as context
lengthens~\citep{wang2026mmlongembed}.
\Cref{tab:evaluation-islands} projects these islands onto one per-modality
table: the primary suites, what each modality still lacks for an omnimodal
claim, and the protocol-validity checks a unified benchmark must retain. The
coverage pattern is uneven across the 3C lens---most islands test coverage
and within-modality comparability, but none measures compositionality across
all modalities under one protocol.

\begin{table*}[t]
\centering
\small
\rowcolors{2}{gray!10}{white}
\caption{Per-modality projection of the existing evaluation islands, and what
each modality lacks for an omnimodal claim. The rightmost column gives the
modality-specific protocol-validity checks a unified benchmark must retain. The
coverage pattern maps onto the 3C lens: most islands test coverage and
comparability within one or two modalities, but none measures
compositionality across all modalities under one protocol.}
\label{tab:evaluation-islands}
\adjustbox{width=0.995\textwidth,center}{
\begin{tabular}{@{}m{0.11\textwidth}m{0.30\textwidth}m{0.24\textwidth}m{0.35\textwidth}@{}}
\toprule
\textbf{Modality} & \textbf{Primary suites / coverage} & \textbf{What remains missing} & \textbf{Protocol-validity checks to retain} \\
\midrule
Text/code & MTEB, MMTEB, BEIR, BRIGHT~\citep{muennighoff2023mteb,enevoldsen2025mmteb,beir,bright2024}; RTEB~\citep{rteb2025}; CoIR, CoREB~\citep{coir2025,coreb2026} & Direct tests of multi-input fusion; held-out code languages and projects; contamination checks & Separable per-category scores (not a mean); code text/code directions; long/low-resource languages; instruction-swap and lexical-sparse controls \\
Image/video & ImageNet linear probe; $\mathcal R$Oxford/$\mathcal R$Paris; MIEB; ViDoRe v1--v3; MVEB~\citep{russakovsky2015imagenet,radenovic2018revisiting,xiao2025mieb,faysse2024colpali,mace2025vidorev2,loison2026vidorev3,elassadi2026mveb} & Broad audio/graph/table coverage; cross-version index compatibility; motion-sensitivity for video & Crop/view/illumination shifts; localization; frame order, rate, duration, and A--V synchronization perturbations; multi-vector bytes/item and latency \\
Audio & SUPERB, HEAR, MARBLE, AV-SUPERB, MAEB, MSEB~\citep{superb,hear,marble,avsuperb,maeb,mseb} & Graph/time-series inputs; broader multi-input combinations & Frozen lightweight probes with swept budget; layer/branch selection; factor retention and removal; multilingual speech vs.\ music/general-audio \\
Graph/3D & OGB; LRGB; text-space and cross-domain GFM benchmarks~\citep{hu2020ogb,dwivedi2022lrgb,chen2024textspacegfm,yu2026gfmevaluation}. 3D: ModelNet40 shape classification and retrieval~\citep{wu2015shapenets} & One common embedding protocol across families; 3D geometry controls & Topology-only, feature-only, fused controls; node-permutation symmetry; visible unseen node/relation/schema/domain/time; coordinate equivariance for graph-anchored 3D \\
Table & WTQ, TabFact, Spider, ToTTo, TAT-QA; TEmBed, TabBench, TRL-Bench~\citep{wtq2015,tabfact2020,spider2018,tembed2026,tabembed2026,trlbench2026} & Frozen-encoder coverage comparable to MTEB; both document and ML readings & Two-readings polarity (document vs.\ i.i.d.\ rows); column-type and relational structure; leakage-resistant splits \\
Time series & UCR, UEA, Monash, GIFT-Eval, QuitoBench~\citep{dau2019ucrarchive,bagnall2018ueaarchive,godahewa2021monash,gifteval,quitobench} & A broadly adopted decoder-independent frozen-encoder suite & Order-destruction; sampling-rate and missingness shifts; channel permutation; unit/scale and amplitude tests; a tabular baseline as the ``does it need time?'' control \\
Cross-modal & M-BEIR/MMEB, MMEB-V2/V3, MixBench, MMLongEmbed, OmniCHOIR, MultiBench~\citep{wei2024uniir,jiang2025vlm2vec,meng2025vlm2vecv2,huang2026mmebv3,li2025grclip,wang2026mmlongembed,huynh2026omniret,liang2021multibench} & All four+-input combinations; role reversal; compositional-gain controls & Bidirectional retrieval; unimodal, score-fusion, additive-vector and fusion baselines; modality-drop/conflict; pairing noise; calibration; component attribution \\
\bottomrule
\end{tabular}
}
\end{table*}

\paragraph{What the silos cannot measure.}
Three limitations follow from this fragmentation.
First, the benchmark silos encode a decoder-coupling problem by
omission: ``a better embedding'' has operational meaning in text (because MTEB
scores a frozen encoder independent of a decoder), but not in time series
benchmarks that report joint properties of
encoder and forecast head~\citep{gifteval, quitobench}.
Second, even within a modality, probe-budget sensitivity can reverse
conclusions: whether one ``frozen encoder + trained head'' is better than
another depends on how the head is sized, and no suite standardizes this
convincingly~\citep{sslbench, contextinv, perch2}.
Third, and most consequentially, no existing benchmark is truly
omnimodal: MTEB and SUPERB are unimodal, the cross-modal suites (M-BEIR, MMEB, and its V2/V3 extensions) are bi- to few-modal, and none measures a
single embedding model's
performance across all modalities under one protocol.
Indeed, recent analysis of MAEB finds that across $53$ encoders and $30$ tasks
no single model dominates, with a clean double dissociation (contrastive
audio-text models strong on environmental sound but near-random on multilingual
speech, and vice versa)---precisely the kind of finding no unimodal silo could
surface~\citep{maeb}.
The instruments that do exist for compositionality are deliberately narrow
diagnostics: Winoground~\citep{thrush2022winoground},
ARO~\citep{yuksekgonul2023aro}, and SugarCrepe~\citep{hsieh2023sugarcrepe}
expose how badly vision--language encoders mishandle word order, attribution,
and binding---state-of-the-art models score close to chance on Winoground's
caption--image swaps---but they are bimodal contrast sets rather than
embedding suites, and none extends to a third modality, let alone to omnimodal
fusion: C3 remains the least measured of the three capabilities.

\paragraph{Toward a unified omnimodal benchmark.}
A benchmark capable of evaluating a true joint embedding should specify three
dimensions.
\begin{itemize}[leftmargin=2em]
  \item \textbf{Multimodal inputs.} The benchmark must cover Text, Code,
    Image, Video, Audio, Graph, Table, and Time Series, both individually
    and in combination, so that a single model can be scored on every
    modality it claims to handle, and on the cross-modal pairs it claims to
    fuse.
  \item \textbf{Multi-task types.} It must span at least three families:
    unimodal tasks (retrieval, classification, similarity) per modality,
    cross-modal retrieval (text-to-image, audio-to-text, query-to-time-series),
    and multimodal fusion tasks in which the input itself is multimodal
    (multimodal question answering, composed image retrieval~\citep{vo2019tirg,
    saito2023pic2word,lookbench2026}, audio-visual event recognition).
    Critically, it must include compositional gain controls: a joint
    embedding should beat the strongest unimodal encoder and the
    additive baselines of score fusion and vector addition, with unimodal
    ablations and feature-fusion controls,
    or the ``fusion'' is not doing work~\citep{jiang2025vlm2vec} (we formalize
    this requirement as the metric $\Delta_{\mathrm{comp}}$ below).
  \item \textbf{A unified evaluation protocol.} All modalities and tasks must
    use the frozen-encoder, deliberately-weak-head protocol that made
    MTEB and SUPERB credible~\citep{superb, muennighoff2023mteb}, with a
    standardized (and reported) probe budget, intrinsic diagnostics
    (alignment--uniformity~\citep{wang2020alignment}, anisotropy~\citep{ethayarajh2019anisotropy},
    inversion attacks~\citep{morris2023inversion}), and---non-negotiable for
    time series---a decoder-independent retrieval instrument that makes
    ``a better embedding'' operational without coupling it to a forecast head.
\end{itemize}

\paragraph{An operational metric for C3.}
So far compositionality has been a verbal criterion---``beats the best single/subset
input, score fusion, and vector addition''.
To give it a form a benchmark can adopt directly, we propose that
every task whose input spans modalities $x_1,\dots,x_k$ report the
\emph{compositional gain}
\begin{equation}
  \Delta_{\mathrm{comp}}
  = \mathrm{Perf}\!\bigl(f_\theta(x_1,\dots,x_k)\bigr)
  - \max\Bigl\{\,
      \max_{\sS \subsetneq \{1,\dots,k\}} \mathrm{Perf}\!\bigl(f_\theta(x_{\sS})\bigr),\ \
      \mathrm{Perf}\!\bigl(\textstyle\sum_i w_i f_\theta(x_i)\bigr),\ \
      \mathrm{Perf}\!\bigl(\textstyle\sum_i v_i s_i\bigr)
  \Bigr\},
  \label{eq:compgain}
\end{equation}
where $f_\theta$ is the frozen embedding model, $\mathrm{Perf}$ a
higher-is-better utility derived from the task metric (evaluated over
either a vector representation or a candidate ranking score),
$x_{\sS} = \{x_i\}_{i \in \sS}$ a non-empty proper subset of the $k \ge 2$ inputs,
$s_i$ the retrieval score that the separately
encoded $x_i$ assigns to a candidate, and the two sets of fusion weights
$\vw$ and $\vv$ fitted on a validation split so that neither side can tune the
baselines in its favor.
A model earns C3 credit on a task only where $\Delta_{\mathrm{comp}} > 0$. A
leaderboard column reporting the fraction of tasks and modality subsets with
positive gain is then comparable across systems, and degenerates gracefully to
$0$ for pairwise cross-modal suites where each query or candidate item is
unimodal ($k=1$).
$\Delta_{\mathrm{comp}}$ is motivated by the synergy term of the
partial information decomposition (PID) of \citet{liang2023quantifying}, which
decomposes the information that several inputs carry about a target into
redundant, unique, and synergistic parts, the synergistic part arising only
from the joint variable.
However, this parallel is not an equivalence: PID is a property of the
data distribution, whereas $\Delta_{\mathrm{comp}}$ measures the
end-to-end performance of one frozen model, so a positive value attests gain
over the specified baselines, not the presence of synergistic information.

The design principles that MSEB articulated for audio are those an
omnimodal benchmark must inherit: prioritize tasks with real applicability,
measure efficiency as a first-class quantity alongside accuracy, include
retrieval and reasoning rather than classification alone, and require that
evaluation involve no task-specific finetuning~\citep{mseb}.

\paragraph{Why such a benchmark is hard to build.}
The specification above is easy to state and hard to realize, because the
real-world scenarios that most need multimodal fusion happen to be the ones
that resist being frozen into a public benchmark.
\begin{itemize}[leftmargin=2em]
\item \textbf{Fusion data is expensive to label and rarely releasable.}
  The compositional tasks that matter most are the most expensive to
  label: constructing composed image retrieval benchmarks is expensive and
  time-consuming~\citep{huynh2025collm}, some benchmarks turned out
  noisy enough that later work had
  to re-label and prune them~\citep{huynh2025collm}, and expert-level
  multimodal benchmarks have to recruit dozens of annotators with domain
  expertise~\citep{yue2024mmmu,xue2024famma}.
  As a result, most aggregation benchmarks
  reuse labels collected for older datasets instead of commissioning new
  ones~\citep{wei2024uniir}, which often diverge from real-world
  queries~\citep{omniembedaudio}.
  Meanwhile, the scenarios that genuinely fuse modalities (e.g., clinical
  encounters, meetings, vehicle fleets) are locked behind privacy and
  licensing regimes: audio must be anonymized before it can be shared at
  all~\citep{tomashenko2022voiceprivacy}, medical imaging paired with
  free-text reports becomes public only after institutional
  de-identification~\citep{johnson2019mimiccxrjpg}, and license tags are
  simply unspecified for over 70\% of popular datasets on the major hosting
  platforms~\citep{longpre2023provenance}---already enough sprawl that MMTEB
  had to prune tasks with underspecified licenses~\citep{enevoldsen2025mmteb}.
  A benchmark assembled from whatever happens to be releasable will
  systematically under-represent exactly the fusion settings that matter
  most.
\item \textbf{Multimodal scores admit unimodal shortcuts.}
  Multimodal benchmarks repeatedly turn out to admit single-modality priors:
  balancing VQA against language priors exposed that top-performing models had
  been relying on shortcut rules~\citep{goyal2017vqa}, a documented failure
  mode that wins on standard benchmarks while failing under real-world
  distribution shifts~\citep{geirhos2020sa}.
  The compositional-gain controls above are therefore a necessity rather than
  a nicety. Furthermore, since deployed sensors frequently drop out and inputs
  are often noisy, robustness to missing and corrupted modalities must be
  a scored axis rather than an afterthought~\citep{liang2021multibench}.
\item \textbf{Static snapshots go stale.}
  Public test sets are a moving target. Evaluation has been declared in
  trouble unless contamination is measured for each
  benchmark~\citep{sainz2023trouble}, and the MTEB maintenance effort already
  documents popular embedding models trained on data from the same
  distributions as benchmark tasks~\citep{chung2025maintainingmteb}.
  Consequently, static snapshots age fast: frozen suites are rapidly
  saturated by models that still fail simple challenge
  examples~\citep{kiela2021dynabench}. This necessitates dynamic benchmarks
  that continuously regenerate test instances from knowledge-base
  deltas~\citep{zhou2025livesearchbench}, alongside sustained engineering
  for versioning, integrity checks, and contamination
  detection~\citep{chung2025maintainingmteb}.
\item \textbf{Evaluation is computationally and storage-wise prohibitive.}
  Evaluating a model on a text-only suite is already costly enough that
  MMTEB had to down-sample datasets and curate hard negatives to keep
  evaluation affordable~\citep{enevoldsen2025mmteb}. An omnimodal
  superset multiplies the bill across modalities: encoding high-resolution
  visual documents, long-context video, or dense audio streams inflates
  token counts by orders of magnitude~\citep{wang2026mmlongembed,mseb}.
  The cost compounds further with modern architectural trends: repurposing
  billion-parameter MLLMs with test-time reasoning adds substantial inference
  latency~\citep{zhu2025retrvr1,cui2025thinkthenembed}, and multi-vector
  late-interaction readouts multiply index storage and memory
  footprints~\citep{faysse2024colpali,xiao2025metaembed}.
\end{itemize}
None of these difficulties is fatal, as the MTEB lineage shows that licensing
curation, down-sampling, and maintenance can carve a usable yardstick amidst
these complexities~\citep{enevoldsen2025mmteb, chung2025maintainingmteb}. Nonetheless,
they explain why the natural first drafts of an omnimodal benchmark tend to be
aggregations of existing islands, and why aggregation alone will miss the
compositional tasks that the joint-embedding claim stands or falls on.

\paragraph{Why this matters now.}
The convergence of \Cref{part:principles} and the fusion models of \Cref{part:path} are only as
credible as the instrument that measures them.
An omnimodal embedding trained without an omnimodal benchmark will be optimized
for tasks the benchmark prioritizes and blind to those it omits, similarly to the
early audio foundation models. And the double dissociations MAEB found will
recur at the omnimodal scale unless the benchmark is built to expose
them~\citep{maeb}.
Building such a benchmark is a prerequisite for, not a consequence of, the
omnimodal embedding; until it exists, the most central claim of this
paper---that joint embedding unifies and surpasses alignment---is strictly a
hypothesis rather than a measured fact.

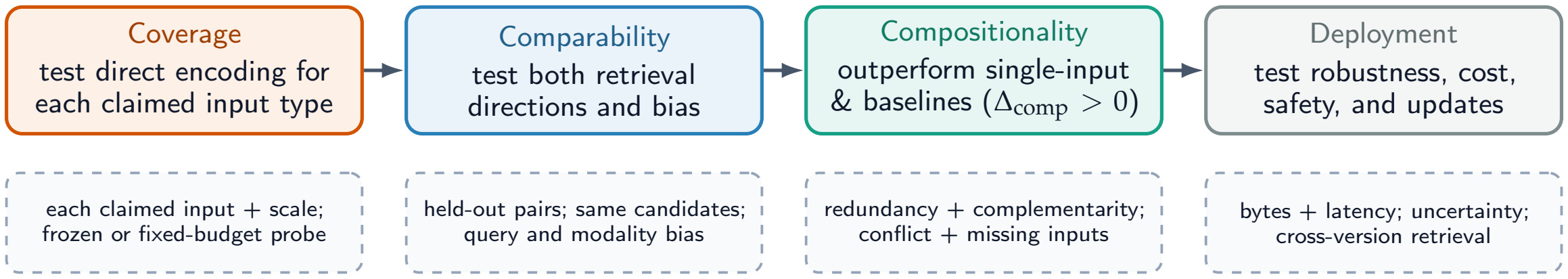
\begin{figure*}[t]
\centering
\begin{adjustbox}{max width=\textwidth}
\begin{tikzpicture}[
  font=\sffamily,
  node distance=4.5mm,
  gate/.style={
    draw, rounded corners=5pt, line width=1.1pt, align=center,
    text width=37mm, minimum height=14mm, font=\small\sffamily, text=paperDark
  },
  check/.style={
    draw=paperBorder, dashed, rounded corners=4pt, align=center,
    text width=37mm, minimum height=11mm, font=\scriptsize\sffamily, fill=bgNeutral,
    line width=0.8pt, text=paperDark 
  },
  arr/.style={-{Latex[length=2.5mm,width=1.7mm]}, draw=paperArrow, line width=1.1pt}
]

\node[gate, fill=c1Bg, draw=c1Border] (C) {{\bfseries\color{c1Text} Coverage}\\[2pt]\footnotesize\color{paperDark} test direct encoding for each claimed input type};
\node[gate, fill=c2Bg, draw=c2Border, right=of C] (P) {{\bfseries\color{c2Text} Comparability}\\[2pt]\footnotesize\color{paperDark} test both retrieval directions and bias};
\node[gate, fill=c3Bg, draw=c3Border, right=of P] (K) {{\bfseries\color{c3Text} Compositionality}\\[2pt]\footnotesize\color{paperDark} outperform single-input \& baselines ($\Delta_{\mathrm{comp}}>0$)};
\node[gate, fill=depBg, draw=depBorder, right=of K] (R) {{\bfseries\color{depText} Deployment}\\[2pt]\footnotesize\color{paperDark} test robustness, cost, safety, and updates};

\draw[arr] (C)--(P); 
\draw[arr] (P)--(K); 
\draw[arr] (K)--(R);

\node[check, below=4mm of C] {each claimed input + scale;\\frozen or fixed-budget probe};
\node[check, below=4mm of P] {held-out pairs; same candidates;\\query and modality bias};
\node[check, below=4mm of K] {redundancy + complementarity;\\conflict + missing inputs};
\node[check, below=4mm of R] {bytes + latency; uncertainty;\\cross-version retrieval};
\end{tikzpicture}
\end{adjustbox}
\caption{Evidence required for broader claims. Omnimodal results must report the three \emph{3C} capabilities---coverage, comparability, and compositionality---within their declared scope, followed by deployment checks (robustness, cost, safety, and cross-version compatibility). The gates list increasing evidence requirements, not a required processing pipeline.}
\label{fig:evaluation-gates}
\end{figure*}

\begin{table*}[t]
\centering
\fontsize{8}{9.4}\selectfont
\setlength{\tabcolsep}{3.3pt}
\renewcommand{\arraystretch}{1.04}
\rowcolors{2}{gray!10}{white}
\caption{Evidence for systems selected for broad input support or alignment.
Coverage ($C1$) and Comparability ($C2$) are the focus; the Compositionality ($C3$)
column only records whether the same work reports any multi-input evidence.
\emph{D} denotes direct testing, \emph{P} partial testing, and dash an
untested requirement. The table separates parameter reuse from cross-modal
retrieval: sharing a backbone and matching separately encoded inputs are
different capabilities.}
\label{tab:evidence-breadth}
\begin{tabular}{@{}m{0.15\textwidth}m{0.21\textwidth}m{0.20\textwidth}ccc m{0.27\textwidth}@{}}
\toprule
\textbf{System} & \textbf{Inputs tested} & \textbf{Mechanism/output} & $C1$ & $C2$ & $C3$ & \textbf{What remains unproven} \\
\midrule
Meta-Transformer~\citep{metatransformer2023} & Twelve reported data types & Modality tokenizers plus shared frozen backbone; task outputs & D & -- & -- & Reusing parameters does not show that outputs share a retrieval space \\
ImageBind~\citep{imagebind} & Image, text, audio, depth, thermal, IMU & Image-anchor alignment; independent global vectors & D & D & -- & Only image--other pairs are trained directly; comparing two non-image inputs is not fusion \\
4M-21~\citep{bachmann20244m21} & Twenty-one visual, semantic, geometric, feature and metadata channels & Discrete any-to-any token prediction & D & -- & -- & Supports broad conditional generation, but not a general dense retrieval vector \\
ViT-Lens / UniBind~\citep{vitlens,lyu2024unibind} & Multiple 2D, 3D, video, audio and sensor inputs & Adapters map new inputs into existing aligned spaces & D & D & -- & Evaluation covers only a subset of possible input pairs \\
LCO-Embedding~\citep{lcoembedding} & Language-centered image, video and related retrieval tasks & MLLM state alignment plus contrastive refinement & D & D & -- & Improved cross-modal comparison does not establish arbitrary multi-input composition \\
jina-embeddings-v4~\citep{guenther2025jinav4} & Multilingual text and images & Universal retrieval vectors & D & D & P & Tested image--text combinations do not establish broader multi-input composition \\
e5-omni~\citep{e5omni2026} & Text, image, video and audio & Temperature calibration, negative mining, and covariance alignment & D & D & -- & The main evidence concerns calibration and cross-modal comparison \\
jina-embeddings-v5-omni~\citep{jinav5omni} & Text, image, audio and video suite & Separately aligned encoders that preserve the text space & D & D & -- & Extends the space efficiently, but does not evaluate joint composition of all four inputs \\
Conan-embedding-v3~\citep{li2026conan} & Text, image, video, documents and audio & Specialist fusion plus projector recovery & D & D & -- & Adding specialists can shift stored embeddings and degrade retrieval compatibility \\
CodeBind~\citep{chen2026codebind} & Nine modalities incl.\ tactile, 3D and EEG & Separate alignment with a compositional codebook & D & D & P & Testing many pairs does not demonstrate joint fusion of all nine modalities \\
\bottomrule
\end{tabular}
\end{table*}

\paragraph{What the evidence we have already shows.}
We can nevertheless read the current literature against the 3C lens.
\Cref{fig:evaluation-gates} summarizes the gates an omnimodal claim must pass,
from coverage through comparability and compositionality to deployment.
Two evidence matrices make the gates concrete. \Cref{tab:evidence-breadth} scores
systems selected for broad input support or alignment, and shows that most of
them establish coverage and comparability but not compositionality: a shared
backbone or a hub anchor is repeatedly mistaken for fusion.
\Cref{tab:evidence-composition} scores systems that explicitly combine several
inputs; as of September 2026, almost all of them carry only a partial
compositionality grade, with only a single strong result,
which itself is restricted to just two modalities. Compositionality remains the least
established capability: today's models cover many inputs and increasingly
match them pairwise, but reliable evidence that an output genuinely depends on
multiple inputs is thin and almost entirely confined to vision--language.

\begin{table*}[t]
\centering
\fontsize{8}{9.4}\selectfont
\setlength{\tabcolsep}{3.3pt}
\renewcommand{\arraystretch}{1.04}
\rowcolors{2}{gray!10}{white}
\caption{Evidence for queries or candidates made from several inputs. In
column $C3$, \emph{P} means that combined inputs were useful, but the study did
not include all of the baselines and stress tests needed for a stronger claim;
\emph{D} means direct compositional-gain evidence against single-input and
simple-fusion baselines. As of September 2026, compositionality remains the
least established capability.}
\label{tab:evidence-composition}
\begin{tabular}{@{}m{0.16\textwidth}m{0.23\textwidth}m{0.20\textwidth}ccc m{0.26\textwidth}@{}}
\toprule
\textbf{System} & \textbf{Multi-input case tested} & \textbf{Mechanism / output} & $C1$ & $C2$ & $C3$ & \textbf{What remains unproven} \\
\midrule
Everything at Once~\citep{shvetsova2022everything} & Available subsets of video, audio and text & Input-agnostic fusion Transformer & D & D & P & Tests missing-input subsets, but not all baselines or both retrieval roles \\
VISTA~\citep{zhou2024vista} & Text, image, and image--text queries & Visual tokens injected into a text encoder & D & D & P & Tests composition directly, but only for vision and language \\
UniIR / M-BEIR~\citep{wei2024uniir} & Text, image and image--text queries/candidates & Instruction-conditioned score or feature fusion & D & D & P & Varies query/candidate roles, but uses only two modalities \\
VLM2Vec~\citep{jiang2025vlm2vec} & Text, image and instructed combinations & Pooled MLLM hidden states & D & D & P & Rewriting the prompt does not show that both inputs affect the result \\
MM-Embed / GME~\citep{lin2025mmembed,zhang2025gme} & Vision--language retrieval and combinations & Negative sampling by modality and fused training & D & D & P & Does not test arbitrary input subsets \\
UniME-V2~\citep{gu2026unimev2} & Universal vision--language embedding tasks & Model-generated semantic supervision & D & D & P & Richer supervision; evaluation remains limited to vision and language \\
OmniRet~\citep{huynh2026omniret} & Text--vision--audio combined queries & Resampling and attention-sliced-Wasserstein pooling & D & D & P & Three media types, but only some combinations \\
SOLAR~\citep{yang2026solar} & Image--text combinations on both sides & Shared/private masking and joint vectors & D & D & D & Both retrieval roles, but only two modalities \\
OmniUE~\citep{wang2026omniue} & Text, image, video, audio, regions and spans & Omni-LLM interactive embeddings & D & D & P & Many prompt formats; no independent benchmark replication at the cutoff \\
DME~\citep{chen2026dme} & Queries over images, videos and visual documents & Evidence-grounded latent reasoning; one serving vector & D & D & P & Full composition tests and serving details are not public \\
\bottomrule
\end{tabular}
\end{table*}

\highlightbox{tred}{Takeaway}{%
We cannot build what we cannot measure. Existing frozen-encoder benchmarks
remain isolated islands: they reliably certify coverage (C1) and pairwise
comparability (C2), but almost entirely overlook compositionality (C3)---the
defining test of genuine joint fusion. Because multimodal evaluation easily
admits unimodal shortcuts and data licensing bottlenecks, evaluating genuine
compositional gain ($\Delta_{\mathrm{comp}}$) across all modalities remains
unresolved. Until a unified, decoder-independent benchmark exists, the thesis
that joint embedding unifies and surpasses alignment remains a compelling
hypothesis rather than an established empirical fact.%
}

\section{Conclusion and Future Outlook}
\label{sec:conclusion}

We have traced the trajectory of embedding research from fragmentation to
fusion.
The field began as separate continents, each modality with its own research
community, its own surveys, and its own benchmarks (\Cref{sec:intro}).
Beneath this fragmentation, four trends quietly converged: contrastive learning
became the lingua franca, the Transformer became the universal encoder, the
pretrain--finetune paradigm standardized training, and the representational
granularity shifted from parts to whole semantic units
(\Cref{sec:objectives,sec:arch,sec:data}).
That convergence is not coincidence but evidence of a deeper
alignment toward a shared ``platonic'' representation of reality, in which a
single geometry underlies embeddings across modalities~\citep{huh2024platonic,
uniembedgeom2025}.

Unification along this trajectory unfolds in three steps.
Transformation (\Cref{sec:transformation}) homogenizes inputs via
rendering, serialization, or discretization, buying coverage at the
cost of native structural loss.
Alignment (\Cref{sec:alignment}) stitches separately encoded spaces
together for comparability, but hits the geometric ceiling of the
modality gap~\citep{liang2022modalitygap}.
Joint embedding (\Cref{sec:fusion}), our advocated terminus,
enables deep cross-modal interaction
to achieve compositionality, fusing when inputs are multimodal and
specializing when unimodal, subsuming alignment provided training
recipes actively mitigate modality competition.

But reaching this terminus requires more than scaling a single backbone.
It demands respecting modality-specific properties rather
than dismissing them: symbolic compositionality, spatial-temporal
topology, factor superposition, explicit graph relations, tabular grid
schemas, and continuous magnitude semantics (\Cref{sec:frontier:properties}).
Rather than reinventing solutions in isolation, communities must pursue
conditional cross-pollination, by porting hardened techniques where
preconditions hold (\Cref{sec:frontier:pollination}).
Crucially, none of this is measurable without a unified, decoder-independent
omnimodal benchmark (\Cref{sec:evaluation}), which is the binding prerequisite
for turning the joint embedding thesis from a hypothesis into a measured fact.

Ultimately, a true omnimodal joint embedding represents a shift in
what embedding models are for.
Moving beyond modality-specific retrieval interfaces, it establishes
a universal representational interface where any datum is indexed,
compared, and composed within a common geometry.
This is what human intelligence has always done, and it is urgent because
omni representation is lagging behind omni generation (\Cref{sec:intro:why}).
If the Platonic conjecture holds and a shared geometry underlies all
modalities, then the omnimodal joint embedding is the vehicle that exposes it,
marking a vital stepping-stone toward more general forms of
machine intelligence that perceive the world not modality by modality, but
whole.

\appendix
\clearpage
\section{Per-Modality Cheat Sheets}
\label{app:cheatsheets}

This section includes cheat sheets for readers who want a quick orientation in a single
modality's history and milestones without interrupting the concept-first main
thread. For each modality we first record the static anatomy of
\Cref{sec:framework,sec:data}---the data objects, units, and granularities it
offers; the backbone and inductive bias that suit it; the recurring
objectives; and the interfaces or exposed embedding object it serves---and
then the developmental arc, a list of milestone modeling works,
a separate list of evaluation benchmarks, and a pointer to where the
modality enters the cross-modal argument.

\subsection{Text}
\label{app:text}

\begin{figure}[htbp]
    \centering
    \includegraphics[width=\textwidth]{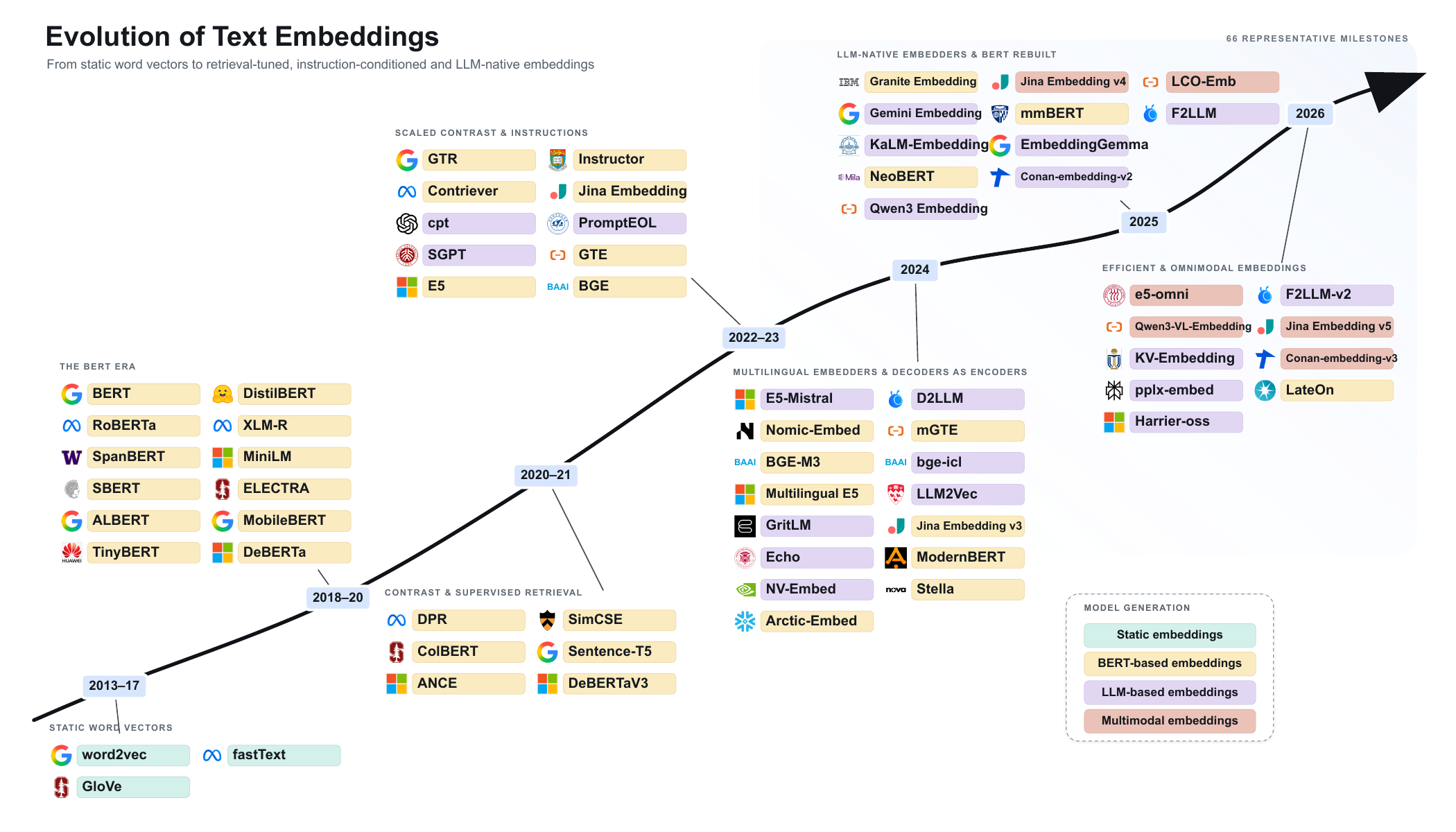}
    \caption{Development history of text embeddings.}
    \label{app:fig:text}
\end{figure}

\paragraph{Data objects, units, and granularities.} Subwords, tokens, spans,
clauses, sentences, paragraphs, and documents; boundaries are partly supplied
by a symbol system. The favored representational object shifts upward toward the sentence and the
document (\Cref{sec:data}), with the token sequence retained as a local object
when grounding is needed.

\paragraph{Backbone and inductive bias.} Encoder-only Transformers and,
increasingly, causal decoder-only LLMs repurposed as embedders; the only
inductive bias beyond attention is a positional (and sometimes
RoPE/ALiBi) encoding over an ordered token sequence.

\paragraph{Objectives.} MLM pretraining giving way to contrastive finetuning
(SimCSE, DPR), contrastive training with role prefixes (E5), and
natural-language instruction conditioning (INSTRUCTOR, E5-Mistral).
Next-token prediction joins as a hybrid head (GRIT) and decoder-only
generative pretraining as a pretext for embedders.

\paragraph{Interfaces / exposed object.} A single global sentence/document
vector for retrieval is the default; multi-vector or late-interaction readouts
(ColBERT-style) retain the local token sequence for grounding, and
instruction-conditioned interfaces expose a task-aware vector.

\paragraph{Arc.} Static word vectors $\to$ contextual representations $\to$
contrastive finetuning $\to$ task-aware (instruction) embeddings $\to$
decoder-only LLMs as embedders.

\paragraph{Modeling milestones.} 
\begin{itemize}[leftmargin=2em]
    \item \textbf{Static word vectors: } word2vec~\citep{mikolov2013distributed},
    GloVe~\citep{pennington2014glove}, fastText~\citep{bojanowski2017fasttext}
    \item \textbf{Contextual \& BERT-based \& contrastive embeddings: } USE~\citep{2018use}, ELMo~\citep{peters2018elmo}, BERT~\citep{devlin2019bert}, RoBERTa~\citep{liu2019roberta}, SpanBERT~\citep{joshi2019spanbert}, ALBERT~\citep{lan2020albert}, TinyBERT~\citep{2020tinybert}, DistilBERT~\citep{sanh2019distilbert}, XLM-R~\citep{2019xlm-r}, MiniLM~\citep{2020minilm}, ELECTRA~\citep{clark2020electra}, MobileBERT~\citep{2020mobilebert}, DeBERTa~\citep{2021deberta,he2021debertav3}, SBERT~\citep{reimers2019sbert},
    DPR~\citep{karpukhin2020dpr}, ColBERT~\citep{khattab2020colbert,Santhanam2022colbert2}, SimCSE~\citep{gao2021simcse}, Sentence-T5~\citep{2021sentence-t5}, GTR~\citep{2021GTR}, Contriever~\citep{izacard2022contriever},
    E5~\citep{wang2022e5,2024multilinguale5}, INSTRUCTOR~\citep{su2022instructor}, GTE~\citep{li2023gte,zhang2024mgte}, BGE~\citep{xiao2023bge,chen2024m3embed}, Jina Embedding v1-3~\citep{2023jina-embedding,2023jina-embedding-v2,2024jina-embedding-v3}, Nomic Embed~\citep{nussbaum2024nomic}, Arctic-Embed~\citep{merrick2024arctic}, ModernBERT~\citep{2024modernbert}, NeoBERT~\citep{lebreton2025neobert}, mmBERT~\citep{marone2025mmbert}, LateOn~\citep{2026lateon}
    \item \textbf{LLM-based embeddings: } cpt~\citep{2022cpt}, SGPT~\citep{muennighoff2022sgpt}, PromptEOL~\citep{jiang2023prompteol},
    E5-Mistral~\citep{wang2024e5mistral}, GRIT~\citep{muennighoff2024grit}, Echo~\citep{2024echo}, NV-Embed~\citep{lee2025nvembed}, LLM2Vec~\citep{llm2vec2024},
    D2LLM~\citep{liao2024d2llm},
    Gemini Embedding~\citep{geminiembed2025}, KaLM-Embedding~\citep{kalmv2_2026}, Qwen3-Embedding~\citep{qwen3embed}, Conan-Embedding~\citep{2025conan-embedding-v2}, F2LLM~\citep{f2llm}, F2LLM-v2~\citep{f2llmv2}, ML-Embed~\citep{mlembed2026}
\end{itemize}

\paragraph{Evaluation benchmarks.} The text community built the canonical
frozen-encoder instrument: MTEB~\citep{muennighoff2023mteb}, its retrieval
focus BEIR~\citep{beir} and AIR-Bench~\citep{airbench2025}, the
reasoning-intensive BRIGHT~\citep{bright2024}, and the multilingual
MMTEB~\citep{enevoldsen2025mmteb}. Beyond English, the MTEB recipe has been
re-instantiated per language, each surfacing where general (often
English-centric) embedders fail: C-MTEB for
Chinese~\citep{xiao2023bge}, MTEB-French~\citep{ciancone2024mtebfrench},
the Scandinavian Embedding Benchmark (Danish, Swedish,
Norwegian)~\citep{enevoldsen2024scandinavian}, ruMTEB for
Russian~\citep{snegirev2025rumteb}, FaMTEB for
Persian~\citep{zinvandi2025famteb}, VN-MTEB for
Vietnamese~\citep{pham2025vnmteb}, MTEB-NL for Dutch~\citep{banar2025mtebnl},
and a German text-embedding clustering benchmark~\citep{wehrli2024german}.

\paragraph{Cross-modal pointers.} Contrastive intersection
(\Cref{sec:objectives:contrastive}); LLM-as-encoder and pooling
(\Cref{sec:arch}); instruction tuning as a cross-pollination export
(\Cref{sec:frontier:pollination}); polysemy and anisotropy as a difference
(\Cref{sec:frontier:properties}); MTEB/MMTEB evaluation silos (\Cref{sec:evaluation}).

\subsection{Code}
\label{app:code}

\begin{figure}[htbp]
    \centering
    \includegraphics[width=\textwidth]{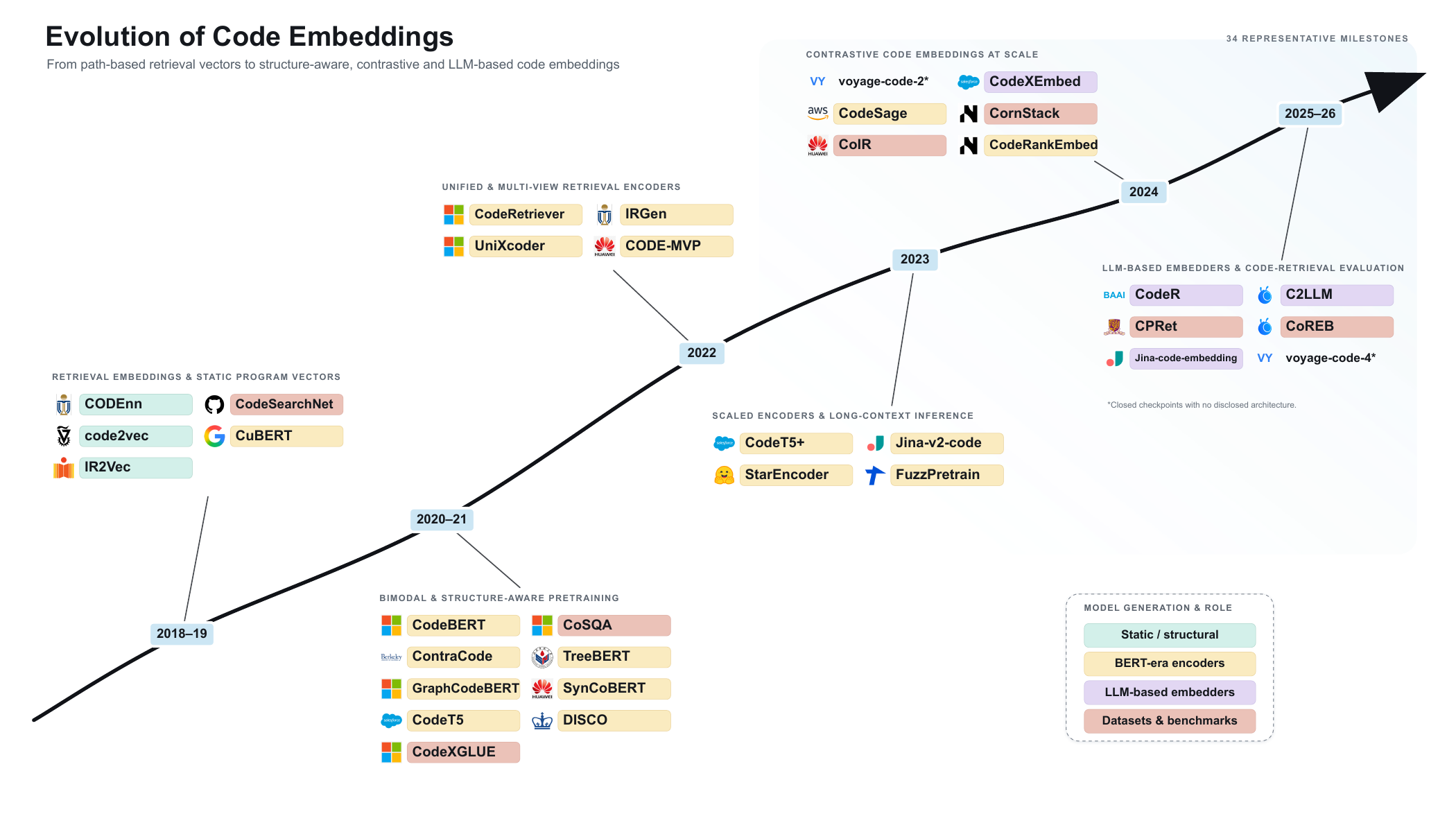}
    \caption{Development history of code embeddings.}
    \label{app:fig:code}
\end{figure}

\paragraph{Data objects, units, and granularities.} Tokens, statements,
blocks, methods, files, modules, projects; an attested hierarchical
structure (the AST and the data-flow graph) sits above the surface text.
The current frontier exposes intermediate-representation units (LLVM IR)
that are behavior- and machine-independent.

\paragraph{Backbone and inductive bias.} Transformer encoders with a
pluggable structural prior: GraphCodeBERT injects data-flow
adjacency as an attention mask, UniXcoder linearizes the AST into the token
stream. Inductive bias is supplied by these structure-bearing inputs and
masks, not by the Transformer backbone itself (\Cref{sec:arch:specific}).

\paragraph{Objectives.} Bimodal masked prediction and replaced-token
detection (CodeBERT) $\to$ identifier-aware and structure-aware masked
targets (GraphCodeBERT, CodeT5) $\to$ contrastive code--code / code--text
views (CodeRetriever, CODE-MVP, CodeSage) $\to$ execution- and trace-grounded
contrastive pretraining (fuzz pretraining).

\paragraph{Interfaces / exposed object.} A global function/method vector for
code search dominates; the AST/data-flow structure is a relational
view and not by itself the served object, so the typical interface is a
single code vector, occasionally with instruction conditioning.

\paragraph{Arc.} General-purpose BERT on code $\to$ bimodal NL-PL pretraining
$\to$ structure-aware (data-flow / AST / multi-view) pretraining $\to$
unified encoder-decoder models $\to$ retrieval-focused contrastive code
embeddings $\to$ Code-LLM-based embedders. Each era asks ``what structure of
code should the representation preserve?'' and answers it one layer
deeper each time: surface tokens, syntax, semantics, behavior~\citep{zhang2024unifying}.

\paragraph{Modeling milestones.} \Cref{app:fig:code} summarizes 34
representative works; the milestones below group them by the layer of code
structure they set out to preserve---the question the Arc above answers one
era deeper each time.
\begin{itemize}[leftmargin=2em]
    \item \textbf{Static structural embeddings (pre-pretraining): }
    CODEnn~\citep{gu2018deepcs} (joint NL--code embedding for
    code search: LSTM max-pooled method-name, API-call, and body features
    aligned with descriptions),
    code2vec~\citep{alon2019code2vec} and
    code2seq~\citep{alon2019code2seq} (AST path-contexts as the unit, with
    attention over sampled paths)
    \item \textbf{Masked, bimodal, and structure-aware pretraining: }
    CuBERT~\citep{kanade2020cubert} (BERT for Python),
    CodeBERT~\citep{codebert2020} (bimodal PL+NL with replaced-token
    detection),
    TreeBERT~\citep{jiang2021treebert} (the AST serialized as composition
    paths and trained with tree masked language modeling plus node order
    prediction),
    GraphCodeBERT~\citep{graphcodebert2021} (data-flow structure; edge
    prediction and code-structure alignment),
    SynCoBERT~\citep{syncobert2021} and CODE-MVP~\citep{codemvp2022}
    (syntax-/multi-view-contrastive code representations),
    CodeT5~\citep{codet5_wang2021} (identifier-aware unified
    encoder-decoder),
    PLBART~\citep{plbart2021} (denoising-autoencoding seq2seq),
    UniXcoder~\citep{unixcoder2022} (cross-modal code+AST+comment with
    contrastive learning),
    CodeT5+~\citep{codet5plus2023} (mixture-of-objectives code LLMs with
    contrastive learning and text-code matching)
    \item \textbf{Semantics-preserving and cross-language contrastive: }
    ContraCode~\citep{jain2021contracode} (semantics-preserving compiler
    transforms as the augmentation that defines functional equivalence),
    DISCO~\citep{ding2022disco} (synthetic clones and injected real-world
    bugs as program contrasts), and
    language-agnostic alignment across programming
    languages~\citep{langagnostic2024}
    \item \textbf{Compiler-IR embeddings: } IR2Vec builds flow-aware LLVM-IR
    embeddings that are language- and machine-independent~\citep{ir2vec2020},
    and IRGen shows that joint training over source code and
    LLVM IR improves clone-detection embeddings beyond source-only
    CodeBERT~\citep{irgen2022}
    \item \textbf{Execution-grounded pretraining: } FuzzPretrain uses
    fuzzer-generated input--output test cases as dynamic complements and
    distills the behavior signal into code-only
    features~\citep{fuzzpretrain2024}
    \item \textbf{Retrieval-focused contrastive embedders: }
    CodeRetriever~\citep{coderetriever2022}, CodeSage~\citep{codesage2024},
    the code variant of Jina Embedding v2~\citep{2023jina-embedding-v2}, and
    CodeRankEmbed~\citep{cornstack2025}
    \item \textbf{Code-LLM-based embedders: }
    StarEncoder~\citep{li2023starcoder} (the encoder-only model of the
    StarCoder family),
    CodeXEmbed~\citep{2024codexembed}, CodeR~\citep{generalistcodeembed2025},
    Jina-code~\citep{2025jina-code}, and C2LLM~\citep{c2llm2025}
\end{itemize}

\paragraph{Evaluation benchmarks.} CodeSearchNet~\citep{codesearchnet2019}
originated large-scale NL$\to$code search (6M functions, 6 languages),
and CodeXGLUE~\citep{codexglue2021} consolidated code intelligence into a
10-task / 14-dataset benchmark (code search, clone detection, defect
detection, completion, translation, summarization), and
CoSQA~\citep{huang2021cosqa} added 20K+ human-annotated web search queries
paired with code for search and question answering. The recent
MTEB-style frozen-encoder suite for code retrieval is
CoIR~\citep{coir2025}, with CPRet~\citep{cpret2025} for competitive
programming and CoREB~\citep{coreb2026} extending the pipeline.

\paragraph{Cross-modal pointers.} Replaced-token detection and
identifier-aware objectives as a reconstructive difference
(\Cref{sec:objectives:reconstructive}); contrastive NL-PL alignment and
zero-shot code-to-code search
(\Cref{sec:objectives:contrastive}, \Cref{sec:alignment}); structure-aware encoders
as a strong-prior architecture (\Cref{sec:arch:specific}); executable
semantics, cross-language equivalence, and hierarchy as the code-specific
difference (\Cref{sec:frontier:properties}); codec/trace discretization and
graph structural modeling as cross-pollination targets
(\Cref{sec:frontier:pollination}); the CoIR/CoREB evaluation silo (\Cref{sec:evaluation}).

\subsection{Table}
\label{app:table}

\begin{figure}[htbp]
    \centering
    \includegraphics[width=\textwidth]{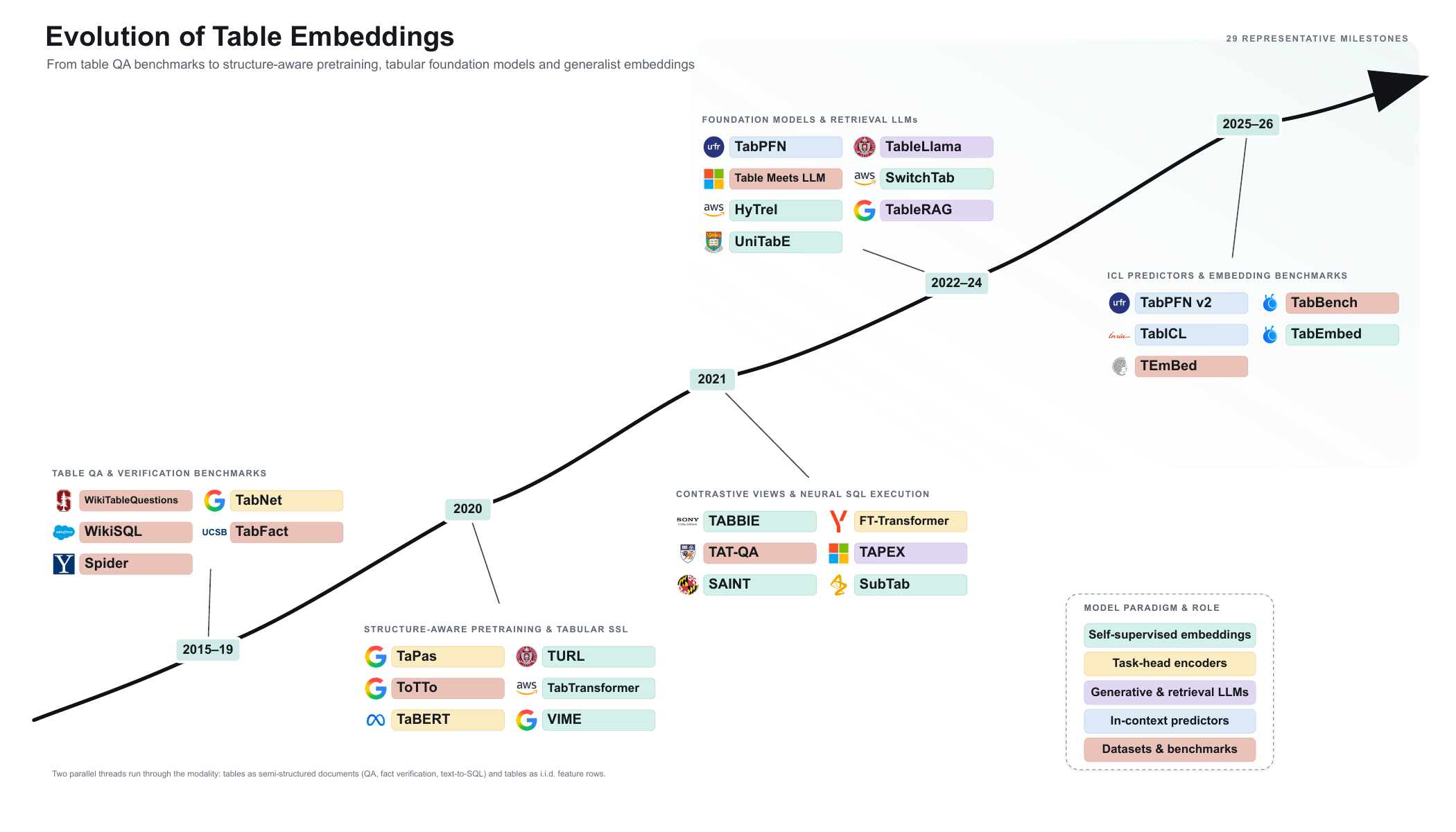}
    \caption{Development history of tabular embeddings.}
    \label{app:fig:table}
\end{figure}

\paragraph{Data objects, units, and granularities.} Cell, header/type, row,
column, and whole table; the affine 2D grid geometry is itself part of the
semantics. Unlike most modalities, all three granularities (cell, row,
column) are legitimate representational objects, and the chosen grain is
task-dependent---TABBIE deliberately exposes all three.

\paragraph{Backbone and inductive bias.} Structure-aware encoders inject the
strong prior that a cell's meaning depends on its value, its column type,
and its row, but they do so differently: TABBIE runs two Transformers over
the rows and the columns to recover the 2D grid that a flat serialization
would hide; TaPas keeps a flat serialization and re-attaches the grid
through explicit row and column index embeddings; and FT-Transformer, on the
tabular-ML reading, tokenizes each feature of a single row with
feature-specific parameters before attending across features. The
inductive bias is strong enough that
a shared text backbone fed a serialized ``column=value'' string underperforms
a structure-aware encoder.

\paragraph{Objectives.} Masked entity recovery over web tables (TURL),
corrupt-cell detection (TABBIE), masked table--text pretraining (masked
language, column, and cell-value prediction in TaBERT; masked language
modeling over text and table cells in TaPas), cross-row intersample
contrast (SAINT); the ML reading prefers a
representation-learning objective that mirrors TabPFN-style in-context
prediction.

\paragraph{Interfaces / exposed object.} From a single global table vector,
to per-cell / per-row / per-column vectors that let consumers query at their
own grain (TABBIE); the document reading serves a global or row-level vector
for QA and retrieval, the ML reading a row embedding for supervised
prediction. The same artifact thus supports two served objects.

\paragraph{Arc.} Tables sit on a fault line between two readings that the
field has never fully reconciled: \emph{tables-as-semi-structured documents}
(a 2D layout of typed cells with headers, captions, and row--column
geometry, the natural unit for retrieval and question answering) and
\emph{tabular ML data} (i.i.d.\ feature rows, the natural unit for
prediction). The document thread runs from serializing tables into BERT
$\to$ structure-aware table encoders (row--column attention) $\to$
table-text joint pretraining $\to$ table retrieval-augmentation; the ML
thread runs from feature engineering $\to$ tabular deep models $\to$
in-context tabular foundation models.

\paragraph{Modeling milestones (documents).} On the document thread:
\begin{itemize}[leftmargin=2em]
    \item \textbf{Table--language pretraining: } TaBERT~\citep{tabert2020}
    (joint NL+table pretraining over 26M tables),
    TaPas~\citep{tapas2020} (BERT extended with table row/column embeddings for
    weakly-supervised QA),
    TURL~\citep{turl2020} (structure-aware Transformer with a Masked Entity
    Recovery objective for web tables),
    TABBIE~\citep{tabbie2021} (two bidirectional Transformers over rows and
    columns, corrupt-cell-detection objective; embeds cells, rows, and
    columns),
    TAPEX~\citep{liu2022tapex} (table pretraining by learning a neural SQL
    executor over the linearized table)
    \item \textbf{Table LLMs and retrieval augmentation: }
    TableLlama~\citep{zhang2024tablellama} (an open table LLM trained over a
    large corpus of serialized tables),
    TableRAG~\citep{tablerag2024} (schema + cell retrieval
    over million-token tables)
\end{itemize}
\paragraph{Modeling milestones (tabular ML).} On the ML thread:
\begin{itemize}[leftmargin=2em]
    \item \textbf{Supervised deep tabular encoders: }
    TabNet~\citep{arik2021tabnet} (attention-based sequential feature
    selection across decision steps, with a self-supervised pretext phase
    for unlabeled data),
    TabTransformer~\citep{huang2020tabtransformer} (self-attention over
    categorical column embeddings while numerical features bypass the
    Transformer---the direct precursor of the feature-tokenizer line),
    FT-Transformer~\citep{fttransformer2021} (feature-tokenizer + Transformer,
    a supervised deep baseline that embeds all columns)
    \item \textbf{Self-supervised tabular pretraining: }
    VIME~\citep{yoon2020vime} (mask-vector estimation plus masked feature
    reconstruction from corrupted inputs),
    SAINT~\citep{saint2021} (intra-row and
    intersample attention with contrastive tabular pretraining),
    SubTab~\citep{ucar2021subtab} (feature subsetting as the multi-view
    augmentation, reconstructing the full row from each subset),
    HyTrel~\citep{chen2023hytrel} (tables as hypergraphs whose row, column,
    and whole-table hyperedges yield cell, row, column, and table
    embeddings),
    SwitchTab~\citep{wu2024switchtab} (autoencoders that split each row
    embedding into a sample-specific and a shared part, switching the shared
    parts across training samples),
    UniTabE~\citep{yang2024unitabe} (a universal cell-level
    mask-then-predict protocol over TabUnits, pretrained on roughly 13B
    Kaggle rows)
    \item \textbf{In-context tabular foundation models: }
    TabPFN~\citep{tabpfn2023} and TabPFN~v2~\citep{tabpfnv2_2025} (prior-data
    fitted networks: in-context tabular prediction, not a reusable embedding),
    TabICL~\citep{qu2025tabicl} (column-then-row attention that scales
    in-context prediction to hundreds of thousands of rows)
\end{itemize}
\paragraph{Evaluation benchmarks.} Document-thread tables are scored on
QA / semantic-parsing tasks rather than a frozen-encoder embedding suite:
WikiTableQuestions~\citep{wtq2015},
WikiSQL~\citep{zhong2017seq2sql} (80K crowdsourced question--SQL pairs over
24K Wikipedia tables---the corpus that launched neural text-to-SQL),
Spider~\citep{spider2018} (text-to-SQL),
TabFact~\citep{tabfact2020} (table fact verification),
ToTTo~\citep{totto2020} (controlled table-to-text), and
TAT-QA~\citep{tatqa2021} (hybrid table+text QA in finance); TableRAG adds
retrieval-recall benchmarks at million-token scale
(ArcadeQA, BirdQA~\citep{tablerag2024}), and Table Meets
LLM~\citep{sui2024tablemeetllm} shows that no serialization format (markdown,
JSON, or others) is uniformly dominant for LLM consumption. Only very
recently has the MTEB
recipe started to reach tables: TEmBed probes frozen tabular encoders at
cell/row/column/table level across 69 datasets~\citep{tembed2026},
TabBench (with the TabEmbed model) evaluates frozen general-purpose embedders
on tabular classification and row retrieval~\citep{tabembed2026}, and
TRL-Bench wraps twenty heterogeneous tabular encoders in one shared embedding
interface probed by lightweight heads~\citep{trlbench2026}. Tabular
ML has otherwise been scored by prediction accuracy
(the OpenML/TabPFN evaluation regime), flagging the same decoder-coupling
gap as time series. A recent survey formalizes the
specialized/transferable/general tiers of tabular representation
learning~\citep{jiangtabularsurvey2026}.

\paragraph{Cross-modal pointers.} Serializing tables as text
(\Cref{sec:transformation:serialization}); row--column structural encoding as a
strong-prior architecture (\Cref{sec:arch:specific}); table-cell discretization
and the only-now-emerging frozen-encoder instruments (\Cref{sec:evaluation});
semi-structured layout as a modality-specific difference
(\Cref{sec:frontier:properties}); column-type and relational structure as a
cross-pollination export to graphs and documents (\Cref{sec:frontier:pollination}).

\subsection{Image and Video}
\label{app:vision}

\begin{figure}[htbp]
    \centering
    \includegraphics[width=\textwidth]{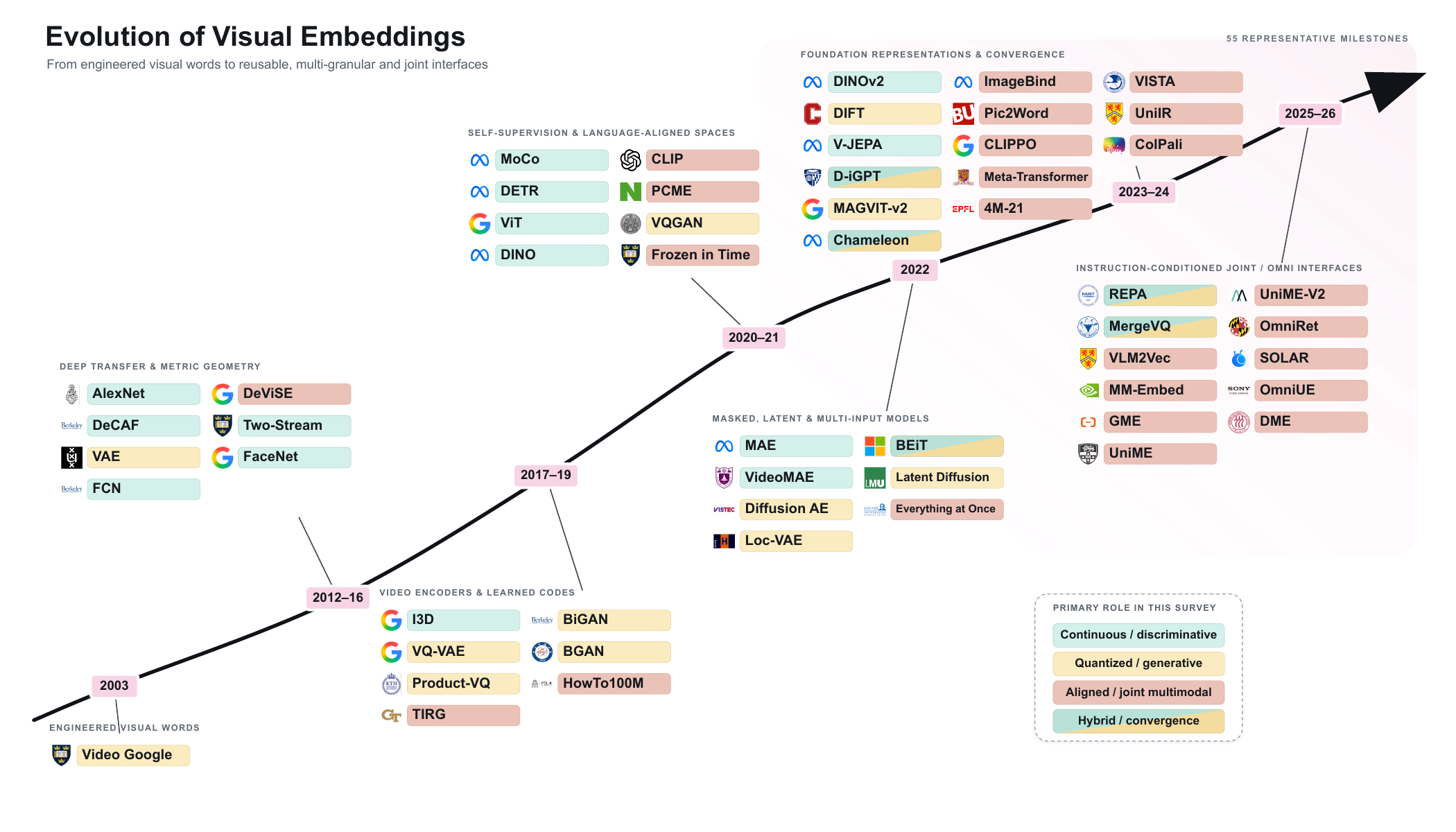}
    \caption{Development history of visual embeddings.}
    \label{app:fig:vision}
\end{figure}

\paragraph{Data objects, units, and granularities.} Pixels, patches,
regions, objects, frames, tracks, clips, and events; most identities above the
pixel are inferred rather than given. The grain moves upward from patches to
whole images, but dense and region-level features are retained where
grounding, correspondence, or motion is needed.

\paragraph{Backbone and inductive bias.} ViTs (patchify + project) became the
default descriptor; 3D-conv patch embeddings handle video tubelets. The
inductive bias CNNs encoded (translation equivariance and locality) is
omitted by the ViT, which retains only the local patch projection and learned
positional encodings, the textbook ``weak prior plus pluggable position'' of
\Cref{sec:arch:transformer}.

\paragraph{Objectives.} Contrastive at the image level (SimCLR, CLIP),
self-distillation (DINO), redundancy reduction (Barlow Twins, VICReg),
masked reconstruction of continuous pixels (MAE),
token-level contrast in early-fusion ViTs. The contrastive and reconstructive
families now coexist; generative VQ objectives (VQ-VAE, MAGVIT-v2) build
discrete visual codes for generation.

\paragraph{Interfaces / exposed object.} One global vector for retrieval and
zero-shot classification is the canonical interface; the local patch set is
preserved by dense and late-interaction models (ColPali) for fine-grained
document retrieval, and equivariance-aware variants keep a set of per-object
embeddings. Video adds the temporal axis on top of an image appearance model.

\paragraph{Arc.} Bag-of-visual-words $\to$ learned CNN features $\to$
instance discrimination $\to$ masked reconstruction $\to$ image-level contrast
and CLIP $\to$ multimodal/dense representations; video adds a temporal axis and
a hierarchy of units.
\Cref{app:fig:vision} summarizes 55 representative works.

\paragraph{Modeling milestones.}
\begin{itemize}[leftmargin=2em]
    \item \textbf{Engineered local descriptors: } Video
    Google~\citep{sivic2003videogoogle} (bag-of-visual-words retrieval),
    SIFT~\citep{lowe2004sift}, Fisher vectors~\citep{perronnin2010fisher}, and
    VLAD~\citep{jegou2010vlad}, which aggregate local descriptors into
    compact global codes
    \item \textbf{Supervised CNN features and readouts: }
    AlexNet~\citep{krizhevsky2012alexnet}, DeCAF~\citep{donahue2014decaf}
    (frozen-CNN features as transferable vectors), ResNet~\citep{he2016resnet};
    FCN~\citep{long2015fcn} and DETR~\citep{carion2020detr} exposed dense and
    object-level readouts of such features, DETR's learned object queries
    serving as detection embeddings
    \item \textbf{Instance retrieval and metric learning: }
    FaceNet~\citep{schroff2015facenet} (triplet-trained identity vectors),
    Product-VQ~\citep{wu2019productvq} (a product quantizer in the VQ-VAE
    bottleneck, so retrieval runs on sub-codeword distance lookups),
    Loc-VAE~\citep{nishimaki2022locvae} (a $\beta$-VAE whose latent
    dimensions localize to brain regions, for medical image retrieval)
    \item \textbf{ViT and self-supervised pretraining: }
    ViT~\citep{dosovitskiy2021vit} (the patch-token interface),
    MoCo~\citep{he2020moco}, SimCLR~\citep{chen2020simclr},
    SwAV~\citep{caron2020swav} (predicting cluster assignments instead of
    pairwise comparison),
    BYOL~\citep{grill2020byol}, Barlow
    Twins~\citep{zbontar2021barlow} and VICReg~\citep{bardes2022vicreg}
    (negative-free redundancy reduction),
    DINO~\citep{caron2021dino},
    iBOT~\citep{zhou2021ibot},
    BEiT~\citep{bao2022beit} (masked prediction against a learned discrete
    codebook), I-JEPA~\citep{assran2023ijepa} (prediction
    moved into latent space), MAE~\citep{he2022mae},
    DINOv2~\citep{oquab2024dinov2}; at the scaling frontier,
    Web-SSL~\citep{fan2025webssl} shows image-only self-supervision on a
    2B-sample web corpus---no language signal at all---can match
    language-aligned pretraining, DINOv3~\citep{dinov3_2025} couples
    web-scale curation with Gram anchoring that keeps patch-level features
    from degrading during long training runs, AIMv2~\citep{aimv2_2024}
    supervises a visual encoder solely through multimodal autoregressive
    prediction, and D-iGPT~\citep{ren2024digpt} rejuvenates autoregressive
    iGPT by predicting semantic tokens from a frozen CLIP teacher
    \item \textbf{Quantized and generative visual codes: }
    VQ-VAE~\citep{vandenOord2017vqvae}, VQGAN~\citep{esser2021taming},
    MAGVIT-v2~\citep{yu2024magvit2} (lookup-free quantization and a causal
    3D tokenizer whose codes let language models beat diffusion on image and
    video generation); the parallel generative-latent
    line---VAE~\citep{kingma2014vae},
    BiGAN/BGAN~\citep{donahue2017bigan,song2018bgan}, latent
    diffusion~\citep{rombach2022ldm}, diffusion
    autoencoders~\citep{preechakul2022diffae}, DIFT~\citep{tang2023dift}, and
    the REPA/MergeVQ representation--generation
    convergence~\citep{yu2025repa,li2025mergevq}---is surveyed in
    \Cref{sec:objectives:generative}, with
    Chameleon~\citep{chameleon2024} interleaving text and visual tokens in one
    autoregressive stream
    \item \textbf{Video encoders: } two-stream~\citep{simonyan2014twostream},
    iDT~\citep{wang2013idt} (camera-motion-compensated dense trajectories),
    C3D~\citep{tran2015c3d}, I3D~\citep{carreira2017i3d},
    SlowFast~\citep{feichtenhofer2019slowfast},
    TimeSformer~\citep{bertasius2021timesformer} (divided space--time
    attention), ViViT~\citep{arnab2021vivit}
    \item \textbf{Video self-supervision and foundation encoders: }
    VideoMAE~\citep{tong2022videomae},
    VideoMAE V2~\citep{wang2023videomaev2} (dual masking that scales masked
    video modeling to millions of videos),
    V-JEPA~\citep{bardes2024vjepa},
    V-JEPA 2~\citep{assran2025vjepa2} (masked latent prediction on more than a
    million hours of video, with an action-conditioned world model that plans
    zero-shot on robots),
    VideoPrism~\citep{zhao2024videoprism} (contrastive pretraining followed by
    masked video modeling that distills a teacher's global and token
    embeddings, one frozen encoder for 30-plus video benchmarks)
    \item \textbf{Video--language pretraining for retrieval: }
    HowTo100M~\citep{miech2019howto100m} (100M narrated video clips),
    Frozen in Time~\citep{bain2021frozen} (a joint video--image encoder for
    end-to-end text--video retrieval), Everything at
    Once~\citep{shvetsova2022everything},
    InternVideo2~\citep{wang2024internvideo2} (joint video--text pretraining
    scaled to billion-parameter regimes)
    \item \textbf{Cross-modal alignment and joint interfaces: }
    DeViSE~\citep{frome2013devise}, TIRG~\citep{vo2019tirg},
    CLIP~\citep{radford2021clip}, PCME~\citep{chun2021pcme},
    Pic2Word~\citep{saito2023pic2word}, CLIPPO~\citep{tschannen2023clippo},
    ImageBind~\citep{imagebind},
    ColPali~\citep{faysse2024colpali} (late interaction over document patch
    tokens), VLM2Vec~\citep{jiang2025vlm2vec}; the many-modality and
    instruction-conditioned successors
    (Meta-Transformer~\citep{metatransformer2023},
    4M-21~\citep{bachmann20244m21}, VISTA~\citep{zhou2024vista},
    UniIR~\citep{wei2024uniir}, MM-Embed~\citep{lin2025mmembed},
    GME~\citep{zhang2025gme}, UniME~\citep{gu2025unime}) are treated as
    fusion-era works in \Cref{sec:fusion}
\end{itemize}
\paragraph{Evaluation benchmarks.} Vision lacks a single MTEB-analog; instead
it relies on task-specific suites---classification on
ImageNet~\citep{russakovsky2015imagenet} (the frozen-encoder linear/zero-shot
probe), instance-level image retrieval on $\mathcal{R}$Oxford/$\mathcal{R}$Paris
with the Easy/Medium/Hard
protocols~\citep{radenovic2018revisiting}, and increasingly the
multimodal embedding benchmark MMEB~\citep{jiang2025vlm2vec, meng2025vlm2vecv2, huang2026mmebv3}.

\paragraph{Cross-modal pointers.} Rendering text as pixels, VisionTS as
transformation (\Cref{sec:transformation}); CLIP/ImageBind alignment hub
(\Cref{sec:alignment}); spatial multi-scale as a difference
(\Cref{sec:frontier:properties}); diffusion latents in hybrid objectives
(\Cref{sec:objectives:generative}); continuous-time index as an import
(\Cref{sec:frontier:pollination}).

\subsection{Audio}
\label{app:audio}

\begin{figure}[htbp]
    \centering
    \includegraphics[width=\textwidth]{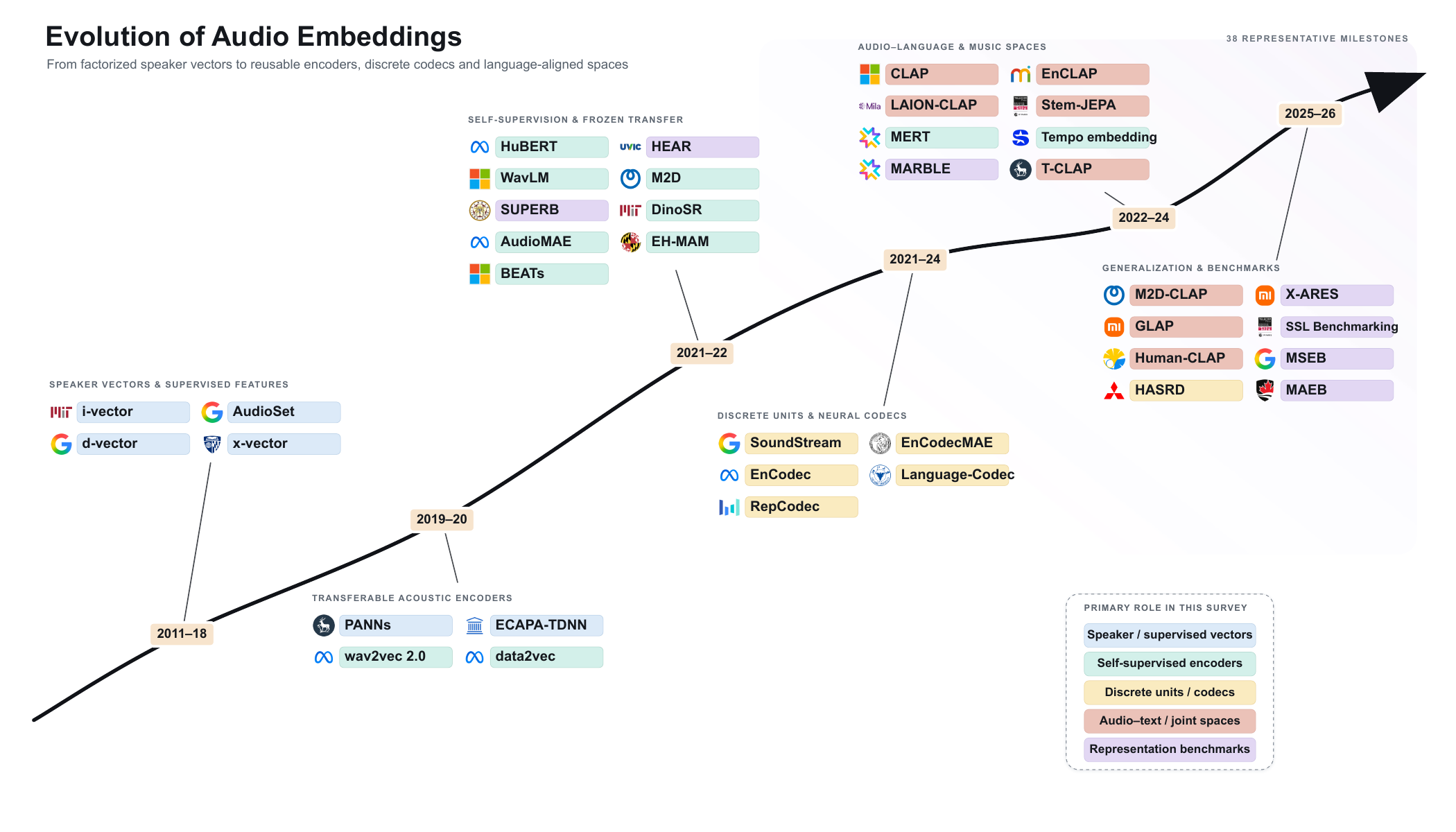}
    \caption{Development history of audio embeddings.}
    \label{app:fig:audio}
\end{figure}

\paragraph{Data objects, units, and granularities.} Samples, frames, phones,
spoken-word segments, utterances, speakers, clips, events, tracks, stems. The
content layer has an attested recurring inventory (phones, syllables, words)
when speech or music is tied to an external symbol system; the factor layer
(content vs.\ speaker vs.\ prosody vs.\ environment) is superposed rather than
hierarchical, so the chosen grain depends on the factor being asked for.

\paragraph{Backbone and inductive bias.} Conformers and transformers
dominate SSL, with speech models reading a temporal feature sequence from a
convolutional encoder over the raw waveform and general-audio encoders
patchifying the time--frequency spectrogram into 2D tokens---so grid
geometry and sampling rate are inductive biases supplied by the front end.
Statistics pooling over frames in the speaker branch (x-vector,
ECAPA-TDNN) deliberately keeps the second moments that text-side global
pooling would erase.

\paragraph{Objectives.} Self-supervised masked prediction over quantized
targets (wav2vec 2.0, HuBERT), contrastive audio--caption alignment (CLAP),
waveform codec discretization into acoustic tokens (SoundStream, EnCodec)
and semantic distillation into a dedicated codebook (SpeechTokenizer),
and removal/anonymization objectives that certify a factor is absent.

\paragraph{Interfaces / exposed object.} The defining ``no single object''
case: a speech-recognition readout is a sequence of contextual frame
representations, a speaker embedding is an utterance-level global vector, a
codec is a discrete code sequence, and a dense frame set is a local object.
The interface must name the factor or
expose a layer-weighted combination (the SUPERB weighted-sum-over-layers
choice).

\paragraph{Arc.} Generative factor analysis (i-vector) $\to$ discriminative
speaker vectors $\to$ self-supervised masked prediction over discovered
units $\to$ discrete codec tokens $\to$ audio-language alignment; four eras
that superseded one another cleanly because a benchmark (SUPERB) could register
the improvement.

\paragraph{Modeling milestones.} \Cref{app:fig:audio} summarizes 38
representative works; the milestones below group them by the factor and
interface they serve.
\begin{itemize}[leftmargin=2em]
    \item \textbf{Speaker vectors: } i-vector~\citep{ivector}, d-vector~\citep{dvector},
    x-vector~\citep{xvector}, ECAPA-TDNN~\citep{ecapa}
    \item \textbf{Self-supervised speech models: } wav2vec 2.0~\citep{baevski2020wav2vec2},
    HuBERT~\citep{hsu2021hubert}, WavLM~\citep{wavlm}, data2vec~\citep{data2vec},
    DinoSR~\citep{dinosr} (online clustering replaces HuBERT's offline k-means
    with machine-discovered, phone-like targets),
    EH-MAM~\citep{ehmam} (easy-to-hard curriculum masking that shifts masking
    toward frames a teacher-predicted loss deems hard)
    \item \textbf{General-audio encoders: } PANNs~\citep{pann} (supervised
    CNNs pretrained on AudioSet and widely reused as frozen feature
    extractors), AudioMAE~\citep{audiomae}, BEATs~\citep{beats},
    M2D~\citep{m2d} (the online network predicts the representations of
    masked patches while the target network encodes only the masked
    patches, so both are pushed to model the input)
    \item \textbf{Neural codecs and discrete units: } SoundStream~\citep{soundstream},
    EnCodec~\citep{defossez2022encodec},
    SpeechTokenizer~\citep{zhang2024speechtokenizer} (distilling HuBERT
    semantics into a dedicated codebook), RepCodec~\citep{repcodec}
    (vector-quantizing pretrained speech representations into semantic
    tokens), Language-Codec~\citep{languagecodec}
    (fewer codebooks so speech LMs consume fewer tokens),
    HASRD~\citep{hasrd} (an RVQ hierarchy whose first codebook carries the
    semantic representation and later ones the acoustic residuals),
    EnCodecMAE~\citep{encodecmae} (masked prediction of codec token indices
    as a universal-audio objective)
    \item \textbf{Audio--language alignment: } CLAP~\citep{clap} and
    LAION-CLAP~\citep{laionclap}, T-CLAP~\citep{tclap} (caption pairs that
    flip event order to teach temporal structure), M2D-CLAP~\citep{m2dclap}
    (masked modeling and contrastive alignment in one model),
    GLAP~\citep{glap} (one contrastive space across speech, sound events, and
    music in many languages), Human-CLAP~\citep{humanclap} (trained against
    human relevance ratings rather than caption identity),
    EnCLAP~\citep{enclap} (codec tokens plus a CLAP embedding feeding a
    caption decoder)
    \item \textbf{Music representations: } MERT~\citep{mert} (masked
    prediction with both codec-token and CQT teachers at 160K-hour scale),
    Stem-JEPA~\citep{stemjepa} (instrument-conditioned latent prediction of a
    target stem for zero-shot stem retrieval),
    tempo embedding~\citep{tempoemb} (a learned translation function that
    maps an embedding to that of its time-stretched version)
    \item \textbf{Disentanglement and privacy: } NeuralFA~\citep{neuralfa},
    Eta-WavLM~\citep{etawavlm}, WavShape~\citep{wavshape}
\end{itemize}
\paragraph{Evaluation benchmarks.} The supervised general-audio era was built
on AudioSet~\citep{audioset} (2.1M ten-second YouTube clips labeled with a
632-class event ontology). Audio converges on the same frozen-encoder protocol as text:
SUPERB~\citep{superb} for speech, HEAR~\citep{hear} for
general audio, MARBLE~\citep{marble} for music, AV-SUPERB~\citep{avsuperb} for
audio-visual, the unified X-ARES~\citep{xares}, the MTEB-style
MSEB~\citep{mseb} and MAEB~\citep{maeb}, and the self-supervised-speech
benchmarking audit~\citep{sslbench}.

\paragraph{Cross-modal pointers.} Factor superposition as the distinctive
audio property (\Cref{sec:frontier:properties}); codec discretization as an
import to code/time series and a precondition test (\Cref{sec:frontier:pollination});
codec tokens and CLAP in objectives (\Cref{sec:objectives}); frozen-encoder
benchmark protocol adopted for speech/audio (\Cref{sec:evaluation}).

\subsection{Graph}
\label{app:graph}

\begin{figure}[htbp]
    \centering
    \includegraphics[width=\textwidth]{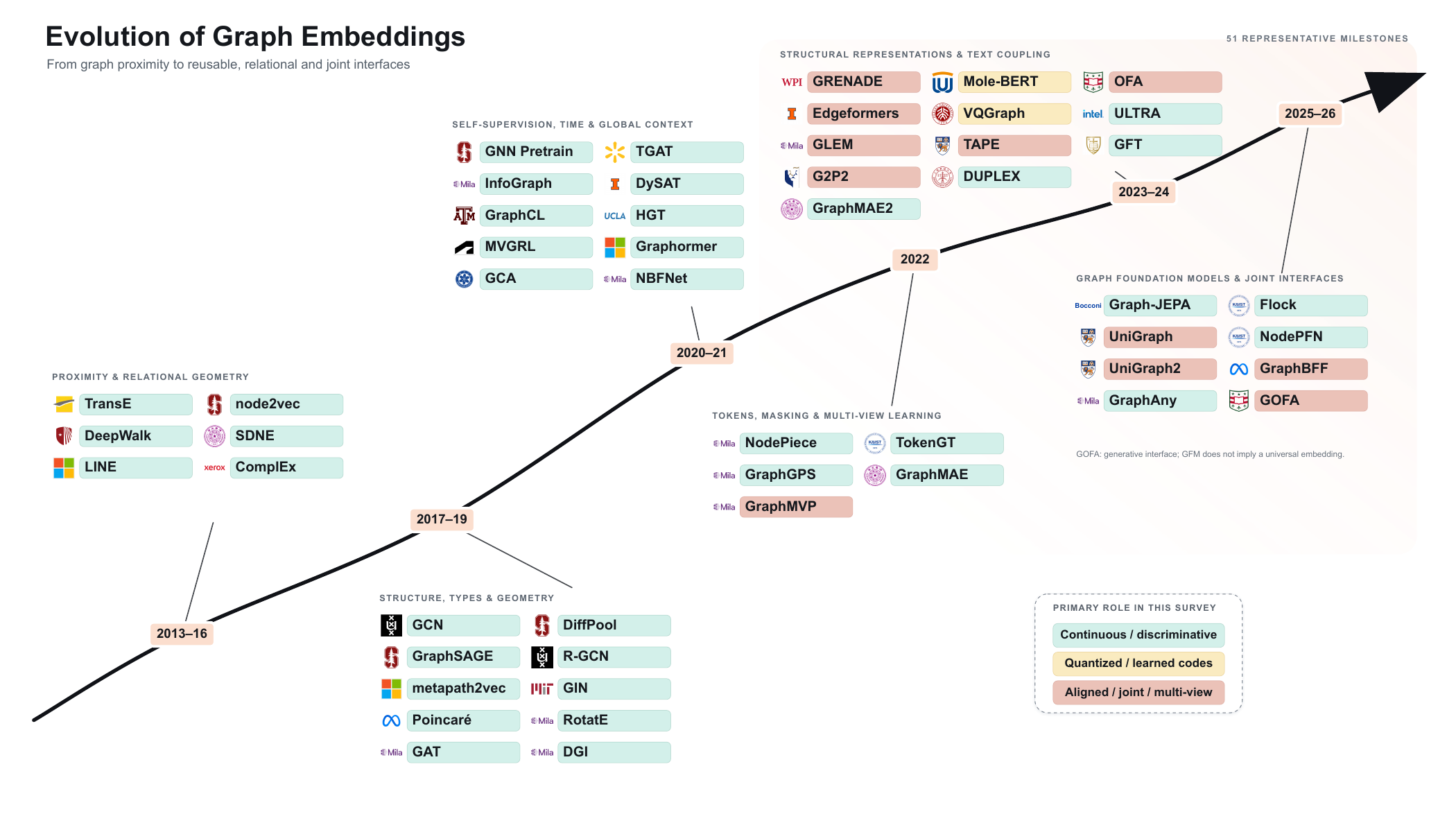}
    \caption{Development history of graph embeddings.}
    \label{app:fig:graph}
\end{figure}

\paragraph{Data objects, units, and granularities.} Node, edge, typed triple,
path, subgraph, whole graph, relational/logical-query record, and
graph-conditioned readout; for dynamic graphs, a time slice. The relations
are part of the represented object, not a by-product of order, so the choice
of granularity coincides with the choice of question (node similarity vs.\
graph similarity vs.\ triple plausibility).

\paragraph{Backbone and inductive bias.} Message-passing GNNs encode the
strong prior of a fixed permutation-equivariant topology; Graph Transformers
reuse attention but rely on structural encodings, attention biases, or edge
features to re-expose the topology (\Cref{sec:arch:specific}). The
inductive bias is therefore in how the graph structure is exposed to the backbone,
not in the attention operator itself.

\paragraph{Objectives.} Random-walk/proximity co-occurrence (DeepWalk,
node2vec); typed-relation scoring and triple plausibility (TransE, RotatE);
contrastive learning over augmented graph views (GraphCL); masked feature
reconstruction (GraphMAE/GraphMAE2); learned structural codes and
distillation (VQGraph); and latent prediction (Graph-JEPA).

\paragraph{Interfaces / exposed object.} Node vectors, triple/entity-relation
scores, subgraph and whole-graph readouts, and relation/edge-conditioned
objects can each be served; the wrong readout silently answers the wrong
question, so the contract must name the represented object and the
permutation/visibility assumptions.

\paragraph{Arc.} Graph representation learning develops through overlapping
threads: graph-specific proximity and relational scoring, structure-aware
encoders, reusable self-supervised representations, and graph foundation
models (GFMs) that seek transfer across graphs and tasks.
A complementary thread aligns graph attributes with language and other
modalities. The change is in what can be reused across domains, not a
necessary progression toward one whole-graph vector.
\Cref{app:fig:graph} summarizes 51 representative works;
the modeling milestones below supply their references.

\paragraph{Modeling milestones.} DeepWalk~\citep{perozzi2014deepwalk},
LINE~\citep{tang2015line},
node2vec~\citep{grover2016node2vec},
SDNE~\citep{wang2016sdne},
TransE~\citep{bordes2013transe},
ComplEx~\citep{trouillon2016complex},
RotatE~\citep{sun2019rotate},
Poincar\'e embeddings~\citep{nickel2017poincare},
DUPLEX~\citep{ke2024duplex};
GCN~\citep{kipf2016semisupervised},
GraphSAGE~\citep{hamilton2017inductive},
GAT~\citep{velikovi2017graph},
GIN~\citep{xu2018how},
DiffPool~\citep{ying2018diffpool},
metapath2vec~\citep{dong2017metapathvec},
R-GCN~\citep{schlichtkrull2018rgcn},
HGT~\citep{hu2020hgt},
NBFNet~\citep{zhu2021nbfnet},
TGAT~\citep{xu2020tgat},
DySAT~\citep{sankar2020dysat},
Graphormer~\citep{ying2021do},
GraphGPS~\citep{rampasek2022graphgps};
GNN pretraining~\citep{hu2020gnnpretrain},
DGI~\citep{velickovic2019dgi},
InfoGraph~\citep{sun2020infograph},
GraphCL~\citep{you2020graphcl},
MVGRL~\citep{hassani2020mvgrl},
GCA~\citep{zhu2021gca},
GraphMAE~\citep{hou2022graphmae},
GraphMAE2~\citep{hou2023graphmae2},
Graph-JEPA~\citep{skenderi2025graph_jepa};
NodePiece~\citep{galkin2021nodepiece},
TokenGT~\citep{kim2022tokengt},
Mole-BERT~\citep{xia2023molebert},
VQGraph~\citep{yang2023vqgraph};
GraphMVP~\citep{liu2022graphmvp},
GRENADE~\citep{li2023grenade},
Edgeformers~\citep{jin2023edgeformers},
G2P2~\citep{wen2023g2p2},
GLEM~\citep{zhao2023glem},
TAPE~\citep{he2023harnessing},
OFA~\citep{liu2024ofa},
UniGraph~\citep{he2025unigraph},
UniGraph2~\citep{he2025unigraph2},
GOFA~\citep{kong2025gofa};
ULTRA~\citep{galkin2024ultra},
Flock~\citep{kim2026flock},
GFT~\citep{wang2024gft},
GraphAny~\citep{zhao2025graphany},
NodePFN~\citep{choi2026nodepfn},
GraphBFF~\citep{bechlerspeicher2026graphbff}; dynamic:
DyG2Vec~\citep{alomrani2022dyg2vec}, TeAST~\citep{teast},
STOKE~\citep{stoke}, ECOLA~\citep{ecola}, TFLEX~\citep{tflex}.

\paragraph{Evaluation benchmarks.} Evaluation spans several complementary
settings: OGB~\citep{hu2020ogb} for node, link, and graph-property prediction;
MoleculeNet~\citep{wu2018moleculenet} for molecular properties; and the
Heterogeneous Graph Benchmark (HGB)~\citep{lv2021hgb} for typed-node and
relation settings. Structural stress tests include the Long Range Graph
Benchmark (LRGB)~\citep{dwivedi2022lrgb} for long-range dependencies and
the heterophily benchmark~\citep{platonov2023heterophily}, including
Roman-empire and Amazon-ratings, for node classification beyond strongly
homophilous graphs. Heterogeneous types and heterophily are distinct axes.
For GFM transfer, the text-space benchmark~\citep{chen2024textspacegfm}
unifies evaluation across graph datasets and tasks, while a recent
benchmark~\citep{yu2026gfmevaluation} jointly varies topic and format domains
through multi-domain pretraining and few-shot adaptation.

\paragraph{Cross-modal pointers.} Explicit relational structure and
non-Euclidean geometry as the distinctive property
(\Cref{sec:frontier:properties}); relational abstraction as an export to
text/code (\Cref{sec:frontier:pollination}); message-passing and Graph
Transformer as a strong-prior architecture (\Cref{sec:arch});
contrastive views and triple reconstruction (\Cref{sec:objectives});
continuous-time machinery shared with time series (\Cref{sec:frontier:pollination}).

\subsection{Time Series}
\label{app:timeseries}

\begin{figure}[htbp]
    \centering
    \includegraphics[width=\textwidth]{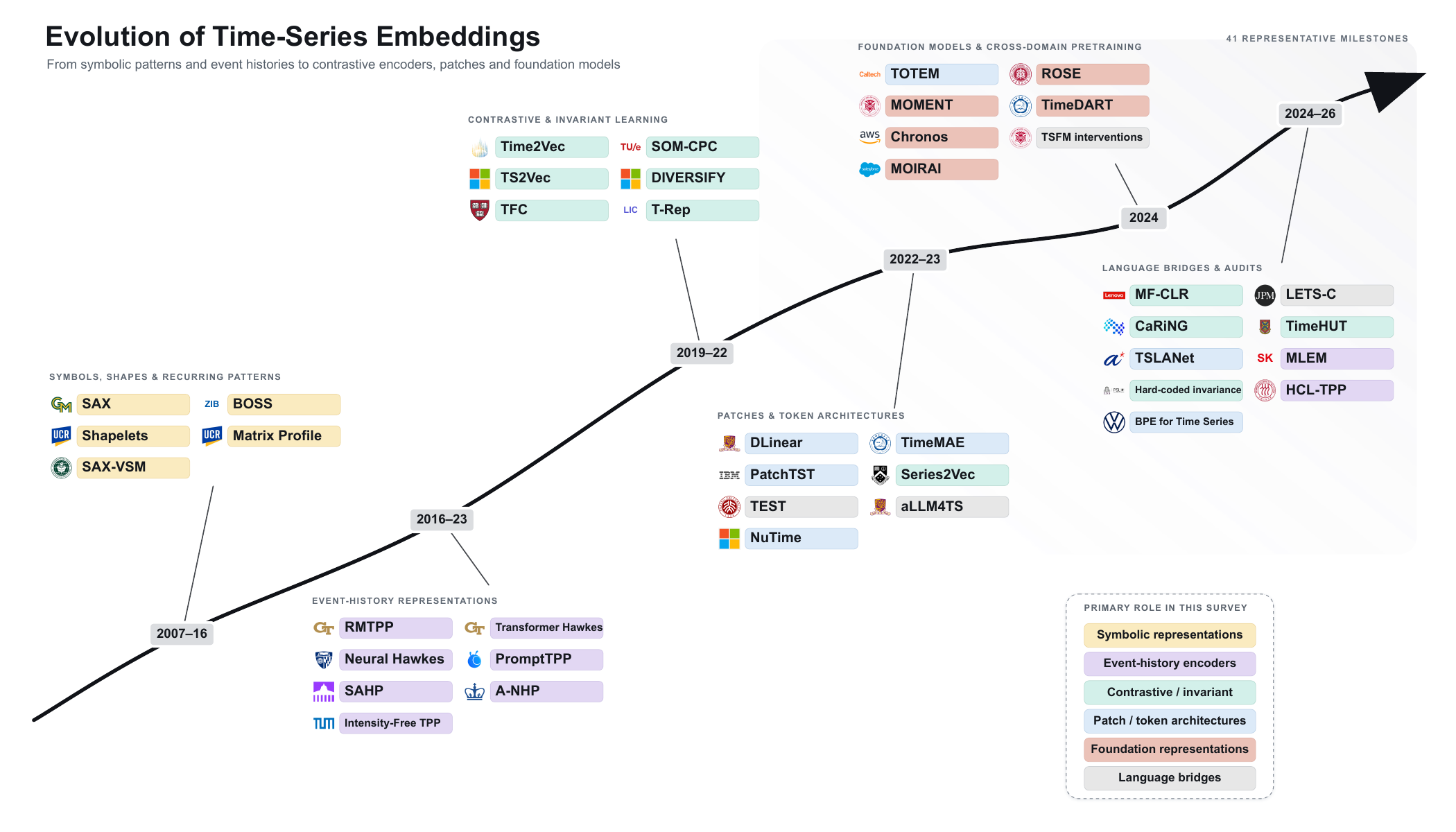}
    \caption{Development history of time-series embeddings.}
    \label{app:fig:timeseries}
\end{figure}

\paragraph{Data objects, units, and granularities.} Sample, channel,
window, patch, complete series, and timestamped event; the window and even
the series boundary are task-dependent, and the recurring units are not
standardized across domains (a QRS complex, a bearing harmonic,
and retail seasonality do not form one vocabulary). Point-process streams add the
mark as an attested discrete token set.

\paragraph{Backbone and inductive bias.} RNN/LSTM encoders carry a
sequential-recurrence prior while TCNs capture temporal context through
dilated causal convolutions; patch Transformers (PatchTST, MOMENT) reuse the
text backbone with a per-channel patch token, and continuous-time LSTMs /
latent ODEs / time-attention add a continuous clock. The counter-evidence
(TSLANet, DLinear) is that attention is not always the right primitive in
small-data, high-noise regimes (\Cref{sec:arch:specific}).

\paragraph{Objectives.} Contrastive over windows, scales, and time--frequency
views (TS2Vec, TF-C); masked patch reconstruction (PatchTST, MOMENT) with the
interpolation-pitfall caveat; probabilistic forecasting pretraining (MOIRAI)
and LLM-bridging serialization (TEST, LETS-C); declared-symmetry objectives
(hard-coded invariances) and magnitude-preserving factorization (NuTime),
which encodes each window's mean and standard deviation separately from its
normalized shape.

\paragraph{Interfaces / exposed object.} A magnitude-preserving global vector
when level matters (amplitude is semantics); a fixed-size patch
embedding otherwise; the event-stream limit exposes a history-conditioned
state whose internal latent is not automatically an evaluated embedding.
A broadly adopted, retrieval-native frozen-encoder instrument is still missing.

\paragraph{Arc.} Features, kernels, and invariants (Era~I) $\to$ contrastive
learning with the augmentation problem (Era~II) $\to$ patching and tokenization
(Era~III) $\to$ foundation models and the LLM bridge (Era~IV)---four eras that,
uniquely, have not cleanly superseded one another.
\Cref{app:fig:timeseries} summarizes 41 representative works.

\paragraph{Modeling milestones.}
\begin{itemize}[leftmargin=2em]
    \item \textbf{Distances, symbols, and shapes: } DTW~\citep{dtw},
    Takens~\citep{takens1981}, SAX~\citep{sax},
    shapelets~\citep{shapelets},
    SAX-VSM~\citep{saxvsm} (tf--idf-weighted SAX words as interpretable
    class vectors),
    BOSS~\citep{boss} (bag-of-SFA-words histograms built to survive noise),
    Matrix Profile~\citep{matrixprofile} (exact all-pairs subsequence joins
    that unify motif, discord, and shapelet search),
    ROCKET~\citep{rocket}
    \item \textbf{Contrastive and invariant learning: }
    T-Loss~\citep{franceschi2019tloss} (a triplet loss over sampled
    subwindows with a dilated-convolution encoder),
    TS-TCC~\citep{eldele2021tstcc} (temporal and contextual contrasting over
    two augmented views),
    TS2Vec~\citep{ts2vec}, TF-C~\citep{tfc}, T-Rep~\citep{trep},
    MF-CLR~\citep{mfclr}, TimeHUT~\citep{timehut},
    Series2Vec~\citep{series2vec} (predicting unsupervised series similarity
    instead of relying on augmentations),
    SOM-CPC~\citep{somcpc} (contrastive predictive coding tied to a
    self-organizing-map codebook for high-rate sensor streams),
    DIVERSIFY~\citep{diversify} (discovering latent sub-domains to learn
    out-of-distribution-robust representations)
    \item \textbf{Patches, tokens, and post-attention encoders: }
    PatchTST~\citep{nie2023times},
    TimeMAE~\citep{timemae} (a decoupled masked autoencoder combining masked
    codeword classification with masked representation regression),
    DLinear~\citep{dlinear}, TSLANet~\citep{tslanet},
    NuTime~\citep{nutime} (magnitude preserving),
    hard-coded invariances~\citep{hardcoded},
    byte-pair encoding of series~\citep{tsbpe}
    \item \textbf{Foundation models and the LLM bridge: }
    MOMENT~\citep{moment}, Chronos~\citep{chronos}, MOIRAI~\citep{moirai},
    TOTEM~\citep{totem}, ROSE~\citep{rose},
    TimeDART~\citep{timedart} (a causal patch encoder trained with a diffusion
    loss in place of autoregressive MSE),
    TEST~\citep{test}, LETS-C~\citep{letsc2025}, aLLM4TS~\citep{allm4ts}
    \item \textbf{Alignment and retrieval: } TS-CLIP~\citep{tsclip},
    CLaSP~\citep{clasp}, TimeRAG~\citep{timerag}, TS-RAG~\citep{tsrag}
    \item \textbf{Event histories (temporal point processes): }
    RMTPP~\citep{rmtpp}, Neural Hawkes~\citep{nhp}, SAHP~\citep{sahp},
    THP~\citep{thp},
    Intensity-Free TPP~\citep{intensityfree} (modeling conditional
    inter-event-time densities directly via normalizing flows and a
    log-normal mixture with closed-form sampling and
    moments instead of an intensity function),
    A-NHP~\citep{attnhp} (attention builds time-varying event and
    participant embeddings),
    Time2Vec~\citep{time2vec} (a learnable, periodic representation of a
    timestamp, used as a drop-in time encoding),
    CoLES~\citep{coles}, HCL-TPP~\citep{hcltpp} (event- and sequence-level
    noise-contrastive estimation on top of maximum likelihood),
    MLEM~\citep{mlem} (masked event reconstruction combined with contrastive
    subsequence learning),
    CaRiNG~\citep{caring} (identifiable temporal causal representations under
    non-invertible generation),
    PromptTPP~\citep{prompttpp} (a retrieval prompt pool that adapts neural
    TPPs to drifting streams), TPP-LLM~\citep{tppllm}
\end{itemize}
\paragraph{Evaluation benchmarks.} The outlier: no broadly adopted
frozen-encoder MTEB analog exists yet. Representation quality is instead
reported as a classification score probed on frozen encodings over the UCR
univariate archive~\citep{dau2019ucrarchive} and the UEA multivariate
archive~\citep{bagnall2018ueaarchive}, or as end-to-end forecasting error on
the Monash forecasting repository~\citep{godahewa2021monash} and the recent
GIFT-Eval~\citep{gifteval}---the former a controlled representation
comparison, the latter a joint property of encoder and forecast head.

\paragraph{Cross-modal pointers.} Continuous axis, no vocabulary, and
magnitude semantics as the distinctive property (\Cref{sec:frontier:properties});
the augmentation problem and declared-symmetry architectures
(\Cref{sec:objectives:contrastive}); serialization (LETS-C),
discretization (Chronos), and rendering (VisionTS)
as transformation (\Cref{sec:transformation}); continuous-time
indexing and magnitude-preserving embedding as exports
(\Cref{sec:frontier:pollination}); the missing broadly adopted frozen-encoder
suite (\Cref{sec:evaluation}).

\subsection{Other Modalities}
\label{app:other}

We review two additional families---3D point cloud and biological
sequences---more compactly since they are more niche and their
embedding literatures are younger.

\paragraph{3D / point cloud.} A point cloud is a sparse, unordered set of
samples over continuous surfaces---an $\mathrm{SE}(3)$-transformable
geometric object rather than a symbol sequence, so its structure is carried
jointly by the input organization and the encoder architecture
(\Cref{sec:arch:specific}).
PointNet established the canonical interface: treat points as a set, apply a
shared pointwise function, and aggregate with a symmetric max-pool that
makes the representation invariant to point order~\citep{qi2017pointnet};
PointNet++ added hierarchical grouping and local aggregation over metric
neighborhoods~\citep{qi2017pointnetpp}. The self-supervised
era then replayed the text recipe in miniature: Point-BERT masks point
groups against a learned discrete vocabulary and recovers them with a
BERT-style encoder~\citep{yu2022pointbert}, Point-MAE masks points and
regresses the missing coordinates with a plain
Transformer~\citep{pang2022pointmae}, and PointGPT moves to the generative
corner, auto-regressively predicting point patches as GPT predicts
tokens~\citep{chen2024pointgpt}. From there the family enters the same
unification routes as every other modality: on the transformation path
(\Cref{sec:transformation}), PointCLIP projects points into multi-view
depth maps so a pretrained vision--language model can read them
directly~\citep{zhang2022pointclip}; on the alignment path
(\Cref{sec:alignment}), ULIP contrasts point
clouds into the pretrained CLIP image--text space~\citep{xue2023ulip},
OpenShape scales zero-shot 3D recognition over language
descriptions~\citep{liu2023openshape}, and Uni3D pushes the point encoder
itself to a billion parameters~\citep{zhou2024uni3d}.
Evaluation, like time series, has no broadly adopted MTEB analog:
representation quality is read out through trained heads on classification and
part-segmentation suites and, on the alignment path, through zero-shot
recognition in a text-aligned space.

\paragraph{Molecules and proteins.} The biomolecular family pairs two
data objects at very different scales: small molecules, whose atoms form
either molecular graphs serializable as SMILES strings
or 3D conformations with explicit atomic coordinates, and proteins,
whose long amino-acid sequences read ever more like a language. That is
precisely what makes them interesting here---each objective family of
\Cref{part:principles} has been run on them. The static word-vector analog is mol2vec,
which trained word2vec verbatim over molecular
substructures~\citep{jaeger2018mol2vec}; a year earlier, SchNet had replaced
engineered fingerprints with continuous-filter message passing over
atomic positions~\citep{schutt2017schnet}. ChemBERTa then ran the BERT
recipe directly on the SMILES
string~\citep{chithrananda2020chemberta}, and GROVER scaled self-supervised
graph pretraining to ten million molecules with a
hundred-million-parameter model~\citep{rong2020grover}.
On the protein side, ESM-2 showed that as protein language models grow,
atomic-level structure emerges directly in the sequence
embedding~\citep{lin2023esm2}---representation quality scaling with
parameters, mirroring the scaling trajectory observed in vision at the molecular level.
The alignment bridge follows the CLIP template as well: MoleculeSTM aligns molecular
graphs with natural language~\citep{liu2023moleculestm}, and CLAMP
pretrains molecules contrastively against assay
descriptions~\citep{seidl2023clamp} (\Cref{sec:alignment}); the modality
gap of \Cref{sec:alignment:gap} persists across very different modality
pairs, amino-acid sequence encoders included~\citep{liang2022modalitygap}. Evaluation again runs through
task heads---property prediction for molecules, structure and function
prediction for proteins---with no embedding-native benchmark on the
horizon yet (\Cref{sec:evaluation}).

\clearpage
\section{Quantitative Analysis of the Literature}
\label{app:quant}

This section grounds three claims of the main text in simple bibliometric
evidence: that embedding research is timely (\ie the object of study is
growing, not receding), that it remains fragmented across communities
(\Cref{sec:intro:silos}), and that early signals of cross-modal convergence
are measurable at the level of publication venue and terminology. Two views
are used: a title-term baseline over arXiv metadata for 2015--2025, and a
manually curated census of formally published conference and journal papers
for 2023--2025.

\subsection{An arXiv title-term analysis (2015--2025)}
\label{app:quant:arxiv}

\paragraph{Method.}
For each year from 2015 to 2025, we queried the arXiv API for every record
in \texttt{cs.CL}, \texttt{cs.CV}, \texttt{eess.AS}, \texttt{cs.LG},
\texttt{cs.IR}, and \texttt{cs.DB} (\ie the main categories for text,
vision, audio processing and other major embedding modalities) whose
title matches the keyword \emph{embedding} or \emph{representation learning}.
Among the matches, 767 records whose only match is a spurious stem (typically
``embedded devices'' or ``embedded systems'') and 87 records whose sole use of the word
``embedding'' is inside a position-embedding-style phrase (\eg
``positional encoding with rotary position embeddings'') are removed. This leaves 10{,}952 records out of 11{,}806
candidates.

\begin{figure*}[th]
    \centering
    \includegraphics[width=\textwidth]{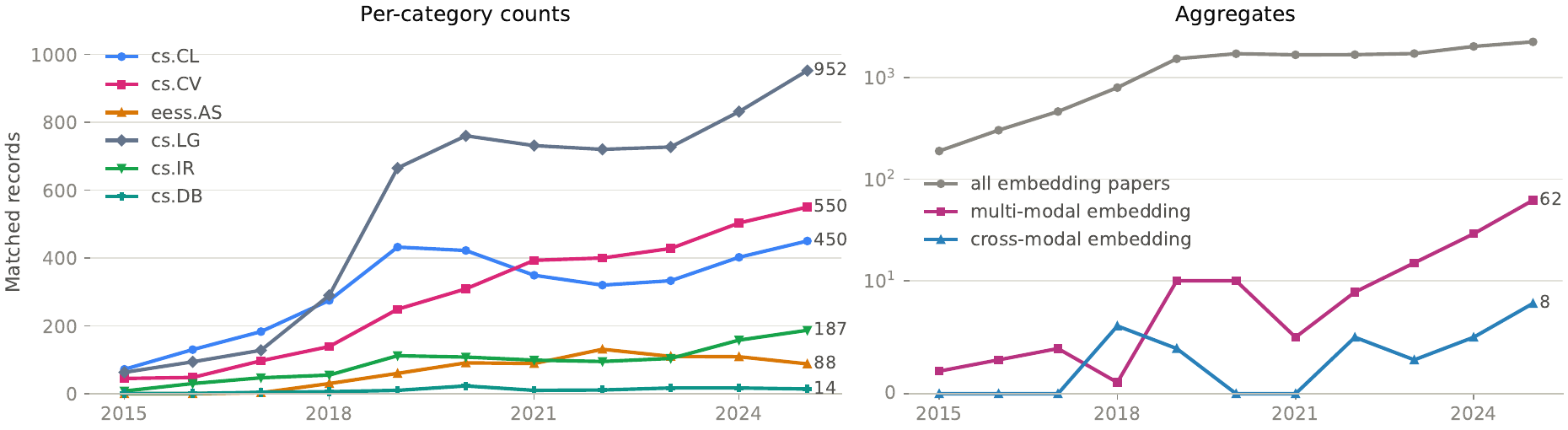}
    \vspace{-0.3cm}
    \caption{\textbf{arXiv title-term analysis, 2015--2025.}
    \emph{Left:} yearly counts of arXiv records containing the keyword \emph{embedding} or \emph{representation learning} in \texttt{cs.CL}, \texttt{cs.CV}, \texttt{eess.AS}, \texttt{cs.LG}, \texttt{cs.IR}, and \texttt{cs.DB} (\ie text, vision, audio, machine learning, information retrieval, and databases).
    \emph{Right:} a comparison between \emph{all embedding papers} and those containing the keyword \emph{multimodal embedding} and \emph{cross-modal embedding}.}
    \label{fig:app-arxiv}
\end{figure*}

\paragraph{Findings.}
\Cref{fig:app-arxiv} shows the resulting findings. First, the field is
large and mostly mature: kept records grow from 140 in 2015 to 1{,}784 in
2025, but growth has slowed---the union count increased by 13\%
from 2024 to 2025, against a 73\% single-year jump from 2018 to
2019---and \texttt{eess.AS} remains an order of magnitude
smaller than the three largest categories, consistent with
the minority position of speech noted in this paper. Second, the center
of gravity has moved: \texttt{cs.CL} peaked in 2019 (432) and was
overtaken by \texttt{cs.CV} in 2021, which reached 550 in 2025; both
trail \texttt{cs.LG}---the general machine-learning category and the
largest of the six since 2018, at 952 records in 2025---while
\texttt{cs.IR} grows steadily to 187 and \texttt{cs.DB} stays marginal.
Third, and most relevant to our argument, multimodal keyword
families behave very differently on this slow-moving baseline: titles containing
\emph{multimodal embedding} (\emph{representation learning} is counted as a synonym of \emph{embedding}) never exceed ten per year before 2023, then rise to 15, 29, and 62
in 2023--2025---a $2.1\times$ single-year jump from 2024 to 2025, against
the 13\% total growth---bringing their share of embedding-titled records
from roughly 1\% or less for most of the decade to 3.5\% in 2025. The
\emph{cross-modal embedding} family, by contrast, never exceeds eight
titles per year across the whole window. In other words, the vocabulary
that names the object of this survey---a unified multimodal embedding
space---is expanding, and it is expanding specifically on the
``multimodal'' side rather than through pairwise ``cross-modal'' phrasing.

\subsection{A curated census of conference papers (2023--2025)}
\label{app:quant:conf}

\paragraph{Method.}
The second view covers works published at peer-reviewed venues.
We manually collected and classified
papers on embedding and representation learning across nine modality
buckets (\emph{text, code, vision, audio, graph, time series,
recommendation, others}, and \emph{cross-modal alignment} as a
deliberately cross-cutting bucket), from the major venues of the different
communities: the NLP conferences (ACL, EMNLP, NAACL),
the vision conferences (CVPR, ICCV, ECCV, ACM MM),
the speech conferences (InterSpeech, ICASSP),
the ML conferences (ICLR, NeurIPS, ICML, and TMLR),
and IR and related venues (SIGIR, WWW, KDD, TPAMI).

\begin{figure*}[th]
    \centering
    \includegraphics[width=0.63\textwidth]{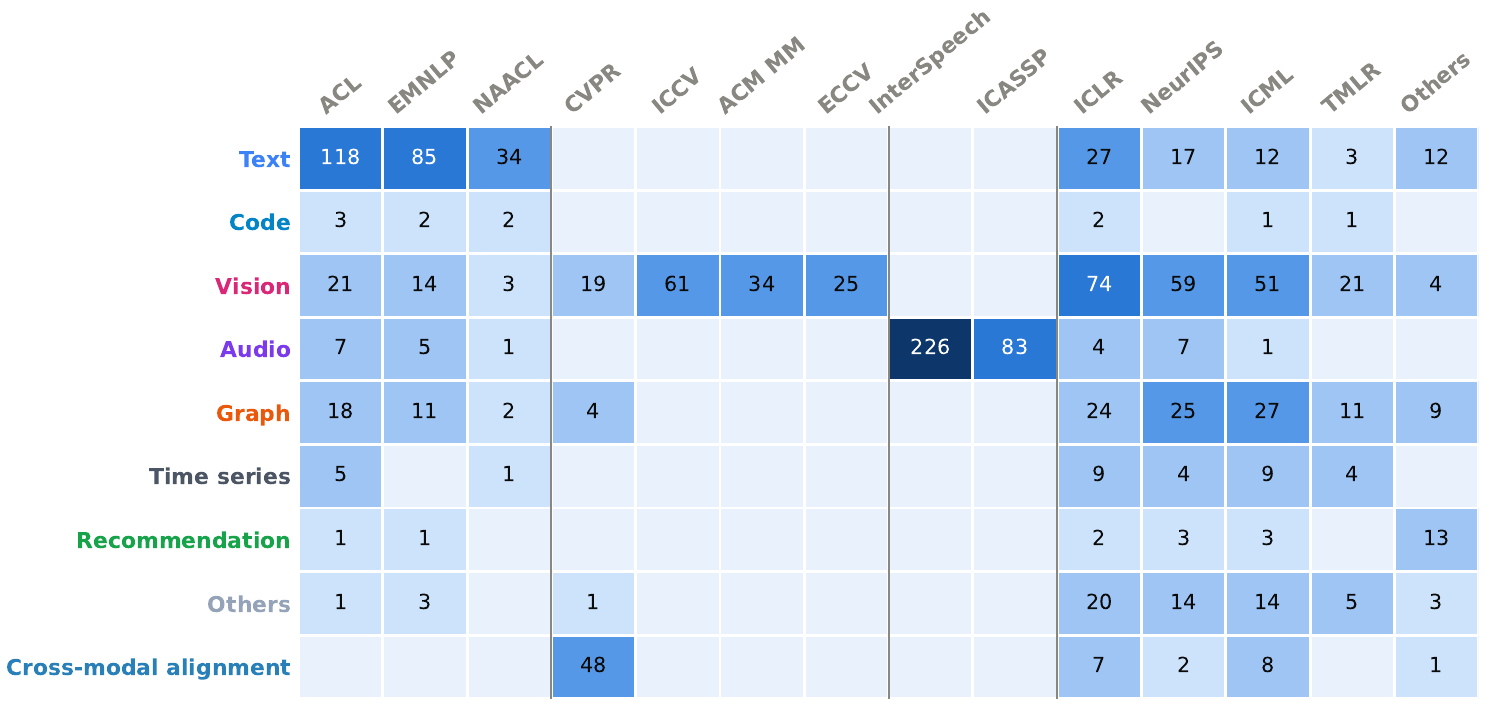}
    \hfill
    \includegraphics[width=0.35\textwidth]{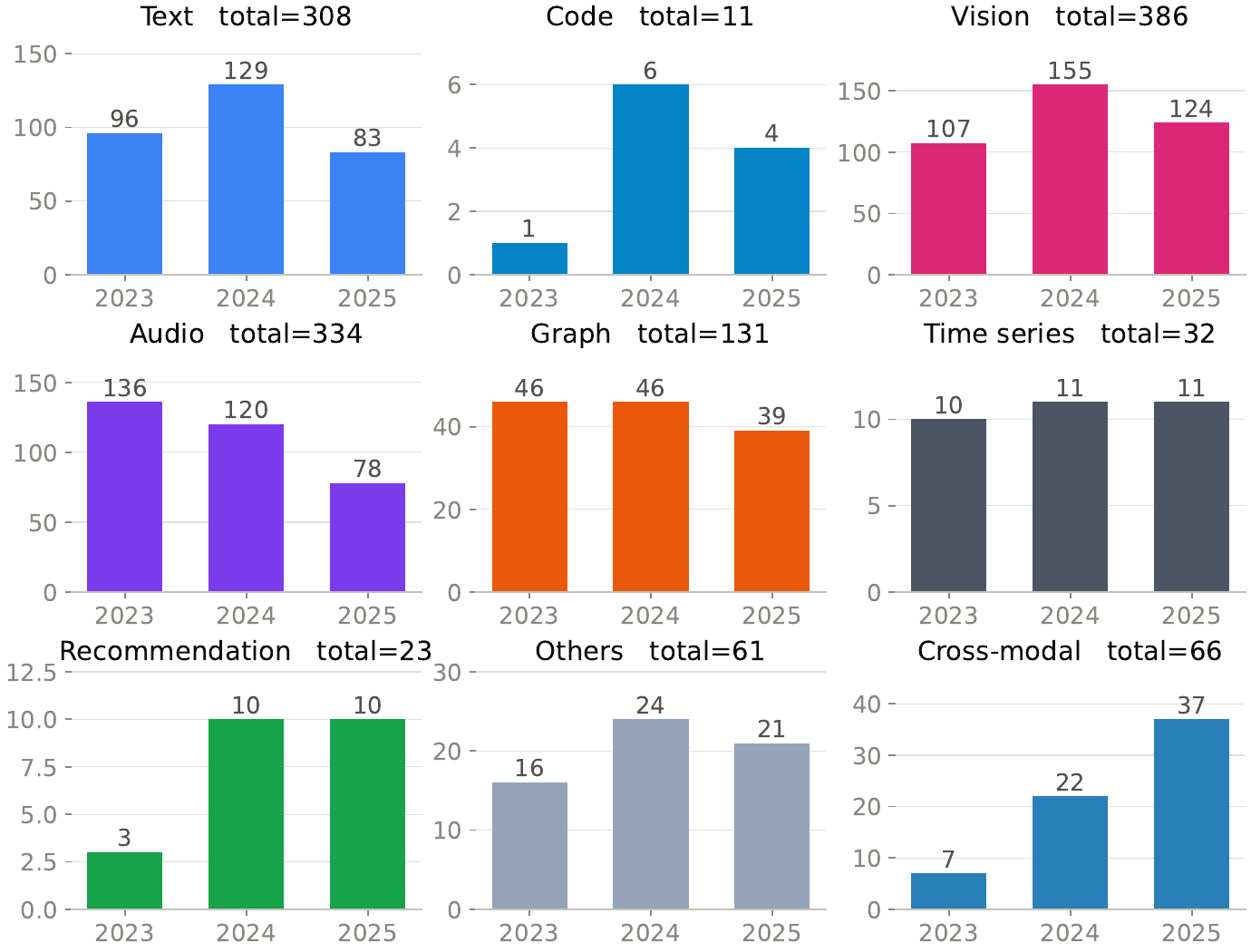}
    \caption{\textbf{Curated conference and journal papers on embedding
    and representation learning, 2023--2025} (1{,}352 formally published
    papers across nine modality buckets). \emph{Left:} modality $\times$
    venue matrix, pooled over the three years;
    the last `Others' column includes SIGIR, WWW, KDD, and TPAMI. \emph{Right:}
    yearly totals per bucket, with cross-modal alignment highlighted:
    $7 \to 22 \to 37$, the fastest-growing bucket in the census while
    most buckets are flat or declining.}
    \label{fig:app-conf}
\end{figure*}

\paragraph{Findings.}
\Cref{fig:app-conf} shows the venue matrix and yearly totals. The
fragmentation claimed in \Cref{sec:intro} is directly visible as structure:
text embeddings live overwhelmingly in the NLP conferences (237 of 308),
audio embeddings in the two speech conferences (309 of 334), and
recommendation at SIGIR, while vision, graph, and time series spread across the ML conferences---ICLR, NeurIPS, and
ICML alone host 205 of the 386 vision-bucket papers. Meanwhile the
cross-cutting bucket behaves unlike any of them: cross-modal alignment
papers are concentrated at CVPR (48 of 66) and the bucket is the only one
that roughly doubles twice in succession, from 7 to 22 to 37 collected
papers across 2023--2025. Speech is also the only bucket in visible
retreat (136$\to$120$\to$78), while remaining the most venue-isolated.
These observations are consistent with the arXiv analysis above: the
communities are siloed by venue precisely where they are largest, and the
fastest-growing published work is the work that crosses modality
boundaries.

\subsection{Limitations}
\label{app:quant:limits}

Both views are deliberately narrow, and we state their limitations here so
their evidence is not over-read. (i) The queries are phrase-based: they
exclude neighboring terminologies (\eg ``unified representation
space''), so all counts are lower bounds for the family of interest
rather than accurate estimates of it. (ii) Title-level matching biases the sample
toward work that signals its subject in the title; methodological work
that uses embeddings without naming them is invisible. (iii) The
counts only measure where the field's attention sits,
how siloed its publishing venues remain, and which corner of the
literature is growing fastest---not whether any
particular embedding construction preserves complementary information,
models cross-modal interactions, or improves downstream reasoning. Those
questions require the controlled evidence surveyed in the main text.

\bibliographystyle{colm2024_conference}
\bibliography{custom}

\end{document}